\documentclass[journal, 10pt]{IEEEtran}  

\IEEEoverridecommandlockouts                              

\usepackage{graphics} 
\usepackage{cite}
\usepackage[T1]{fontenc}
\usepackage{amsmath} 
\usepackage{amssymb}  
\usepackage{hyperref}
\hypersetup{
    colorlinks=true,
    linkcolor=blue,
    filecolor=magenta,      
    urlcolor=blue,
}

\usepackage{pst-all}
\usepackage{epsfig}
\usepackage{pst-plot,pst-text}
\usepackage{pst-grad}
\usepackage{relsize}    
\usepackage{graphicx}
\usepackage[T1]{fontenc}
\usepackage{xcolor}
\usepackage{pst-circ}
\usepackage{fontawesome}
\usepackage{colortbl}
\usepackage{hhline}

\usepackage{comment}
\usepackage[utf8]{inputenc}
\usepackage{pgfplots}
\usepackage{pgf}
\usepgfplotslibrary{groupplots,dateplot}
\usetikzlibrary{patterns,shapes.arrows}
\pgfplotsset{compat=newest}
\usetikzlibrary[pgfplots.groupplots]

\usepackage{color, colortbl}
\definecolor{LightCyan}{rgb}{0.88,1,1}
\usepackage{multirow}
\usepackage{siunitx}
\DeclareSIUnit\RPM{rpm}
\DeclareMathOperator{\sinc}{sinc}
\usepackage{color, colortbl}
\definecolor{LightCyan}{rgb}{0.88,1,1}
\usepackage{multirow}
\usepackage{siunitx}
\DeclareSIUnit\RPM{rpm}
\DeclareSIUnit\pixel{pixel}
\DeclareSIUnit\pixels{pixels}

\usepackage{enumitem} 

\usepackage{wrapfig}
\usepackage{xcolor}
\usepackage{amssymb}
\usepackage{tikz}
\usetikzlibrary{shapes,arrows}
\usepackage{verbatim}
\usetikzlibrary{positioning}
\usepackage[normalem]{ulem} 
\usepackage{tabstackengine}
\usepackage{algorithm}
\usepackage[end]{algpseudocode} 
\usepackage{etoolbox}
\usepackage{subcaption}

\usepackage{eso-pic}
\newcommand\AtPageUpperMyright[1]{\AtPageUpperLeft{%
 \put(\LenToUnit{0.5\paperwidth},\LenToUnit{-1.5cm}){%
     \parbox{0.5\textwidth}{\raggedright\fontsize{11}{11}\selectfont #1}}%
 }}%
\newcommand{\conf}[1]{%
\AddToShipoutPictureBG*{%
\AtPageUpperMyright{#1}
}
}

\title{\LARGE \bf \textit{MPPI-VS}: Sampling-Based Model Predictive Control Strategy for Constrained Image-Based and Position-Based Visual Servoing}

\conf{\hspace*{-9cm}\textcolor{gray}{This paper is an extension of \cite{mohamed2021sampling} that has been published in the \textit{2021 IEEE/RSJ International Conference on Intelligent Robots and Systems \textcolor{blue}{\href{https://www.iros2021.org/}{(IROS 2021)}}}} }

\author{Ihab S. Mohamed$^{1}$ 
\thanks{$^{1}$Ihab S. Mohamed is with  the Luddy School of Informatics, Computing, and Engineering, Indiana University, Bloomington, IN 47408 USA (e-mail: {\tt\small mohamedi@iu.edu)}}%
}

\begin{document}

\maketitle

\thispagestyle{plain} 
\pagestyle{plain}

\begin{abstract}
In this paper, we open up new avenues for visual servoing systems built upon the \textit{Path Integral (PI)} optimal control theory, in which the non-linear partial differential equation (PDE) can be transformed into an expectation over all possible trajectories using the \textit{Feynman-Kac} (FK) lemma. 
More precisely, we propose an \textit{MPPI-VS} control strategy, a \textit{real-time} and \textit{inversion-free} control strategy on the basis of sampling-based model predictive control (namely, Model Predictive Path Integral (\textit{MPPI}) control) algorithm, for both image-based, \textit{3D} point, and position-based visual servoing techniques, taking into account the system constraints (such as visibility, 3D, and control constraints) and parametric uncertainties associated with the robot and camera models as well as measurement noise. 
Contrary to classical visual servoing control schemes, our control strategy directly utilizes the approximation of the interaction matrix, without the need for estimating the interaction matrix inversion or performing the pseudo-inversion. 
We validate the \textit{MPPI-VS} control strategy as well as the classical control schemes on a 6-DoF Cartesian robot with an \textit{eye-in-hand} camera based on the utilization of four points in the image  plane as visual features.
To better assess and demonstrate the robustness and potential advantages of our proposed control strategy compared to classical schemes, intensive simulations under various operating conditions are carried out and then discussed.
The obtained results demonstrate the effectiveness and capability of the proposed scheme in coping easily with the system constraints, as well as its robustness in the presence of large errors in camera parameters and measurements.



\end{abstract}
\section*{Multimedia Material}
The supplementary video attached to this work is available at:
\url{https://bit.ly/3JrSJUu}
\section{INTRODUCTION}\label{Introduction}

\IEEEPARstart{V}{isual} servoing, also known as vision-based control, has been widely used in robotics and automation society, providing more flexible and intelligent machines utilized in many applications such as assembly tasks \cite{hamner2010autonomous}, detection and tracking missions 
\cite{minaeian2015vision, qiu2019visual, do2019perception}, surgical tasks \cite{azizian2014visual}, unmanned aerial vehicles 
\cite{serra2016landing}, ultrasound probe guidance \cite{mebarki20102}, welding \cite{zhou2006autonomous}, and spray painting \cite{chen2017fringe}. Generally speaking, visual servoing refers to the use of computer vision data, acquired from one or several cameras, for controlling the motion of a robot. 
Typically, the camera has two fundamental configurations. First, it can be mounted directly on a robot (e.g., robot manipulator or mobile robot), where the robot motion produces the motion of the camera. Second, it can be fixed in the workspace, in order to observe the motion of the robot. The former configuration refers to \textit{eye-in-hand systems}, whereas the latter indicates \textit{eye-to-hand systems}. 
From the control schemes perspective, visual servoing can be typically classified into three categories: (i) image-based visual servoing (\textit{IBVS}), in which the control law is directly based on the error between the current and desired image features on the \textit{2D} image plane;
(ii) position-based visual servoing (\textit{PBVS}), where the pose of the camera, relative to some reference coordinate frame, is computed to be used by the control law  \cite{hutchinson1996tutorial};
and (iii) hybrid visual servoing, in which, one way or another, a combination of \textit{IBVS} and \textit{PBVS} is utilized \cite{malis19992}. 
 
 Generally, the \textit{IBVS} control scheme has received much attention both in scientific literature and in industry-oriented research due to its inherent robustness against not only camera calibration imperfections but also against modelling errors \cite{espiau1993effect}, exhibiting better local stability and convergence properties. 
 In contrast to \textit{PBVS} control strategy that is often very sensitive to camera calibration errors, which lead to errors in the \textit{3D} reconstruction of the target object in the environment and accordingly errors during task execution. 
 In addition, as the control law of \textit{PBVS} is explicitly expressed in Cartesian space and there is apparently no direct control to the visual features on the image plane, these features or some may leave the camera's field of view (FoV).
 Furthermore, compared with \textit{PBVS}, \textit{IBVS} does not require a perfect knowledge of the \textit{3D} model of the target, and only some information about the object depths are required. 
 Nevertheless, there exist some issues associated with \textit{IBVS}, which should be addressed, such as: (i) its convergence is theoretically constrained to a region around the desired camera pose; (ii) in some cases, singularities (or poor conditioning) in the interaction matrix\footnote{In \textit{IBVS}, the mapping between visual features velocities in image space and camera velocity in Cartesian space is encoded in the interaction matrix, which will be briefly discussed in Section \ref{VS-model}.} and image local minima may occur, leading to control problems and probably spilling down the servoing task; (iii) since there is no direct control over the camera velocity in Cartesian space, as the control law is defined in the image plane, the executed trajectories by the robot, in the Cartesian space, could be quite contorted; (iv) other concern is the difficulty of constraints handling such as visibility constraints, which imposed by the fact that the visual features should constantly remain within the camera's FoV. It is noteworthy that these major problems have been clearly pointed out in \cite{chaumette1998potential}. 
\subsection{Related Work}
To this so, several control approaches have been proposed in the literature, so as to improve the visual servoing performance and overcome the previously highlighted difficulties. 
Just to name a few, the problem of image singularities could be solved by finding the suitable visual features of visual servoing, such as Cylindrical \cite{iwatsuki2002new}, Spherical \cite{fomena2009improvements}, Polar \cite{liu2020robust} 
coordinate systems, and moment features \cite{chaumette2004image}. A good alternative solution based on the task function approach is investigated in \cite{marchand1996using}, for avoiding robot joint limits and kinematic singularities. While, in \cite{malis2004improving}, several control schemes based on second-order minimization techniques have been conducted for avoiding the camera retreat problem (as introduced in \cite{chaumette1998potential}). In \cite{siradjuddin2013image}, the authors proposed a Takagi–Sugeno fuzzy framework for modelling the \textit{IBVS} scheme, in which the singularity can be handled and the stability can be easily verified. 
Another improvement is achieved by so-called 2-1/\textit{2D} visual servoing approach, which is one of the well-known hybrid schemes, where the characteristics of the \textit{IBVS} and \textit{PBVS} methods are combined \cite{malis19992}. Similarly, soon afterwards, authors in \cite{cervera1999combining} proposed a new approach based on augmenting depth information within the visual features vector. 
While Thuilot et  al. \cite{thuilot2002position}
proposed a new control strategy for \textit{PBVS} in which an online trajectory is planned on the image plane in order to ensure that the object remains within the FoV.
In addition, other hybrid or partitioned schemes have been developed based on (i) decoupling the translational camera motions from the rotational ones \cite{corke2001new, tahri2010decoupled, xu2017partially}; or (ii) switching either between different control schemes (i.e., \textit{IBVS} and \textit{PBVS} control schemes) \cite{gans2003asymptotically, chesi2004keeping, gans2007stable} or between two different coordinate systems \cite{allibert2012switching, ye2015novel}. For instance,  
Allibert and Courtial \cite{allibert2012switching}
proposed a switching control scheme for \textit{IBVS} tasks based on Cartesian and Polar image coordinates, taking the advantages of both Cartesian- and Polar-based \textit{IBVS} strategy. Other promising methods based on combining path planning and trajectory tracking were developed in
the literature for coping with constraints handling (see, e.g., \cite{kazemi2010path} and \cite{zheng2017planning}).

Recently, model predictive control (\textit{MPC}) strategies have been widely used in the related literature, with the aim of improving the quality of visual servoing and coping with constraints which have not been explicitly handled by most of the previously mentioned approaches. 
For example, an unconstrained stabilizing receding horizon control strategy is applied to \textit{3D} visual servoing \cite{murao2006predictive}, while other predictive control approaches have been proposed to deal with constraints in \textit{IBVS} \cite{allibert2010predictive, heshmati2014robustness, wang2015quasi, qiu2019model1, roque2020fast}. Particularly, in \cite{allibert2010predictive}, an \textit{MPC}-based approach is proposed for constraints handling and image prediction, demonstrating its robustness with respect to errors in the camera parameters and noise in the measured visual features. Nonetheless, no robustness (i.e., stability) analysis was performed, which has been lately addressed, e.g., in \cite{heshmati2014robustness} and \cite{ roque2020fast}, and the controller has not been experimentally validated. Based on the polytopic transformation of the interaction matrix, a robust quasi-min-max \textit{MPC} strategy was proposed in \cite{wang2015quasi}, which was also validated via simulation studies.
The two proposed approaches in \cite{allibert2010predictive} and \cite{wang2015quasi} suffer from the computational burden since solving the optimization problem exceeds the real system-sampling time (i.e., they are not implemented \textit{online}). Sequentially, it can be quite difficult to be applied to the real system. In order to alleviate and tackle this problem, various methods have been developed to guarantee a real-time solving optimization problem such as in \cite{ke2016visual, hajiloo2015robust, fusco2020integrating}.
In \cite{ke2016visual}, the authors utilized the primal-dual neural network (PDNN) as a promising computational tool for solving the quadratic programming (QP) problem and achieving a real-time implementation of \textit{MPC}, instead of using the sequential quadratic programming (SQP) method which requires repeatedly a calculation of Hessian matrix to solve a QP problem \cite{hajiloo2015robust}.
Fusco et al. \cite{fusco2020integrating} presented a real-time \textit{IBVS} prediction control scheme for a robotic arm based on integrating the acceleration of the visual features, which allows the controller to produce better input signals compared to classical predictive strategies.

On the other hand, a new algorithmic methodology based on Path Integral (PI) optimal control theory has recently been proposed by Kappen in \cite{kappen2005path}, for solving the nonlinear Stochastic Optimal Control (SOC) problem. Traditionally, on the basis of dynamic programming, the SOC problem is defined by a partial differential equation (PDE) known as the \textit{Hamilton-Jacobi-Bellman} (HJB) equation. More generally, this equation can not be solved analytically. On other words, it can only be solved numerically \textit{backward-in-time} and, unfortunately, numerical solutions are intractable due to the curse of dimensionality \cite{oksendal2013stochastic}. 
This is actually one of the primary reasons for proposing a \textit{PI control theory}, in which the non-linear PDE can be transformed into an expectation over all possible trajectories using the \textit{Feynman-Kac} (FK) lemma. This transformation allows the SOC to be solved by sampling methods, such as \textit{forward-in-time} Monte-Carlo approximation, instead of solving the HJB equation \textit{backward-in-time}. 
Inspired by the \textit{PI control theory}, Williams et al. \cite{williams2017model} proposed a sampling-based model predictive control algorithm known as Model Predictive Path Integral (\textit{MPPI}) control framework, which has been successfully applied to a variety of robotic systems for tasks such as aggressive autonomous driving and autonomous flying through \textit{2D} cluttered environments. 
More recently, a generic and elegant \textit{MPPI} control framework has been presented in \cite{mohamed2020model}, which enables the robot to navigate autonomously in either \textit{2D} or \textit{3D} environments which are inherently uncertain and partially observable.

It is obviously sufficiently that both \textit{MPPI} and \textit{MPC} follow the same control strategy, which can be summarized as: (i) the optimal control action is obtained by solving, at each time step, a finite-horizon constrained optimization problem, considering the current state of the system to be controlled as the initial state; (ii) then, the optimization yields an optimal control sequence; (iii) finally, the first control in this sequence is applied to the system being controlled. 
However, we believe that the \textit{MPPI} control framework significantly outperforms the conventional \textit{MPC} strategy for the following reasons. 
First, since it is a sampling-based and derivative-free optimization method, it does not require the computation of gradients (i.e., derivatives) so as to find the optimal solution. As a consequence, it can be readily implemented \textit{online} and applied to the real system.
Second, it does not require the first- or second-order approximation of the system dynamics and quadratic approximation of the cost functions, i.e., highly non-linear and non-convex functions can be naturally utilized. Moreover, this flexibility also allows the dynamics to be easily represented by neural networks \cite{williams2017information}.
Third, its ability to cope with hard and soft constraints easily, making it so attractive in many robotics fields. Additionally, discontinuous cost functions, i.e., indicator functions, can be easily handled and added to the running cost function. For instance, in the context of autonomous flying tasks, a \textit{large-weighted} indicator function can be employed as part of the running cost, for penalizing the collision with ground or obstacles.
Forth, \textit{MPPI} can still provide a reasonable solution when there exists \textit{no feasible} solution, which (i.e., feasibility issue) represents one of the most predominant issues in \textit{MPC} schemes that adds additional complexity to the optimization problem \cite{mayne2000constrained}.
\subsection{Contributions}
 Motivated by the observations above, within this paper, we present the framework of \textit{PI control theory} to the visual servoing systems. 
 More precisely, we propose a \textit{real-time} and \textit{inversion-free} control method on the basis of the \textit{MPPI} control framework for both image-based (\textit{IBVS}), \textit{3D} point (\textit{3DVS}), and position-based (\textit{PBVS}) visual servoing control schemes, which has been validated on a 6-DoF Cartesian robot (namely, Gantry robot) with an \textit{eye-in-hand} camera. 
 We call this new approach the \textit{MPPI-VS} control strategy.
Consequently, the underlying goal of the work described in this paper is two-fold: 
 \begin{enumerate}
     \item examining the possibility of employing the \textit{MPPI} control framework for visual servoing (VS) systems, particularly, in our case, for VS systems based on the utilization of four points in the image plane as visual features; to the best of the authors' knowledge, this is the first attempt to develop a VS control strategy based on \textit{MPPI};
     \item highlighting its potential advantages for improving the quality of visual servoing, in terms of the image prediction, constraints handling, and overcoming other aforementioned difficulties as far as possible.
\end{enumerate}
The major contributions of the work described in this paper can be summarized as follows:
\begin{itemize}
\item We propose a \textit{real-time} sampling-based \textit{MPC} algorithm for predicting the future behavior of the VS systems, without solving the online optimization problem which usually exceeds the real system-sampling time and suffers from the computational burden.
    \item No need for estimating the interaction matrix inversion or performing the pseudo-inversion in real-time \cite{lapreste2004efficient}; our proposed \textit{MPPI-VS} control strategy directly utilizes the approximate interaction matrix, i.e., it is an \textit{inversion-free} control method.
    \item We also propose a direct estimation (i.e., online estimation) of \textit{3D} parameters (i.e., depth $Z_i(t)$) of the interaction matrix under the assumption that the initial object depths, $Z_i(0)$, are known; on other words, no need for solving a set of linear equations using least-squares or using iterative updating schemes \cite{piepmeier2004uncalibrated}.
    \item The system constraints (namely, visibility, three-dimensional (i.e., \textit{3D}), and control constraints) and parametric uncertainties associated with the robot and camera models can be easily posed and handled.
    \item For \textit{IBVS}, the proposed approach (so-called \textit{MPPI-IBVS}) has the capability of tackling the camera retreat problem without proposing hybrid methods or decoupling the $z$-axis motions from the others degrees of freedom \cite{corke2001new}, thanks to the prediction process which allows to explicitly enforce better behaviour of the camera's motion. 
    \item Finally, for \textit{PBVS}, we propose two methods to ensure that the object always remains within the camera's FoV since the control law is explicitly expressed in Cartesian space and there is no direct control to the visual features on the image plane.
\end{itemize}

This paper is an extension of our previously-published study in \cite{mohamed2021sampling}, with a more detailed explanation and analysis.
As an improvement over the previous study, this article demonstrates in detail the intensive simulations jointly with a set of examples from the successful servoing tasks that shows the behavior of the proposed control strategy compared to the classical control schemes, including: (i) the trajectories in both the image and Cartesian space, (ii) the camera velocity components, and (iii) the visual features error (for instance, see Fig.~\ref{fig:s113_ibvs}). Moreover, it studies the convergence time of the proposed control schemes, as well as how the camera local minimum can be effectively avoided. 

\subsection{Paper Organization}
The rest of this paper is organized as follows. In Section~\ref{VS-model}, we briefly recall the mathematical formulation of the classical \textit{IBVS}, \textit{3DVS}, and \textit{PBVS} control schemes for a pinhole camera model.
Section~\ref{MPPI Control Strategy} describes our proposed \textit{MPPI-VS} control strategy, detailing how the system constraints can be handled, whereas Section~\ref{Simulation Details and Results} is dedicated to the intensive simulation results and discussion. Finally, conclusions and future work are given in Section~\ref{sec:conclusion}.

\section{Classical Visual Servoing Control Schemes}\label{VS-model}
Broadly speaking, the main objective of all vision-based control schemes is to minimize the error $\mathbf{e}(t)$ between the current visual features $\mathbf{s}(t)$ and the desired features $\mathbf{s}^{*}$, which is typically defined as
\begin{equation}\label{eq:e(t)}
   \mathbf{e}(t)=\mathbf{s}(t)-\mathbf{s}^{\ast}, 
\end{equation}
assuming, herein, that the camera observes stationary visual features (i.e., $\mathbf{s}^{\ast}$ is constant) and any changes in $\mathbf{s}$ depend only on the motion of the camera. 
\begin{figure}[!ht]
\begin{center}
\includegraphics[scale=0.75]{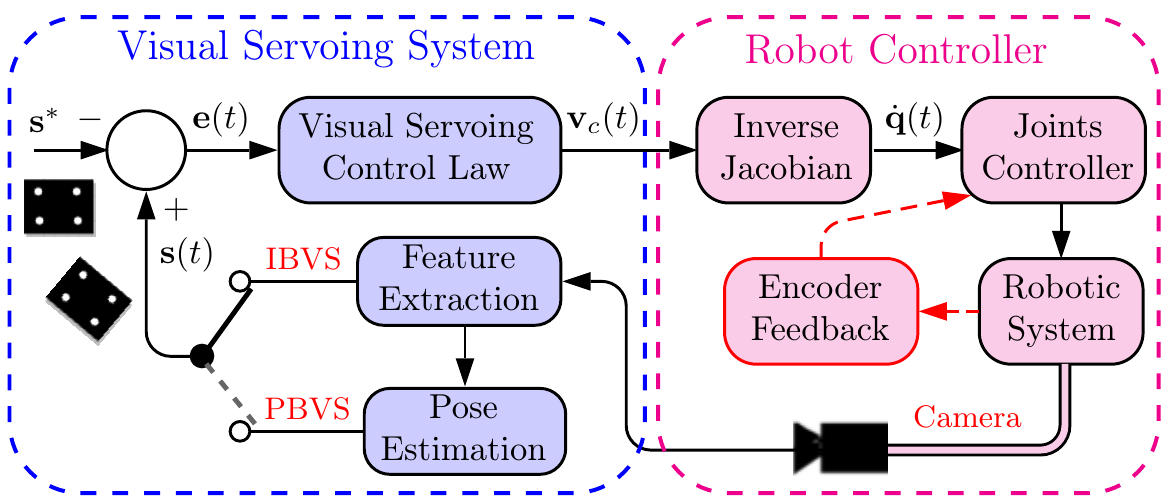}
\caption{Global structure of \textit{IBVS} and \textit{PBVS} control schemes.}
\label{fig:VS-framework}
\end{center}
\end{figure}
The global structure of both \textit{IBVS} and \textit{PBVS} control schemes, for a 6-DoF Grandy robot, is illustrated in Fig.~\ref{fig:VS-framework}.   
The main difference between these control schemes is in the way that $\mathbf{s}$ is designed. 
More specifically, for \textit{IBVS} control schemes, $\mathbf{s}$ consists of a set of image features that expressed directly in the \textit{2D} image space; 
whilst for \textit{PBVS} control schemes, it is composed of a set of \textit{3D} parameters that estimated from the visual features extracted from the image. 
In this section, we briefly recall the mathematical formulation of the classical \textit{IBVS}, \textit{3DVS}, and \textit{PBVS} (namely, \textit{C-IBVS}, \textit{C-3DVS}, and \textit{C-PBVS}) control schemes for a pinhole camera model, considering only the utilization of a set of point-like features as visual features extracted from a \textit{2D} image.
\subsection{Image-Based Visual Servoing (IBVS)}\label{classic-ibvs}
Consider $\boldsymbol{P}=[X, Y, Z]^T \in \mathbb{R}^{3}$ as coordinates of a \textit{3D} point feature expressed in the camera frame in meter units, which can be projected onto the image plane as a \textit{2D} image point feature with coordinates $\mathbf{s}=[u, v]^{T}$ expressed in pixel units. 
Let $\mathbf{v}_{c}=[\boldsymbol{v}_{c}^T, \boldsymbol{\omega}_{c}^T]^T \in \mathbb{R}^{6}$ be the spatial velocity of the camera in the world frame, composed of
the linear velocities $\boldsymbol{v}_{c}=\left[v_{x}, v_{y}, v_{z}\right]^T$ and the angular
velocities $\boldsymbol{\omega}_{c}=\left[\omega_{x}, \omega_{y}, \omega_{z}\right]^T$.
For the \textit{eye-in-hand} configuration, the time derivative of the coordinates of point $\boldsymbol{P}$ can be directly related to the camera velocity $\mathbf{v}_{c}$ by 
\begin{equation}
\dot{\boldsymbol{P}}=-\boldsymbol{v}_{c}-\boldsymbol{\omega}_{c} \times \boldsymbol{P} \Leftrightarrow\left\{\begin{array}{c}
\dot{X}=-v_{x}-\omega_{y} Z+\omega_{z} Y, \\
\dot{Y}=-v_{y}-\omega_{z} X+\omega_{x} Z, \\
\dot{Z}=-v_{z}-\omega_{x} Y+\omega_{y} X.
\end{array}\right.   
\end{equation}
This equation can be written in the matrix form as
\begin{equation}\label{eq:3dvs}
\underbrace{\begin{pmatrix}
\dot{X} \\
\dot{Y} \\
\dot{Z}
\end{pmatrix}}_{\dot{\boldsymbol{P}}}
=
\underbrace{\begin{pmatrix}
-1 & 0 & 0 & 0 & -Z & Y \\
0 & -1 & 0 & Z & 0 & -X \\
0 & 0 & -1 & -Y & X & 0
\end{pmatrix}}_{{}^{3d}\mathbf{L}_{\mathbf{s}}(X,Y,Z)}
\underbrace{\vphantom{\begin{pmatrix}
-1 & 0 & 0 & 0 & -Z & Y \\
0 & -1 & 0 & Z & 0 & -X \\
0 & 0 & -1 & -Y & X & 0
\end{pmatrix}}
\begin{pmatrix}
\boldsymbol{v}_{c} \\
\boldsymbol{\omega}_{c}
\end{pmatrix}}_{\mathbf{v}_{c}},
\end{equation}
where ${}^{3d}\mathbf{L}_{\mathrm{s}}(X,Y,Z) \in \mathbb{R}^{3 \times 6}$ refers to the \textit{interaction matrix} related to the \textit{3D} point $\boldsymbol{P}$ relative to the camera frame. 
Furthermore, according to the principle of a pinhole camera perspective projection, we can relate the normalized image-plane coordinates $(x=\frac{X}{Z},y=\frac{Y}{Z})$ to the pixel coordinates $(u,v)$ using the well-known equation
\begin{equation}\label{eq:uv}
    \left\{\begin{array}{c}
    u=f_u x+u_{0}, \\
    v= f_v y+v_{0},
\end{array}\right.
\end{equation}
where 
$\Gamma=(f_u, f_v, u_0, v_0)$
denotes the set of camera intrinsic parameters: $f_u = \frac{f}{\rho_{u}}$, $f_v = \frac{f}{\rho_{v}}$, $f$ is the camera focal length, $(\rho_{u}, \rho_{v})$ are the horizontal and vertical dimension of a pixel, and $(u_0, v_0)$ are the coordinates of the principal point.
Using~(\ref{eq:3dvs}) and~(\ref{eq:uv}), the relationship between the time variation of visual features $\mathbf{s}$ and the camera velocity screw $\mathbf{v}_{c}$ can be modeled as
\begin{equation}\label{eq:ibvs}
\dot{\mathbf{s}} = {}^{2d}\mathbf{L}_{\mathbf{s}}(u,v,Z,\Gamma) \mathbf{v}_{c},
\end{equation}
where the \textit{interaction matrix} (also called image Jacobian matrix) ${}^{2d}\mathbf{L}_{\mathbf{s}}(u,v,Z,\Gamma) \in \mathbb{R}^{2 \times 6}$ of a \textit{2D} point feature $\boldsymbol{s}$, with respect to the principal point, is given by
\begin{equation*}
{}^{2d}\mathbf{L}_{\mathbf{s}}=
\begin{pmatrix}
-\frac{f_{u}}{Z} & 0 & \frac{u}{Z} & \frac{u v}{f_{v}} & -f_{u}-\frac{u^{2}}{f_{u}} & \frac{f_{u}}{f_{v}} v \\
0 & -\frac{f_{v}}{Z} & \frac{v}{Z} & f_{v}+\frac{v^{2}}{f_{v}} & -\frac{u v}{f_{u}} & -\frac{f_{v}}{f_{u}} u
\end{pmatrix},
\end{equation*}
and $Z$ in the interaction matrix ${}^{2d}\mathbf{L}_{\mathbf{s}}$ denotes the depth of the feature point with respect to the camera frame. More details, concerning the derivations, can be found in \cite{corke2011robotics}.

The simplest and well-known control strategy to \textit{IBVS} is to merely use (\ref{eq:e(t)}) and (\ref{eq:ibvs}) to construct the control law that drives the current features toward their desired values on the image plane. Thus, this classical feedback control law is given by
\begin{equation}\label{eq:classical-ibvs}
    \mathbf{v}_{c}=-\lambda_{\mathbf{s}} {}^{2d}\widehat{\mathbf{L}_{\mathbf{s}}^{+}} \mathbf{e}, \quad \quad \forall \lambda_{\mathbf{s}} >0,
\end{equation}
where ${}^{2d}\widehat{\mathbf{L}_{\mathbf{s}}^{+}}$ denotes the pseudo-inverse of an estimation of ${}^{2d}\mathbf{L}_{\mathbf{s}}$, since it is impossible in practice to know exactly either ${}^{2d}\mathbf{L}_{\mathbf{s}}$ or ${}^{2d}\mathbf{L}_{\mathbf{s}}^{+}$.
In this work, the four well-known cases for constructing the approximate ${}^{2d}\widehat{\mathbf{L}}_{\mathbf{s}}$ are considered and studied, to be used in both classical \textit{IBVS} control strategy and our proposed one, which are as follows:
\begin{enumerate}
    \item \textit{CASE \#0}: ${}^{2d}\widehat{\mathbf{L}}_{\mathbf{s}} = {}^{2d}\mathbf{L}_{\mathbf{s}}(u_{t}, v_{t}, \hat{Z}_{t}, \hat{\Gamma})$;
    in the present case, we assume that each visual feature $\mathbf{s}=[u, v]^{T}$ and its depth $Z$ are estimated at each iteration of the control scheme under the assumption that the initial depth $Z(0)$ of each point-like feature is known.
    \item \textit{CASE \#1}: ${}^{2d}\widehat{\mathbf{L}}_{\mathbf{s}} = {}^{2d}\mathbf{L}_{\mathbf{s}}(u_{t}, v_{t}, \hat{Z}^{*}, \hat{\Gamma})$;
    herein, it is supposed that the feature points $\mathbf{s}$ must be updated at each iteration, whereas the depth of each point is considered to be fixed and is set to the estimated value of $Z$ at the desired camera position, i.e., $\hat{Z}=\hat{Z}^{*}$.
    \item \textit{CASE \#2}: ${}^{2d}\widehat{\mathbf{L}}_{\mathbf{s}} = {}^{2d}\mathbf{L}_{\mathbf{s}}(u^{\ast}, v^{\ast}, \hat{Z}^{\ast}, \hat{\Gamma})$; the simplest case is to consider the interaction matrix as a constant matrix with not only $\hat{Z}=\hat{Z}^{*}$ but with also $\mathbf{s}=\mathbf{s}^{*}$. This means that no estimation will be carried out, at each iteration, during the visual servoing task, and only the desired depth of each point needs to be set.
    \item \textit{CASE \#3}: Finally, we considered the choice that has been proposed in \cite{malis2004improving}, which relies on the use of the mean of the estimation of the interaction matrix ${}^{2d}\widehat{\mathbf{L}}_{\mathbf{s}}$ at the current iteration (namely, \textit{CASE \#0}) and at equilibrium (namely, \textit{CASE \#2}); i.e., in this case,  ${}^{2d}\widehat{\mathbf{L}}_{\mathbf{s}} = \frac{1}{2}\left(
    {}^{2d}\mathbf{L}_{\mathbf{s}}(u_{t}, v_{t}, \hat{Z}_{t}, \hat{\Gamma}) 
    +
    {}^{2d}\mathbf{L}_{\mathbf{s}}(u^{\ast}, v^{\ast}, \hat{Z}^{\ast}, \hat{\Gamma})\right)$.
\end{enumerate}

In a similar way, the classical control law relative to a \textit{3D} point feature $\boldsymbol{P}$ can be computed using (\ref{eq:e(t)}) and (\ref{eq:3dvs}) as
\begin{equation}\label{eq:classical-3dvs}
    \mathbf{v}_{c}=-\lambda_{\mathbf{s}} {}^{3d}\widehat{\mathbf{L}_{\mathbf{s}}^{+}} \mathbf{e}, \quad \quad \forall  \lambda_{\mathbf{s}} >0,
\end{equation}
where $\mathbf{s}$, in this case, consists of a set of \textit{3D} point features expressed in the camera frame, i.e., $\mathbf{s}\equiv \boldsymbol{P}$. Herein, we called this intermediate approach a \textit{3DVS} control strategy.

\subsection{Position-Based Visual Servoing (PBVS)}
In the \textit{PBVS} control strategy, the visual features extracted from the image are utilized for estimating the pose of the camera relative to either (i) a reference frame $\mathcal{F}_{o}$ tied to the object, or (ii) the desired camera frame $\mathcal{F}_{c^{*}}$. 
Thus, $\mathbf{s}$ can be generally defined as  $\mathbf{s}= \left[\mathbf{t}, \theta \mathbf{u}\right]^{T} \in \mathbb{R}^{6}$, where $\mathbf{t} = \left[t_{x}, t_{y}, t_{z}\right]$ denotes a translation vector; whereas $\theta \mathbf{u}$ represents a rotation vector, where $\mathbf{u} = \left[u_{x}, u_{y}, u_{z}\right]$
is a unit vector representing the rotation axis and $\theta$ is the rotation angle. 
Within this work, we define $\mathbf{t}$ as the translation vector between the current camera frame $\mathcal{F}_{c}$ and its desired frame $\mathcal{F}_{c^{*}}$.
Therefore, the aim of the control scheme is to drive the camera so that $\mathcal{F}_{c}$ converges to $\mathcal{F}_{c^{*}}$.
As a result, we have $\mathbf{s}=\left[{}^{c^*}\mathbf{t}_{c}, \theta \mathbf{u}\right]^T$, $\mathbf{s}^{\ast}=0$, and $\mathbf{e}=\mathbf{s}$. 
Furthermore, by following the developments presented in \cite{chaumette2006visual}, the interaction matrix relative to $\mathbf{s}$ can be obtained by 
\begin{equation}
{}^{\text{pbvs}}\mathbf{L}_{\mathbf{s}}({}^{c^*}\mathbf{t}_{c}, \theta \mathbf{u})=
\begin{pmatrix}
{}^{c^*}\mathbf{R}_{c} & \mathbf{0}_3 \\
\mathbf{0}_3 & \mathbf{L}_{\theta \mathbf{u}}
\end{pmatrix},
\; {}^{\text{pbvs}}\mathbf{L}_{\mathbf{s}} \in \mathbb{R}^{6\times 6},
\end{equation}
where ${}^{c^*}\mathbf{R}_{c} \in \mathbb{R}^{3\times 3}$ refers to the rotation matrix between frames $\mathcal{F}_{c}$ and $\mathcal{F}_{c^{*}}$, and the Jacobian matrix $\mathbf{L}_{\theta \mathbf{u}}$ is given by 
\begin{equation}
\mathbf{L}_{\theta \mathbf{u}}=\mathbf{I}_{3}-\frac{\theta}{2}[\mathbf{u}]_{\times}+\left(1-\frac{\sinc(\theta)}{\sinc^{2}( \frac{\theta}{2})}\right)[\mathbf{u}]_{\times}^{2}, \mathbf{L}_{\theta \mathbf{u}} \in \mathbb{R}^{3\times 3},
\end{equation}
in which $\mathbf{I}_{3}$ denotes a $3 \times 3$ identity matrix, $[\mathbf{u}]_{\times}$ is the skew-symmetric matrix associated with
vector $\mathbf{u}$, and $\sinc(\theta)=\sin(\theta)/\theta$. 
Similar to \textit{IBVS} scheme, the time derivative of $\mathbf{s}$ can be related to the camera velocity screw $\mathbf{v}_{c}$ by 
\begin{equation}\label{eq:pbvs}
\dot{\mathbf{s}} = {}^{\text{pbvs}}\mathbf{L}_{\mathbf{s}} \mathbf{v}_{c}.
\end{equation}
Accordingly, we can obtain the simplest, i.e., classical, \textit{PBVS} control strategy by
\begin{equation}\label{eq:classical-pbvs}
    \mathbf{v}_{c}=-\lambda_{\mathbf{s}} {}^{\text{pbvs}}\widehat{\mathbf{L}_{\mathbf{s}}^{-1}} \mathbf{e}, \quad  \quad \forall \lambda_{\mathbf{s}} >0,
\end{equation}
which can be decoupled into: $\boldsymbol{v}_{c} =-\lambda_{\mathbf{s}} {}^{c^*}\mathbf{R}_{c}^{T} {}^{c^*}\mathbf{t}_{c}$, 
and $\boldsymbol{\omega}_{c} =-\lambda_{\mathbf{s}} \theta \mathbf{u}$, where $\mathbf{L}_{\theta \mathbf{u}}^{-1} \theta \mathbf{u}=\theta \mathbf{u}$.



\section{MPPI Control Strategy for Visual Servoing}\label{MPPI Control Strategy}
In this section, we briefly present the control strategy of our proposed sampling-based \textit{MPC} approach (namely, \textit{MPPI}) for visual servoing systems; then, we state the mathematical formulation of \textit{MPPI} in the presence of constraints such as the visibility, three-dimensional (i.e., \textit{3D}), and control constraints. 

\subsection{Review of MPPI}
The \textit{MPPI} control strategy is a sampling-based and derivative-free optimization method to model predictive control (\textit{MPC}) that can be easily applied in \textit{real-time} (i.e., \textit{online}) to the real system, without requiring the first- or second-order approximation of the system dynamics and quadratic approximation of the objective functions. 
\begin{wrapfigure}{r}{0.21\textwidth}
\begin{center}
\includegraphics[scale=1]{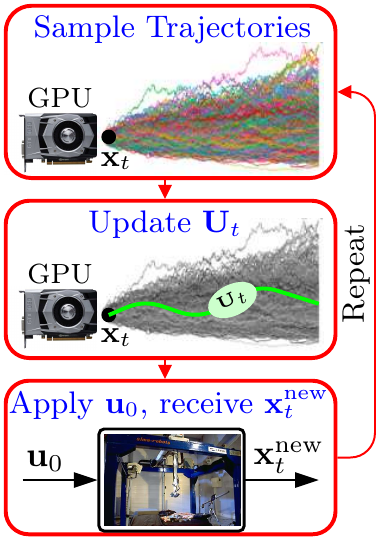}
\caption{Real-time \textit{MPPI} control loop \cite{williams2017model}.} 
\label{fig:mppi_concept}
\end{center}
\end{wrapfigure}
The \textit{real-time} control cycle of \textit{MPPI} is shown in Fig.~\ref{fig:mppi_concept}. 
At each time-step $\Delta t$, \textit{MPPI} samples thousands of trajectories from the system dynamics (e.g., in our case, from Eq. (\ref{eq:3dvs}), (\ref{eq:ibvs}), or (\ref{eq:pbvs})), using a Graphics Processing Unit (GPU) so as to ensure a \textit{real-time} implementation. 
Afterward, on the basis of the parallel nature of sampling, each of these trajectories is individually executed and then evaluated according to its expected cost. 
In sequential, the optimal control sequence $\mathbf{U}$, over a finite prediction time-horizon $t_p \in\{0,1,2, \dots, T-1\}$, is updated based on a weighted average cost over these generated trajectories, where $\mathbf{U} = \left(\mathbf{u}_{0}, \mathbf{u}_{1}, \dots,\mathbf{u}_{T-1}\right) \in \mathbb{R}^{m \times T}$, and $T \in \mathbb{R}^{+}$ refers to the number of timesteps. 
Finally, the first control $\mathbf{u}_{0}$ is applied to the system, while the remaining control sequence of length $T-1$ is slid-down to be used for providing a warm-starting to the optimization at the next time-step.

Let $\delta \mathbf{u}_{t} \in \mathbb{R}^{m}$ be a zero-mean Gaussian noise vector with a variance of $\Sigma_{\mathbf{u}}$, i.e., $\delta \mathbf{u}_{t} \sim \mathcal{N}(\mathbf{0}, \Sigma_{\mathbf{u}})$, where $\delta \mathbf{u}_{t}$ represents the random noise associated with the commanded control input $\mathbf{u}_{t}$ to the system; in other words, it represents the control input updates as the actual input is $\mathbf{v}_{t} = \mathbf{u}_{t}+\delta \mathbf{u}_{t}$. 
Suppose that the number of the samples (namely, trajectories or rollouts) drawn from the discrete-time dynamics system, $\mathbf{x}_{t+1}=f\left(\mathbf{x}_{t},\mathbf{v}_{t}\right)$, is $K$, where $\mathbf{x}_{t} \in \mathbb{R}^{n}$ denotes the state of the system at time $t$. Moreover, let $\tilde{S}\left(\tau_{t, k}\right) \in \mathbb{R}^{+}$ be the \textit{cost-to-go} of the $k^{th}$ trajectory from time $t$ onward. Then, based on the detailed derivation given in \cite{williams2017model} and well-summarized in \cite{mohamed2020model}, the optimal control sequence $\left\{\mathbf{u}_{t}\right\}_{t=0}^{T-1}$ can be readily updated using the following iterative update law: 
\begin{equation}\label{eq:mppi_optimal-control}
 \mathbf{u}_{t} \leftarrow \mathbf{u}_{t} +\frac{\sum_{k=1}^{K} \exp \left(-(1 / \lambda) \tilde{S}\left(\tau_{t, k}\right)\right) \delta \mathbf{u}_{t, k}}{\sum_{k=1}^{K} \exp \left(-(1 / \lambda) \tilde{S}\left(\tau_{t, k}\right)\right)}, 
 \end{equation}
where $\lambda \in \mathbb{R}^{+}$ is so-called the inverse temperature which determines the level of the selectiveness of the weighted average. 
We defined the \textit{cost-to-go} of each trajectory $\tau$ over the predefined prediction time-horizon as
\begin{equation}
 \tilde{S}\left(\tau\right) =\phi\left(\mathbf{x}_T\right)+\sum_{t=0}^{T-1} \tilde{q}\left(\mathbf{x}_{t}, \mathbf{u}_{t}, \delta \mathbf{u}_{t}\right),  
\end{equation}
in which $\phi\left(\mathbf{x}_T\right)$ refers to the terminal cost. Whilst $\tilde{q}\left(\mathbf{x}_t, \mathbf{u}_t, \delta \mathbf{u}_t\right)$ denotes the instantaneous running cost, which composed of the sum of state-dependent running cost $q\left(\mathbf{x}_{t}\right)$ and quadratic control cost, and is defined as follows: 
\begin{equation*}
\tilde{q} 
=
\underbrace{
\vphantom{
\frac{\left(1-\nu^{-1}\right)}{2} \delta \mathbf{u}_{t}^{T} R \delta \mathbf{u}_{t}+\mathbf{u}_{t}^{T} R \delta \mathbf{u}_{t}+\frac{1}{2} \mathbf{u}_{t}^{T} R \mathbf{u}_{t}}
q\left(\mathbf{x}_{t}\right)}_{\text{\color{red}{\textit{State-dep.}}}} 
+
\underbrace{\frac{\left(1-\nu^{-1}\right)}{2} \delta \mathbf{u}_{t}^{T} R \delta \mathbf{u}_{t}+\mathbf{u}_{t}^{T} R \delta \mathbf{u}_{t}+\frac{1}{2} \mathbf{u}_{t}^{T} R \mathbf{u}_{t}}_{\text{\color{red}{\textit{Quadratic Control Cost}}}},
\end{equation*}
where $R\in \mathbb{R}^{m \times m}$ is a positive definite control weight matrix, and $\nu \in \mathbb{R}^{+}$ is so-called the exploration noise which determines how aggressively \textit{MPPI} explores the state-space; the impact of changing $\nu$ has been studied in \cite{mohamed2020model}.

\subsection{MPPI-VS Control Strategy}\label{sec:MPPI-VS Control Strategy}
In order to apply the \textit{MPPI} control strategy to the visual servoing systems, the non-linear discrete-time form of the continuous-time dynamical model given in (\ref{eq:3dvs}), (\ref{eq:ibvs}), or (\ref{eq:pbvs}) is required, for simulating and propagating thousands of trajectories in parallel. 
This discrete-time model can be approximated using the Newton–Euler method as
\begin{equation}\label{eq:DT-model}
    \mathbf{s}(t+1) = \mathbf{s}(t)+ \widehat{\mathbf{L}}_{\mathbf{s}}(t) \mathbf{v}_{c}(t) \Delta t,
\end{equation}
whereby $\mathbf{s}(t)$ defines the state vector of the model (i.e., the \textit{MPPI} states) at time instant $t$, where 
(i) for \textit{MPPI-IBVS} control scheme, the state vector $\mathbf{s}$ refers to a set of $n_p$ \textit{2D} points, i.e. $\mathbf{s} = [u_i, v_i]^{T} \in \mathbb{R}^{2n_p} \, \forall i=1,2,\dots, n_p$; 
(ii) for \textit{MPPI-3DVS} control scheme, it is composed of a set of $n_p$ \textit{3D} point features, i.e., $\mathbf{s} = [X_i, Y_i, Z_i]^{T} \in \mathbb{R}^{3n_p}$; 
while (iii) for \textit{MPPI-PBVS} control scheme, the state vector is defined as $\mathbf{s}=\left[{}^{c^*}\mathbf{t}_{c}, \theta \mathbf{u}\right]^T \in \mathbb{R}^{6}$, as described 
in Section~\ref{VS-model}. 
Similarly, the approximate interaction matrix is given by either $\widehat{\mathbf{L}}_{\mathbf{s}} = {}^{2d}\widehat{\mathbf{L}}_{\mathbf{s}}(u_i,v_i,Z_i,\Gamma)\in \mathbb{R}^{2n_p \times 6}, {}^{3d}\widehat{\mathbf{L}}_{\mathbf{s}}(X_i,Y_i,Z_i)\in \mathbb{R}^{3n_p \times 6},$ or $ {}^{\text{pbvs}}\widehat{\mathbf{L}}_{\mathbf{s}}({}^{c^*}\mathbf{t}_{c}, \theta \mathbf{u})\in \mathbb{R}^{6 \times 6}$ for \textit{MPPI-$\{$IBVS, 3DVS, PBVS\}} control schemes, respectively.
It is generally apparent that our \textit{MPPI-VS} control strategy directly utilizes the estimation of the interaction matrix, without the need for estimating the interaction matrix inversion or performing the pseudo-inversion in real-time.
Furthermore, it is noteworthy that for the \textit{MPPI-IBVS} control scheme, particularly for \textit{CASE \#0} and \#3 of the approximate ${}^{2d}\widehat{\mathbf{L}}_{\mathbf{s}}$ where the depth $Z_i$ of each point-like feature needs to be estimated, we use the discrete-time model of the \textit{3D} point features given in (\ref{eq:DT-model}) for predicting the future evolution of the object depths $Z_i(t)$, assuming that, at each time-step $\Delta t$, the initial depth $Z_i(0)$ of a set of $n_p$ point-like features is known.

\subsubsection{\textbf{Handling Visibility and \textit{3D} Constraints}}
One of the most attractive features of \textit{MPPI}, compared to the classical \textit{MPC}, is its capability of coping easily with the hard and soft constraints, without adding additional complexity to the optimization problem. 
More precisely, a \textit{large-weighted} indicator function can be employed as part of the state-dependent running cost function $q(\mathbf{x})$ for handling the constraints. In our case, for \textit{MPPI-VS} control scheme, the instantaneous state-dependent running cost function, $q(\mathbf{x})\equiv q(\mathbf{s})$, is defined as
\begin{equation}\label{eq:running-cost}
q(\mathbf{s}_t)= q_1(\mathbf{s}_t) + q_2(\mathbf{s}_t),  
\end{equation}
where $q_1(\mathbf{s}_t)=(\mathbf{s}_t-\mathbf{s}^{\ast})^{\mathrm{T}} Q (\mathbf{s}_t - \mathbf{s}^{\ast})$ denotes the state-dependent cost which is a simple quadratic cost for enforcing the current state $\mathbf{s}_t$ to reach its desired value $\mathbf{s}^{\ast}$. 

On the other hand, $q_2(\mathbf{s}_t)$ refer to an indicator function utilized for handling:
\begin{enumerate}[label=(\roman*)]
\item the visibility constraints to ensure that the visual features always remain within the image plane; they can be simply expressed as $\mathbf{s}_{\min} \leq \mathbf{s}_t \leq \mathbf{s}_{\max}$, where $\mathbf{s}_{\min} = [u_{\min},v_{\min}]^T$ and $\mathbf{s}_{\max} = [u_{\max},v_{\max}]^T$ refer to the lower and upper ranges, in pixels, of the image point coordinates,
and
\item the three-dimensional (i.e., \textit{3D}) constraints such as workspace or joint limits; 
in the present work, we did not consider the joint limits as a part of the optimization problem, in order to better understand the behavior of \textit{MPPI} whenever one of the joints reaches its given bounds. 
In practice, the joint limits are handled by the robot controller as illustrated in Fig.~\ref{fig:VS-framework}; thus, if the robot reaches the joint bounds, the velocity of each joint will be set to zero.
\end{enumerate}
The definition of $q_2(\mathbf{s})$ is mainly based on the \textit{MPPI-VS} control scheme to be implemented. Therefore, in the \textit{MPPI-IBVS} control scheme, $q_2$ is formulated as 
    \begin{equation}
        q_2 = \num{d7}C_1+ \num{d5}C_2,  
    \end{equation}
where 
$C_{1}$ and $C_{2}$ are Boolean variables that are used to heavily penalize each trajectory that violates the visibility and \textit{3D} constraints, respectively. For instance, $C_1$ is turned on (i.e., $C_1=1$) if, at any time, $\mathbf{s}$ exceeds its given bounds, i.e., $C_{1} = 1 \Leftrightarrow \exists \mathbf{s}| \; \big((\mathbf{s}> \mathbf{s}_{\max}) \lor (\mathbf{s} < \mathbf{s}_{\min})\big)$.
Since the joint limits have not been considered as a part of the \textit{MPPI} optimization problem, we can add the second term, 
if needed, for constraining the future evolution of the \textit{3D} point features, namely, $C_{2} = 1 \Leftrightarrow \exists \boldsymbol{P}_i | \; \big((\boldsymbol{P}_i > \boldsymbol{P}_{\max}) \lor (\boldsymbol{P}_i < \boldsymbol{P}_{\min})\big) \, \forall i=1,\dots, n_p$. 
Just to name a few, this term can be added for avoiding the camera retreat motion, which occurs when there is a large rotation between the initial and desired camera configuration (this phenomenon is known as the \textit{camera retreat} problem), by simply penalizing the $Z$-axis translation motion of each \textit{3D} point feature $\boldsymbol{P}$ relative to the camera frame. In the present case, $C_2$ will be active if at any time the depth of one point, at least, exceeds a threshold $Z_{\max}$, where $Z_{\max}$ defines the maximum allowable camera retreat motion along its optical axis, i.e., $C_{2} = 1 \Leftrightarrow \exists Z_i| \;(Z_i > Z_{\max})$.

In the \textit{MPPI-PBVS} control scheme, 
the controller might produce control input that would ultimately leads to that some visual features may leave the camera's FoV, just as in the classical \textit{PBVS} control strategy expressed in (\ref{eq:classical-pbvs}). 
Sequentially, the pose of the object can not be estimated and, then, the robot will be stopped as the control loop is no longer closed. 
This is mainly due to the fact that the control law is explicitly expressed in Cartesian space and there is apparently no direct control to the visual features on the image plane.
Therefore, we propose two methods so as to guarantee that the object always remains within the camera's FoV.
Concerning the first proposed method, the indicator function $q_2$ is defined as a \textit{large-weighted} exponential penalty function, as follows:
\begin{equation}\label{eq:17}
q_2=\beta \sum_{i=1}^{n_p} \bigg(
e^{-\alpha \big(x_{\max} - \mid x_i\mid \big)}
+ e^{-\alpha \big(y_{\max} - \mid y_i\mid \big)}
\bigg),
\end{equation}
so that the visibility constraints are constantly satisfied which can be written as
$ |p_i| \leq p_{\text{max}} \, \forall i = 1,\dots, n_p$, where $p_i = [x_i, y_i]^{T} = [\frac{X_i}{Z_i},\frac{Y_i}{Z_i}]^{T} \in \mathbb{R}^{2}$ denotes the normalized Cartesian coordinates of an image point, $p_{\max}= [x_{\max}, y_{\max}]^{T}$, $p_{\min}= -p_{\max}$ are the minimum and maximum bounds, in meters, of the point feature projected on the normalized image plane, while $\beta$ and $\alpha$ are positive scalar variables.
In the second method, we proposed an alternative solution based on augmenting the \textit{3D} point features within the nominal state vector. In further words, we combine both \textit{MPPI-PBVS} and \textit{MPPI-3DVS} control schemes. Thus, in this case, we have $\mathbf{s}=\left[{}^{\text{pbvs}}\mathbf{s}, {}^{3d}\mathbf{s} \right]^T= \left[{}^{c^*}\mathbf{t}_{c}, \theta \mathbf{u}, X_i, Y_i, Z_i\right]^T \in \mathbb{R}^{(6+3n_p)}$, whilst   
$\widehat{\mathbf{L}}_{\mathbf{s}} = \left[{}^{\text{pbvs}}\widehat{\mathbf{L}}_{\mathbf{s}}; 
{}^{3d}\widehat{\mathbf{L}}_{\mathbf{s}}
\right] \in \mathbb{R}^{(6+3n_p) \times 6}$. 
In the present scheme, the penalty (i.e., indicator) function $q_2(\mathbf{s})$ is replaced by a \textit{large-weighted} quadratic state-dependent cost function which modulates how fast the current state vector of the \textit{3D} point features ${}^{3d}\mathbf{s}$ converges to its desired one ${}^{3d}\mathbf{s}^{*}$. Therefore, the running cost function $q(\mathbf{s})$, which is given in (\ref{eq:running-cost}), is reformulated as 
\begin{equation}\label{eq:18}
q(\mathbf{s})
=
\underbrace{\!\!\!\!
w_1\sum_{i=0}^{5} (\mathbf{s}_i-\mathbf{s}^{\ast}_i)^2
}_{\color{red}{:=q_1(\mathbf{s}), \, \mathbf{s}_i\in\, {}^{\text{pbvs}}\mathbf{s}, \, \mathbf{s}^{\ast}_i\in\, {}^{\text{pbvs}}\mathbf{s}^{\ast}}}\!\!\!
+\!
\underbrace{w_2\sum_{i=6}^{n-1} (\mathbf{s}_i-\mathbf{s}^{\ast}_i)^2
}_{\color{red}{:=q_2(\mathbf{s}), \,\mathbf{s}_i\in\, {}^{\text{3d}}\mathbf{s},\,\mathbf{s}^{\ast}_i\in \, {}^{\text{3d}}\mathbf{s}^{\ast}}}\!\!,
\end{equation}
where $\mathbf{s}^{\ast}=\left[{}^{\text{pbvs}}\mathbf{s}^{\ast}, {}^{3d}\mathbf{s}^{\ast} \right]^T$, $n = 3n_p+6$ is the length of the state vector $\mathbf{s}$, whereas $w_1, w_2 \in \mathbb{R}^{+}$ refer to the cost weighting of $q_1$ and $q_2$, respectively.
It should be noted that assigning very high weights, particularly for $w_2$, is sufficient for enforcing the visual features to stay within the image plane, without imposing additional constraints on the optimization problem.

\subsubsection{\textbf{Handling Control Constraints}}
Most robotic systems, including the 6-DoF Cartesian robot we consider here, have constraints on the actuators, such as range or velocity limitations, that must be taken into consideration in the control law design. These constraints are known as control constraints which considered to be hard constraints that should not be violated. In the case of \textit{MPPI} control strategy, there exist two ways for handling the control constraints.
First, they can be implemented as a natural part of the running cost by adding an appropriate term (e.g., indicator function), as we previously explained, to penalize all the trajectories that violate the constraints.
However, the common issue associated with this way is that the control constraints acting as soft constraints, and, accordingly, it is notoriously difficult to ensure that the control input, obtained by the controller, remains always within its allowed bounds, even after rejecting each trajectory that violates the control input limits.
Therefore, within this work, we utilize an alternative method based on pushing the control constraints into the dynamics system \cite{williams2017information}, meaning that $\mathbf{x}_{t+1}=f\big(\mathbf{x}_{t},g(\mathbf{v}_{t})\big)$, where $g(\mathbf{v}_{t})$ is an element-wise clamping function that is used to restrict the control input $\mathbf{v}_{t}\sim \mathcal{N}(\mathbf{u}_t, \Sigma_{\mathbf{u}})$ to remain within a given range, for all samples drawn from the dynamics system. Thus, $g(\mathbf{v})$ can be defined as $g(\mathbf{v})= \max \big(\mathbf{v}_{\min}, \min(\mathbf{v}, \mathbf{v}_{\max})\big)$, where $\mathbf{v}_{\min}$ and $\mathbf{v}_{\max}$ are
the lower and upper bounds of the control input.
The major advantage, here, is that the problem formulation of \textit{MPPI}, which takes into account control constraints, is converted into an unconstrained one, without violating the constraints and affecting the convergence of the \textit{MPPI} algorithm as $g(\mathbf{v})$ has an impact only on the dynamics system.

\section{Simulation-Based Evaluation}\label{Simulation Details and Results} 
In the present section, we conduct extensive simulations on a simulated 6-DOF Cartesian robot with an \textit{eye-in-hand} camera configuration so as to evaluate and demonstrate the potential advantages of our proposed control strategy for improving the quality of visual servoing tasks, along with a comparison to the classical schemes (namely, \textit{C-IBVS}, \textit{C-3DVS}, and \textit{C-PBVS} schemes).
\subsection{Simulation Settings and Performance Metrics}

We validated the \textit{MPPI-VS} control strategy as well as the classical schemes on a 6-DOF Cartesian robot (namely, Gantry robot) based on the utilization of four points (i.e., $n_p=4$) in the image plane as visual features.\footnote{The reason behind choosing the Cartesian robot is that it has a quite large workspace that helps to verify the robustness of the \textit{MPPI-VS} control scheme. For more information about the real robot simulator: \url{https://visp.inria.fr/robot-interface/} \& \url{https://github.com/lagadic/visp\_ros/blob/master/nodes/afma6.cpp}}
During all the simulations, it is assumed that the \textit{MPPI} algorithm runs with a time horizon $t_p$ of \SI{3.5}{\second}, a control frequency of \SI{50}{\hertz} (i.e., $T=175$), and generates \num{2000} samples each time-step $\Delta t$ with an exploration variance $\nu$ of \num{1000}. Moreover, the control weighting matrix $R$ is set to $\frac{1}{2}\lambda \Sigma_{\mathbf{u}}^{-1}$, assuming that the random noise associated with the control input has a variance of $\Sigma_{\mathbf{u}} = \operatorname{Diag}\left(0.02, 0.01, 0.01, 0.02, 0.02, 0.01\right)$. 
On the other side, the remaining parameters, which needed to be adjusted on the basis of the \textit{MPPI-VS} scheme to be implemented, are tabulated in Table~\ref{table:SysParameters}, as well as the camera intrinsic parameters. 
More concretely, in the case of \textit{MPPI-IBVS} scheme, it can be noticeable that the inverse temperature $\lambda$ is set to a higher value. Conversely, it is tuned to much lower values in both \textit{MPPI-3DVS} and \textit{MPPI-PBVS} control schemes. 
It is important to bear in mind that fine-tuning $\lambda$ is mainly based on the state-dependent running cost function $q\left(\mathbf{s}\right)$ and in which plane (i.e., \textit{2D} image plane or Cartesian plane) it is expressed. 
The real-time \textit{MPPI-VS} algorithm is executed on an NVIDIA GeForce GTX 1080 Ti desktop GPU. In addition, all the  control schemes, including the classical ones, were developed in Python and C++ and were implemented using Visual Servoing Platform (ViSP) \cite{marchand2005visp} integrated with the Robot Operating System (ROS) framework.
\definecolor{applegreen}{rgb}{0.7, 1, 0.0}
\definecolor{LightCyan}{rgb}{0.88,1,1}
\definecolor{atomictangerine}{rgb}{1.0, 0.6, 0.4}
\begin{table}[!ht]
\caption{Camera and \textit{MPPI-VS} Parameters}
\centering
\begin{tabular}{|c| c||c| c|}
\hline
\rowcolor{applegreen}
 Parameter  & Value  & Parameter     & Value\\
 \hline  
 \hline
 \rowcolor{LightCyan}
 \multicolumn{4}{|c|}{Camera Intrinsic Parameters} \\
 \hline
 Image Res. [\si{\pixels}]  & \si{ 480 \times 640}  & $f$ [\si{\milli\metre}]   & \num{8.40}\\
 \hline
 $u_0$  [\si{\pixels}]   & \num{320}  & $f_u$ [\si{\pixels}] & \num{840}\\
 \hline
 $v_0$  [\si{\pixels}]   & \num{240}  & $f_v$ [\si{\pixels}]  & \num{840}\\
 \hline \hline
 \rowcolor{LightCyan}
 \multicolumn{4}{|c|}{Parameters of \textit{MPPI-IBVS} Control Scheme} \\
 \hline  
$\lambda$  & \num{100} & $Q$ & $2.5\mathbf{I}_{8}$\\
\hline \hline
 \rowcolor{LightCyan}
 \multicolumn{4}{|c|}{Parameters of \textit{MPPI-3DVS} Control Scheme} \\
 \hline  
$\lambda$  & \num{d-2} & $Q$ & $35\mathbf{I}_{12}$\\
 \hline \hline
 \rowcolor{LightCyan}
 \multicolumn{4}{|c|}{Parameters of \textit{MPPI-PBVS} Control Scheme} \\
\hline
$\lambda$  & \num{d-3} & $Q$ & $35\mathbf{I}_{6}$\\
\hline
$\beta, \alpha$ \quad (\ref{eq:17}) & \num{150}, \num{d3} & $w_1, w_2$ \quad (\ref{eq:18}) & \num{35}, \num{150}\\
\hline
\end{tabular}
\label{table:SysParameters}
\end{table}

\subsubsection{\textbf{Initial Camera Configurations}} To better assess and demonstrate the robustness level of our proposed control strategy, \num{120} initial camera configurations were randomly extracted from a uniform distribution within the robot's workspace, with a guarantee that (i) the robot kinematics has initially the capability of reaching these generated poses, and (ii) the visual features are initially located within the camera's FoV.
The four point-like features were located in the $xy$-plane at the positions $P_{1}=[-0.1,-0.1,0]^{T}\!, P_{2}=[0.1,-0.1,0]^{T}$\!, $P_{3}=\![0.1,0.1,0]^{T}$\!, and $P_{4}=[-0.1,0.1,0]^{T}$ [\si{\metre}] (see Fig.~\ref{fig:3Dtrajectory_S113_ibvs}). 
\subsubsection{\textbf{Desired Camera Configurations}}
In this work, we considered two different desired poses of the camera, all expressed in the reference frame $\mathcal{F}_{o}$ attached to the object with $z$-axis pointing downward.
The first desired camera pose ${ }^{o}\boldsymbol{P}_{c_1}^{\ast} = [{ }^{o}\boldsymbol{t}_{c}, \theta \mathbf{u}]^T$ was chosen along the $-z$-axis with a translation vector ${ }^{o}\boldsymbol{t}_{c} = \left[0,0, -0.75\right]$ [\si{\metre}], heading toward the features with an orientation vector $\boldsymbol{\theta \mathbf{u}} = \left[0,0, 0\right]$ [\si{\deg}].
Hence, the desired coordinates of the four points in the image plane are $\mathbf{s}^{\ast} = [\mathbf{s}_1^{\ast}, \mathbf{s}_2^{\ast}, \mathbf{s}_3^{\ast},\mathbf{s}_4^{\ast}]^T$, where  $\mathbf{s}_1^{\ast} =[128,208]$, $\mathbf{s}_2^{\ast}=[352,208]$, $\mathbf{s}_3^{\ast}=[352,432]$, and $\mathbf{s}_4^{\ast}=[128,432]$ [\si{\pixels}]; meaning that the object lies in the center of the image, as illustrated in Fig.~\ref{fig:s113_ibvs}.
We have defined the second camera configuration in such a way that the features are much closer to the edge of the camera's FoV (i.e., the object is located near one of the image borders), to demonstrate how robust the proposed control strategy is against violating the system constraints particularly visibility constraints.
Herein, the camera pose relative to the object frame ${ }^{o} \boldsymbol{P}_{c_2}^{\ast}$ is set to $\left(\left[0.076, 0.202, -0.727\right], \left[10, -10, -15\right]\right)^{T}$ in ([\si{\metre}], [\si{\deg}]). As a result, the object lies in the top-right corner of the image.

\subsubsection{\textbf{Performance Metrics}}
For each initial camera configuration, we considered the following. First, for the conventional VS control schemes, the simulation is run for \SI{60}{\second} on a realistic simulator of the real robot. Whereas for our proposed \textit{MPPI-VS} scheme, we adopted the simulation time to be \SI{90}{\second} as it is noticed that the convergence rate is empirically slow, especially in the case of \textit{MPPI-IBVS} scheme, as will be discussed later. 
Second, the visual servoing task is considered to be successful if the following are satisfied: 
\begin{enumerate}[label=(\roman*), leftmargin=0.6cm]
\item The camera does not reach a local minimum (which indicated as $\mathcal{R}_{\text{LM}}= 0$ or $\textit{False}$, during the simulations). Such configuration corresponding to a local minimum occurs when $\mathbf{v}_c=0$ and
${ }^{o} \boldsymbol{P}_{c}\neq { }^{o} \boldsymbol{P}_{c}^{\ast}$. 
Mathematically, we utilized the mean-squared error (\textit{MSE}) as a metric that measures the positioning error $\epsilon$, which can be formulated as $\textit{MSE}=\frac{1}{n} \mathbf{e}^T\mathbf{e}$ in which $\mathbf{e} = { }^{o} \boldsymbol{P}_{c} - { }^{o} \boldsymbol{P}_{c}^{\ast}$ and $n=6$. 
Thence, we considered that the local minimum is completely avoided (i.e., $\mathcal{R}_{\text{LM}}= 0$) if and only if $\textit{MSE}_1 = \frac{2}{n} \mathbf{e}_1^T\mathbf{e}_1 <\epsilon_1$ and $\textit{MSE}_2 = \frac{2}{n} \mathbf{e}_2^T\mathbf{e}_2 <\epsilon_2$, where $\mathbf{e}_1$ (in [\si{\metre}]) and $\mathbf{e}_2$ (in [\si{\radian}]) refer to the translational and rotational errors between ${ }^{o} \boldsymbol{P}_{c}$ and ${ }^{o} \boldsymbol{P}_{c}^{\ast}$, respectively. During the simulations, we set the two thresholds as $\epsilon_1= \num{d-5}$ and $\epsilon_2 = \num{d-4}$.
\item The robot joint limits are avoided (i.e., $\mathcal{R}_{\text{JL}}=0$). If the robot reaches one of the joint limits, the velocity of each joint will be set to zero by the robot controller (for safety reasons) and, accordingly, $\mathcal{R}_{\text{JL}}$ will be turned to 1. 
\item The visual features (in our case, the four points) always remain within the camera's FoV (i.e., $\mathcal{P}_{\text{out}}=0$). 
Furthermore, in the experimental validation, it has been assumed that the robot will immediately stop as soon as one of the features is no longer visible in the image.
\end{enumerate}
Finally, the number of successful servoing tasks, with respect to the total initial camera configurations, is denoted as $\mathcal{N}_{\text{success}}$, while $\mathcal{S}_{\text{rate}}$ indicates the success rate.

\subsection{Intensive Simulation Details and Results}
In order to validate the theoretical findings and emphasize the efficiency of the \textit{MPPI-VS} control strategy, intensive simulation studies have been performed under various operating conditions, with the aim of (i) highlighting the potential advantages of our proposed strategy for improving the performance of visual servoing in terms of visual features prediction and constraints handling,
(ii) fine-tuning the parameters of the \textit{MPPI} algorithm and, then, studying the impact of changing these parameters as they play a critical role in determining the robustness and ensuring the convergence of the algorithm,
and (iii) studying the influence of both modeling errors (e.g., errors in camera calibration and depth estimation) and measurement noises on the visual servoing. 
\definecolor{applegreen}{rgb}{0.7, 1, 0.0}
\definecolor{LightCyan}{rgb}{0.88,1,1}
\definecolor{atomictangerine}{rgb}{1.0, 0.6, 0.4}
\definecolor{amber}{rgb}{1.0, 0.75, 0.0}
\definecolor{aqua}{rgb}{0.0, 1.0, 1.0}
\definecolor{almond}{rgb}{0.94, 0.87, 0.8}
\definecolor{aquamarine}{rgb}{0.5, 1.0, 0.83}
\definecolor{babyblue}{rgb}{0.54, 0.81, 0.94}
\definecolor{babyblueeyes}{rgb}{0.63, 0.79, 0.95}
\definecolor{asparagus}{rgb}{0.53, 0.66, 0.42}
\definecolor{auburn}{rgb}{0.43, 0.21, 0.1}
\definecolor{brilliantlavender}{rgb}{0.96, 0.73, 1.0}
\definecolor{bittersweet}{rgb}{1.0, 0.44, 0.37}
\definecolor{blue-violet}{rgb}{0.54, 0.17, 0.89}
\definecolor{capri}{rgb}{0.0, 0.75, 1.0}
\definecolor{celadon}{rgb}{0.67, 0.88, 0.69}
\definecolor{darkcyan}{rgb}{0.0, 0.55, 0.55}
\definecolor{deepskyblue}{rgb}{0.0, 0.75, 1.0}
\definecolor{dogwoodrose}{rgb}{0.84, 0.09, 0.41}
\begin{table}[!ht]
\small\addtolength{\tabcolsep}{-4.0pt} 
\caption{Overall performance of \textit{VS} control schemes}
\centering
\begin{tabular}{|c||c|c|c|c|c|c|} 
\hline
 \rowcolor{applegreen}
 Test No. & Control Scheme & $\mathcal{R}_{\text{LM}}$  &  $\mathcal{P}_{\text{out}}$ & $\mathcal{R}_{\text{JL}}$   & $\mathcal{N}_{\text{success}}$ & $\mathcal{S}_{\text{rate}}$ \\
 \hline  
 \hline
 \rowcolor{LightCyan}
 \multicolumn{7}{|c|}{\textit{MPPI-IBVS} and \textit{C-IBVS} Control Schemes} \\
 \hline
 \noalign{\vspace*{2pt}}
 \arrayrulecolor{pink}\hline  
\cellcolor{pink!40}Test \#1 & \textit{MPPI- (CASE \#0)} & 0 & 0 & 1 & 119 & \textcolor{green}{99.16\%} \\
\hline
\cellcolor{pink!40}Test \#2 & \textit{MPPI- (CASE \#1)} & 0 & 0 & 29 & 91 & 75.8\% \\
\hline
\cellcolor{pink!40}Test \#3 & \textit{MPPI- (CASE \#2)} & 0 & 0 & 54 & 66 & \textcolor{red}{55\%} \\
\hline
\cellcolor{pink!40}Test \#4 & \textit{MPPI- (CASE \#3)} & 0 & 0 & 6 & 114 & 95\% \\
\hline
\cellcolor{pink!40}Test \#5 & \textit{MPPI- (CASE \#0)} & 0 & 0 & 0 & 120 & \textcolor{green}{100\%}\\
\hline
\noalign{\vspace*{2pt}} 
\arrayrulecolor{amber}\hline 
\cellcolor{amber!20}Test \#6 & \textit{C- (CASE \#0)} & 0 & 0 & 37 & 83 & 69.17\% \\
\hline
\cellcolor{amber!20}Test \#7 & \textit{C- (CASE \#0)} & 0 & 0 & 36 & 84 & 70\% \\
\hline 
\noalign{\vspace*{2pt}}
\arrayrulecolor{auburn}\hline
\cellcolor{auburn!15}Test \#8 & \textit{MPPI- (CASE \#0)} & 0 & 0 & 19 & 101 & 84.17\% \\
\hline
\cellcolor{auburn!15}Test \#9 & \textit{MPPI- (CASE \#0)} & 0 & 0 & 35 & 85 & 70.8\% \\
\hline
\cellcolor{auburn!15}Test \#10 & \textit{MPPI- (CASE \#0)} & 0 & 0 & 7 & 113 & \textcolor{green}{94.17\%} \\
\hline 
\noalign{\vspace*{2pt}}
\arrayrulecolor{bittersweet}\hline
\cellcolor{bittersweet!40}Test \#11 & \textit{MPPI- (CASE \#0)} & 0 & 0 & 15 & 105 & 87.5\% \\
\hline 
\noalign{\vspace*{2pt}}
\arrayrulecolor{babyblue}\hline
\cellcolor{babyblue!30}Test \#12 & \textit{MPPI- (CASE \#0)} & 0 & 0 & 25 & 95 & 79.17\% \\
\hline
\cellcolor{babyblue!30}Test \#13 & \textit{C- (CASE \#0)} & 0 & 0 & 42 & 78 & 65\% \\
\hline
\noalign{\vspace*{2pt}}
\arrayrulecolor{asparagus}\hline
\cellcolor{asparagus!30}Test \#14 & \textit{MPPI- (CASE \#0)} & 0 & 0 & 0 & 120 & \textcolor{green}{100\%} \\
\hline
\cellcolor{asparagus!30}Test \#15 & \textit{MPPI- (CASE \#0)} & 0 & 0 & 2 & 118 & 98.33\% \\
\hline
\cellcolor{asparagus!30}Test \#16 & \textit{MPPI- (CASE \#0)} & 0 & 0 & 5 & 115 & 95.83\% \\
\hline
\cellcolor{asparagus!30}Test \#17 & \textit{MPPI- (CASE \#0)} & 0 & 0 & 17 & 103 &85.83\% \\
\hline
\cellcolor{asparagus!30}Test \#18 & \textit{MPPI- (CASE \#0)} & 0 & 0 & 4 & 116 & 96.67\% \\
\hline
\noalign{\vspace*{2pt}}
\arrayrulecolor{black}\hline
 \rowcolor{LightCyan}
 \multicolumn{7}{|c|}{\textit{MPPI-3DVS} and \textit{C-3DVS} Control Schemes} \\
 \hline
 \noalign{\vspace*{2pt}}
\arrayrulecolor{blue-violet}\hline
\cellcolor{blue-violet!30}Test \#19 & \textit{MPPI- (${ }^{o}\boldsymbol{P}_{c_1}^{\ast}$)} & 0 & 0 & 0 & 120 & \textcolor{green}{100\%} \\
\hline
\cellcolor{blue-violet!30}Test \#20 & \textit{C- (${ }^{o}\boldsymbol{P}_{c_1}^{\ast}, \lambda_{\mathbf{s}}=0.5)$} & 0 & 0 & 4 & 116 & 96.67\% \\
\hline
\cellcolor{blue-violet!30}Test \#21 & \textit{MPPI- (${ }^{o}\boldsymbol{P}_{c_2}^{\ast}$)} & 0 & 0 & 5 & 115 & 95.83\% \\
\hline
\cellcolor{blue-violet!30}Test \#22 & \textit{C- (${ }^{o}\boldsymbol{P}_{c_2}^{\ast}, \lambda_{\mathbf{s}}=0.5$)} & 0 & 1 & 6 & 113 & 94.17\% \\
\hline
\noalign{\vspace*{2pt}}
\arrayrulecolor{black}\hline
 \rowcolor{LightCyan}
 \multicolumn{7}{|c|}{\textit{MPPI-PBVS} and \textit{C-PBVS} Control Schemes} \\
 \hline
 \noalign{\vspace*{2pt}}
\arrayrulecolor{capri}\hline
\cellcolor{capri!30}Test \#23 & \textit{MPPI- (${ }^{o}\boldsymbol{P}_{c_1}^{\ast}$)} & 0 & 45 & 0 & 75 & \textcolor{red}{62.50\%} \\
\hline
\cellcolor{capri!30}Test \#24 & \textit{C- (${ }^{o}\boldsymbol{P}_{c_1}^{\ast}, \lambda_{\mathbf{s}}=0.5$)} & 0 & 26 & 0 & 94 & 78.33\% \\
\hline
\cellcolor{capri!30}Test \#25 & \textit{C- (${ }^{o}\boldsymbol{P}_{c_2}^{\ast}, \lambda_{\mathbf{s}}=0.5$)} & 0 & 22 & 0 & 98 & \textcolor{green}{81.17\%} \\
\hline
\noalign{\vspace*{2pt}}
\arrayrulecolor{dogwoodrose}\hline
\cellcolor{dogwoodrose!30}Test \#26 & \textit{MPPI (${ }^{o}\boldsymbol{P}_{c_1}^{\ast}$, (\ref{eq:17}))} & 0 & 0 & 0 & 120 & \textcolor{green}{100\%} \\
\hline
\cellcolor{dogwoodrose!30}Test \#27 & \textit{MPPI (${ }^{o}\boldsymbol{P}_{c_2}^{\ast}$, (\ref{eq:17}))} & 0 & 0 & 0 & 120 & \textcolor{green}{100\%} \\
\hline
\cellcolor{dogwoodrose!30}Test \#28 & \textit{MPPI (${ }^{o}\boldsymbol{P}_{c_1}^{\ast}$, (\ref{eq:18}))} & 0 & 0 & 0 & 120 & \textcolor{green}{100\%} \\
\hline
\end{tabular}
\label{table:intensiveSimulationTable}
\end{table}

Table~\ref{table:intensiveSimulationTable} summarizes the overall performance of each control scheme individually, given the $120$ initial camera configurations.   
In total, we conduct a set of $28$ different tests. The first $18$ tests, included in the intensive studies, focus on studying the quality and robustness of both \textit{MPPI-IBVS} and \textit{C-IBVS} control schemes, while the remaining tests assess the performance of \textit{MPPI-3DVS} and \textit{MPPI-PBVS} compared to \textit{C-3DVS} and \textit{C-PBVS} control schemes.
Furthermore, the simulations are carried out considering ${ }^{o}\boldsymbol{P}_{c_1}^{\ast}$ as the desired camera configuration (unless mentioned otherwise, i.e., ${ }^{o}\boldsymbol{P}_{c_2}^{\ast}$). 

\subsubsection{\textcolor{pink}{\textbf{Prediction Process Influence}}}
In the first four tests (namely, from Test \#1 to \#4), we analyzed the impact of the image prediction process on the performance of the \textit{MPPI-IBVS} control strategy taken into account the four cases of the approximate interaction matrix ${}^{2d}\widehat{\mathbf{L}}_{\mathbf{s}}$, that have been previously discussed in Section~\ref{classic-ibvs}, and the \textit{MPPI-IBVS} parameters listed in Table~\ref{table:SysParameters}. 
It can be clearly seen that the prediction process exerts a high influence as \textit{MPPI} provides better performance in \textit{CASE} \#0 and \#3, in which the \textit{2D} visual features and their depth information are assumed to be estimated each iteration.
While the worst performance is achieved when ${}^{2d}\widehat{\mathbf{L}}_{\mathbf{s}}$ is assumed to be constant, as depicted in Test \#3. 
More accurately, for Test \#1, we conducted $3$ trials. For all the trials, our proposed control scheme succeeded in completing $119$ out of \num{120} tasks/configurations (i.e., $\mathcal{N}_{\text{success}}=119$), while there existed only one initial camera configuration led the robot to reach one of its joint limits (i.e., $\mathcal{R}_{\text{JL}} = 1$). 
It is noteworthy that this one failure case can be readily tackled by increasing the prediction horizon as illustrated in Test \#5 where we adopted $t_p=\SI{5}{\second}$ and $K=1000$ instead of $t_p=\SI{3.5}{\second}$ and $K=2000$, showing the superiority of our proposed control strategy even without involving the robot joint limits in the prediction algorithm. Moreover, $K$ is decreased to half of its nominal value so as to ensure a \textit{real-time} implementation of \textit{MPPI-IBVS}, without compromising its robustness level and convergence rate.
\subsubsection{\textcolor{amber}{\textbf{\textit{C-IBVS} Control Scheme}}} To better understand the potential advantages of \textit{MPPI-IBVS} in improving the performance of visual servoing, intensive simulations of the classical IBVS control scheme are carried out in Test \#6 and \#7 in which we set $\lambda_{\mathbf{s}}$ to \num{0.5} and \num{0.2}, respectively. 
\begin{figure}[th!]
\centering
\begin{subfigure}{.25\textwidth}
  \centering
  \includegraphics[width=1\columnwidth,height=1.4in]{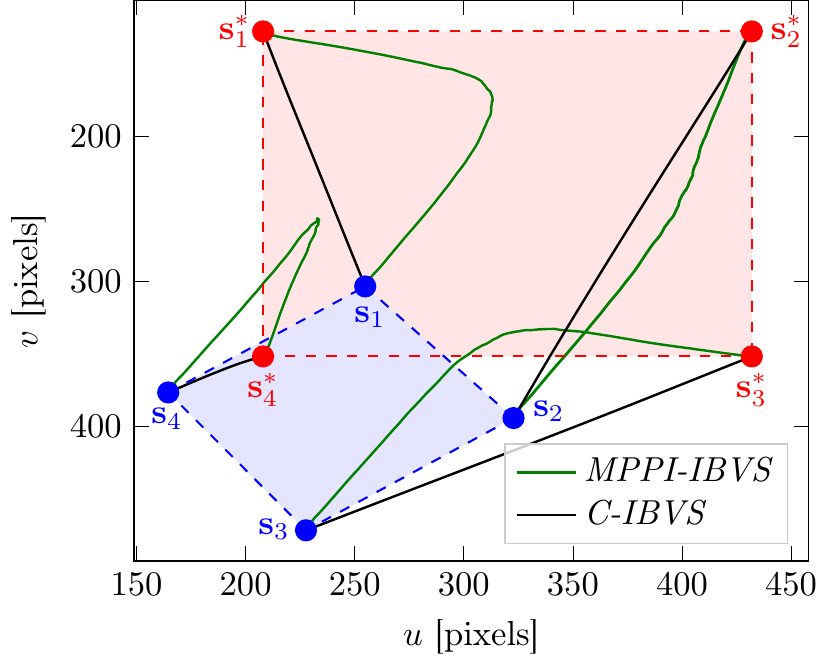}
  \caption{Image points trajectories} 
  \label{fig:2Dtrajectories_S113_ibvs}
\end{subfigure}%
\begin{subfigure}{.25\textwidth}
  \centering
  \includegraphics[width=1\columnwidth,height=1.4in]{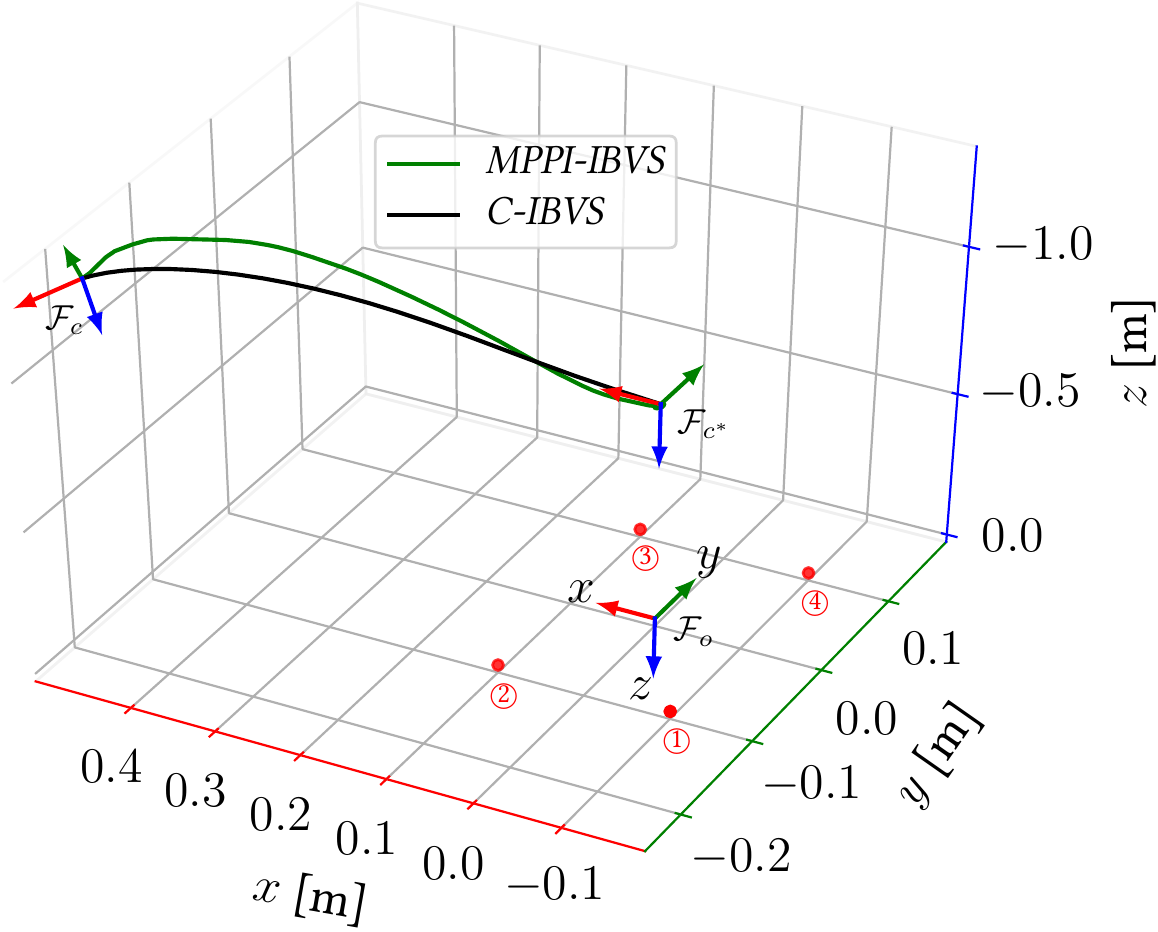}
  \caption{3D camera trajectories}
  \label{fig:3Dtrajectory_S113_ibvs}
\end{subfigure}
\par\medskip 
\begin{subfigure}{.25\textwidth}
  \centering
  \includegraphics[width=1\columnwidth,height=0.75in]{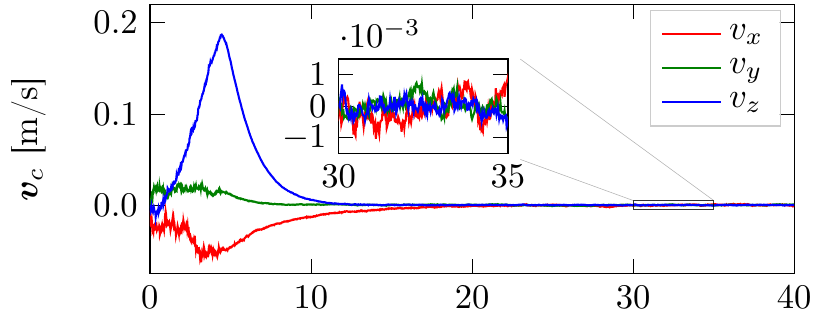}
  \includegraphics[width=1\columnwidth,height=0.85in]{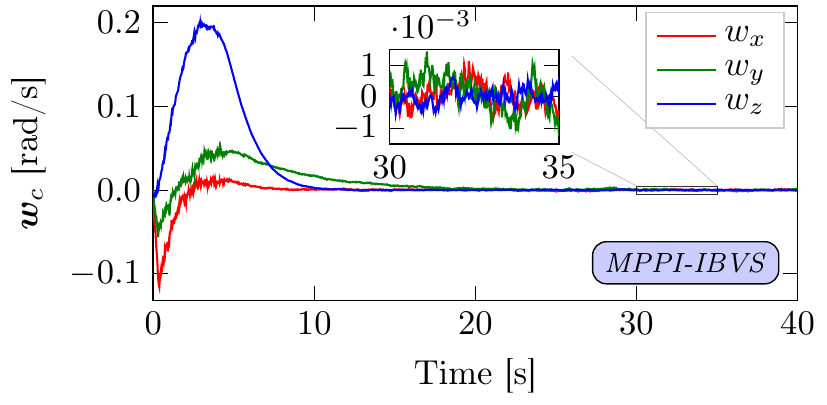}
  \caption{Camera velocity \textit{(MPPI-IBVS)}}
  \label{fig:cameraVelocity_S113_mppi_ibvs}
\end{subfigure}%
\begin{subfigure}{.25\textwidth}
  \centering
  \vspace{-3pt}
  \includegraphics[width=1\columnwidth,height=0.8in]{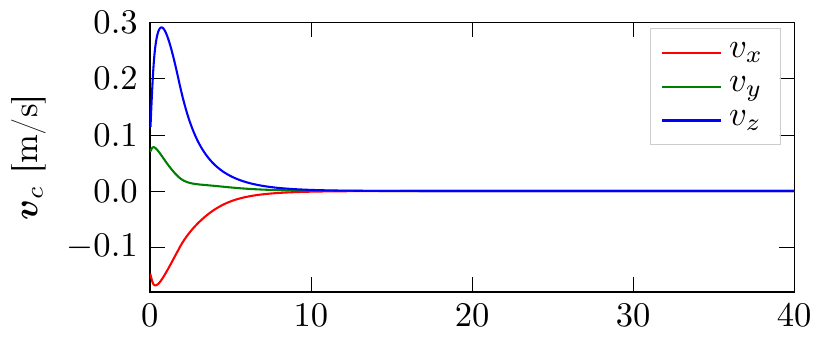}
  \includegraphics[width=1\columnwidth,height=0.85in]{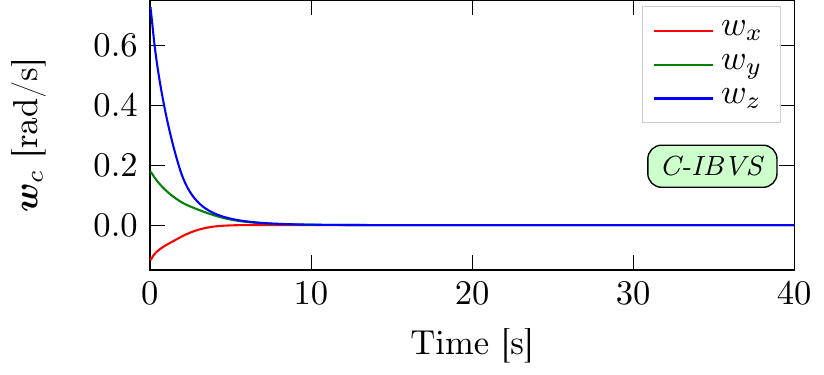}
  \caption{Camera velocity \textit{(C-IBVS)}}
  \label{fig:cameraVelocity_S113_classic_ibvs}
\end{subfigure}
\par\medskip 
\begin{subfigure}{.25\textwidth}
  \centering
  \includegraphics[width=1\columnwidth,height=0.8in]{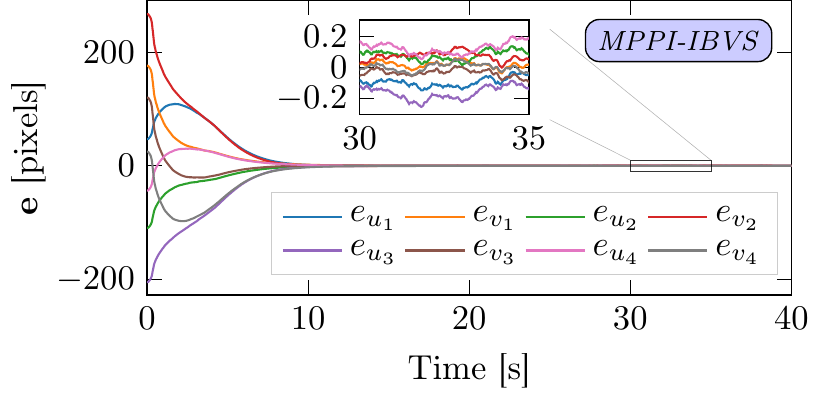}
  \caption{\textit{2D} features error \textit{(MPPI-IBVS)}}
  \label{fig:2DErrors_S113_mppi_ibvs}
\end{subfigure}%
\begin{subfigure}{.25\textwidth}
  \centering
  \includegraphics[width=1\columnwidth,height=0.8in]{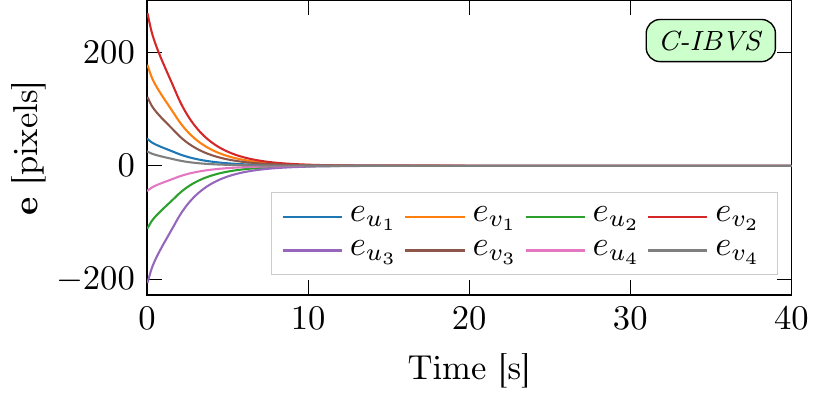}
  \caption{\textit{2D} features error \textit{(C-IBVS)}}
  \label{fig:2DErrors_S113_classic_ibvs}
\end{subfigure}%
\caption{Simulation results of \textit{MPPI-IBVS} and \textit{C-IBVS} for the positioning task \#113 that has an initial camera configuration of $\left(\left[0.44, -0.23, -1.35\right], \left[10.95, -20.48, -50.15\right]\right)^T$ in ([\si{\metre}], [\si{\deg}]), considering ${ }^{o}\boldsymbol{P}_{c_1}^{\ast}$ as the desired configuration.}
\label{fig:s113_ibvs}
\end{figure}
As anticipated, the success rate of visual servoing tasks was reduced to $70\%$, with \num{36}-$\mathcal{R}_{\text{JL}}$ failure cases. In fact, these failure cases occurred due to the fact that their corresponding initial camera configurations have large rotations around the camera optical axis as well as their positions along the $z$-axis are much closer to the minimum allowable limit of the third joint which is approximately \SI{-1.45}{\metre} relative to the object reference frame $\mathcal{F}_{o}$. Thus, thanks to the prediction process, it has been possible to provide better camera trajectories without violating the robot joint bounds.
In Fig.~\ref{fig:s113_ibvs}, we illustrate the behavior of \textit{MPPI-IBVS} compared to \textit{C-IBVS} through an example from the successful servoing tasks (namely, task \#113) in both Tests \#1 and \#6. 
We can observe that despite the trajectories of the four points in the image are less satisfactory than those obtained from \textit{C-IBVS}, our proposed control scheme provides a smooth trajectory of the camera in the Cartesian space (see Figs.~\ref{fig:2Dtrajectories_S113_ibvs} and \ref{fig:3Dtrajectory_S113_ibvs}), with very low fluctuations in the camera velocity components as depicted in Fig.~\ref{fig:cameraVelocity_S113_mppi_ibvs}.
Furthermore, it is worthy to notice that \textit{MPPI-IBVS} ensures convergence in both the image and \textit{3D} space as in \textit{C-IBVS}; however, the convergence of the image points is no longer perfectly exponential (see Fig.~\ref{fig:2DErrors_S113_mppi_ibvs}).
Herein, the system converges within approximately \SI{19.82}{\second} (or, 991 iterations) compared to \SI{14.46}{\second} (or, 723 iterations) in the case of \textit{C-IBVS}.
\subsubsection{\textcolor{auburn}{\textbf{\textit{MPPI} Parameters Influence}}} 
In Tests \#1 and \#5, we have clearly seen that fine-tuning the parameters of \textit{MPPI} plays an important role in determining its behavior and improving the quality of positioning tasks. 
As a consequence, the boundary of the workspace and joint limits have been avoided during the servoing operation.
Therefore, in the next three tests starting from Test \#8, the influence of both control input updates $\Sigma_{\mathbf{u}}$, prediction horizon $t_p$, and number of sampled trajectories $K$ is individually studied. 
In Test \#8, we adopted $\Sigma_{\mathbf{u}}=\operatorname{Diag}\left(0.009, 0.009, 0.009, 0.03, 0.03, 0.02\right)$, while in Test \#9 $t_p$ is set to \SI{1}{\second}. Finally, we set $K$ to \num{100} in Test \#10.
In general, the intensive simulations demonstrate that the impact of either having short-time horizons $t_p$ or changing the control input updates $\Sigma_{\mathbf{u}}$ is appreciably higher than the influence of decreasing the number of samples $K$, as the success rate $\mathcal{S}_{\text{rate}}$ in Tests \#8 and \#9 is significantly lower than that in Test \#10.
Concerning the quality of successful servoing tasks, it is interesting to notice that assigning very low values to  $t_p$ and $K$ affects the convergence rate\footnote{Notice that the simulation time of Test \#9 is adopted to be \SI{180}{\second} as the convergence rate of \textit{MPPI}, in this case, is extremely slow.} and quality of the trajectories in both the image and \textit{3D} space.  
Just to name a few, for the positioning task \#113 given in Fig.~\ref{fig:s113_ibvs}, the system converges to its desired pose within about \SI{165.84}{\second} with $t_p=\SI{1}{\second}$, instead of taking \SI{19.82}{\second} in case of $t_p=\SI{3.5}{\second}$.

\subsubsection{\textcolor{bittersweet}{\textbf{Handling Control and \textit{3D} Constraints}}} 
In Test \#11, we repeated the simulations of Test \#1 under the assumption that the maximum control input of the camera velocity screw $\mathbf{v}_{\max}$ is limited to \SI{0.5}{\metre/ \second} for the translational speed and \SI{0.3}{\radian/ \second} for the rotational speed, while the minimum control input $\mathbf{v}_{\min}$ equals to $- \mathbf{v}_{\max}$.
Broadly speaking, we can observe that restricting the control input affects the overall performance of \textit{MPPI-IBVS}, particularly for those camera configurations that require large motions especially, in our simulations, those containing large rotations around $Z$-axis. 
Within the present test, the \num{15}-$\mathcal{R}_{\text{JL}}$ failure cases can be easily avoided and tackled by involving the \textit{3D} constraints, especially joint limits constraints, as a part of the \textit{MPPI-IBVS} optimization problem.
On the other side, the obtained results of the \num{105} successful tasks demonstrate that our proposed control scheme performs perfectly with a high capability of handling the control constraints.
\begin{figure}[th!]
\centering
\begin{subfigure}{.25\textwidth}
  \centering
  \includegraphics[width=1\columnwidth,height=1.85in]{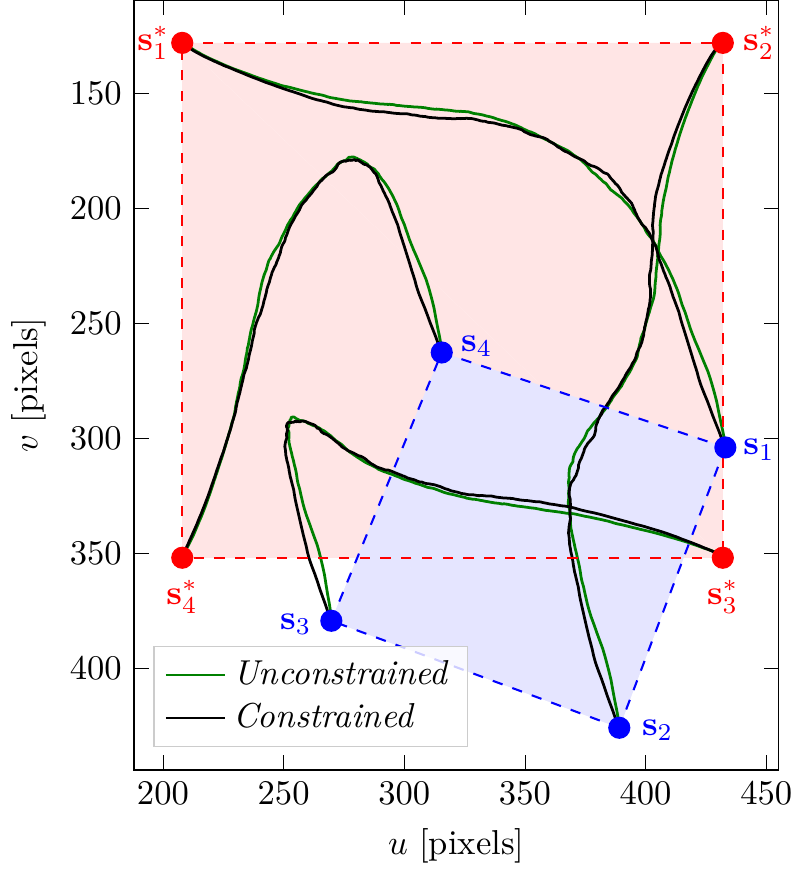}
  \caption{Image points trajectories} 
  \label{fig:2Dtrajectories_S45_ibvs}
\end{subfigure}%
\begin{subfigure}{.25\textwidth}
  \centering
  \includegraphics[width=1\columnwidth,height=1.1in]{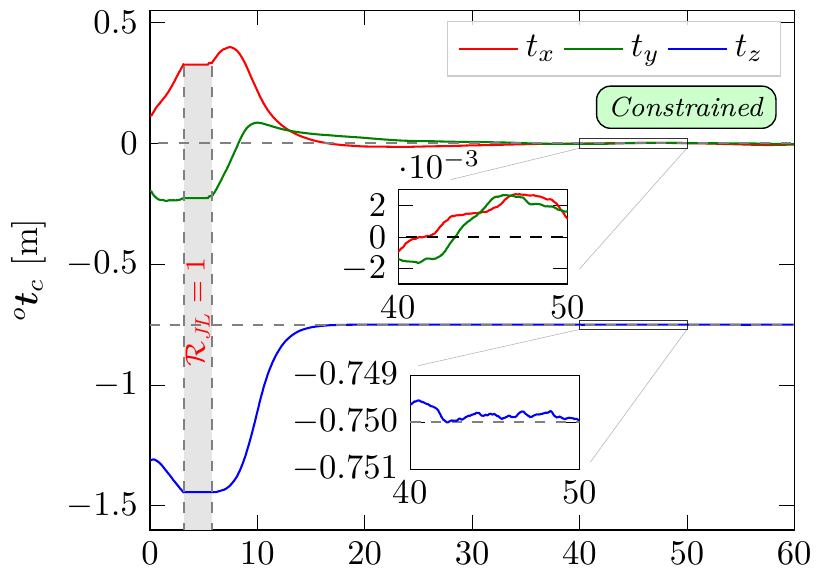}
  \includegraphics[width=1\columnwidth,height=0.75in]{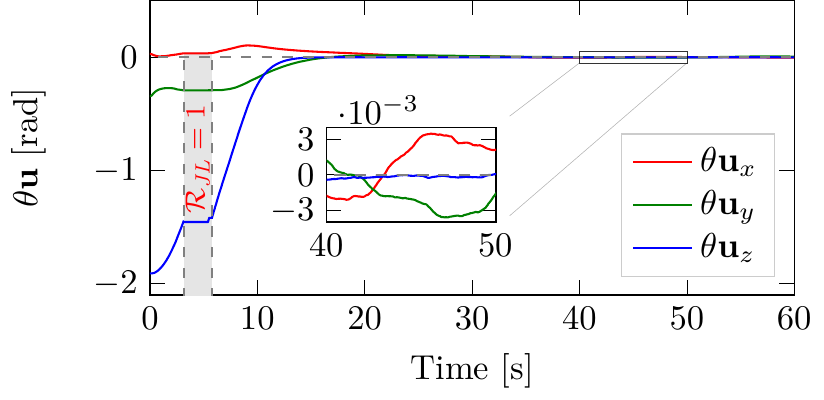}
  \caption{Camera pose ${ }^{o}\boldsymbol{P}_{c}$ in $\mathcal{F}_{o}$}
  \label{fig:CameraPose_S45_ibvs}
\end{subfigure}
\par\medskip 
\begin{subfigure}{.25\textwidth}
  \centering
  \includegraphics[width=1\columnwidth,height=0.75in]{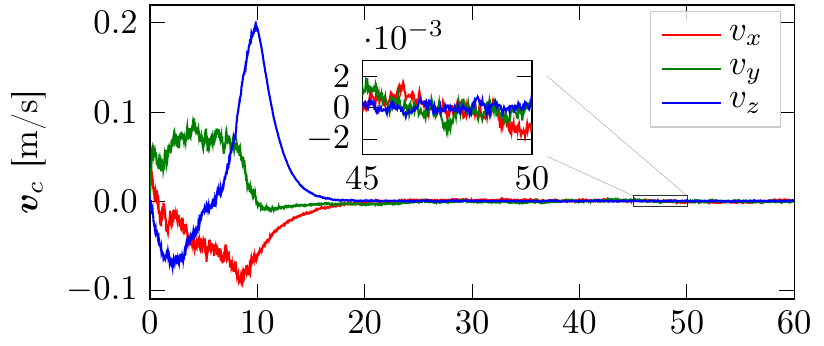}
  \includegraphics[width=1\columnwidth,height=0.85in]{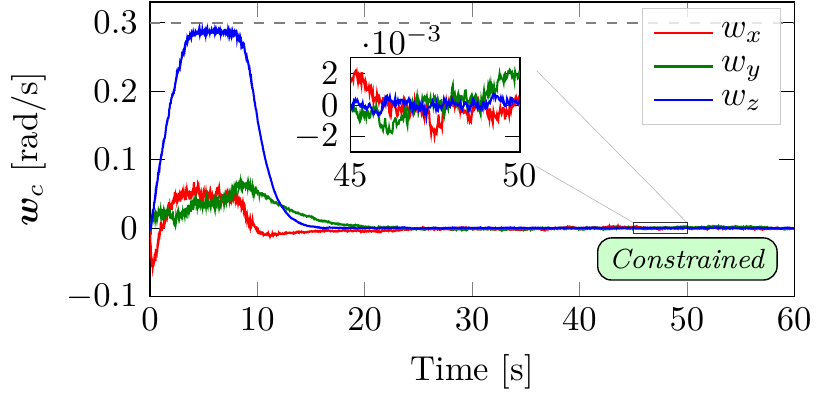}
  \caption{Constrained camera velocity}
  \label{fig:cameraVelocity_S45_mppi_with}
\end{subfigure}%
\begin{subfigure}{.25\textwidth}
  \centering
  \includegraphics[width=1\columnwidth,height=0.75in]{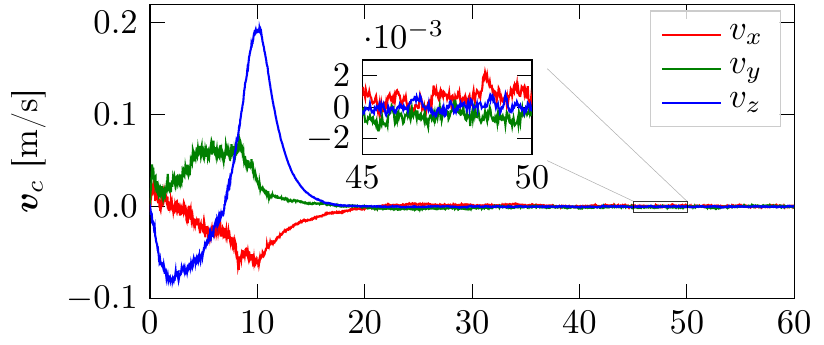}
  \includegraphics[width=1\columnwidth,height=0.85in]{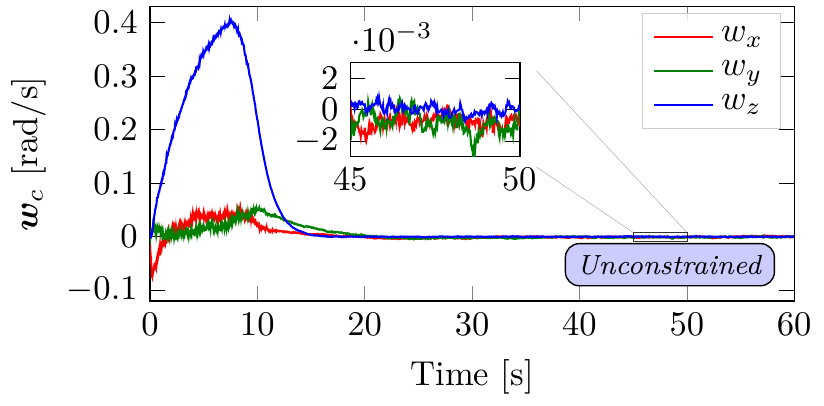}
  \caption{Unconstrained camera velocity}
  \label{fig:cameraVelocity_S45_mppi_without}
\end{subfigure}
\caption{Simulation results of \textit{MPPI-IBVS} for the positioning task \#45 that has an initial camera configuration of $\left(\left[0.115 , -0.197, -1.311\right], \left[1.67, -19.58, -109.34\right]\right)^T$ in ([\si{\metre}], [\si{\deg}]), considering ${ }^{o}\boldsymbol{P}_{c_1}^{\ast}$ as the desired configuration.}
\label{fig:s45_ibvs}
\end{figure}

Figure~\ref{fig:s45_ibvs} shows the behavior of \textit{MPPI-IBVS} for servoing task \#45; in such a configuration, \textit{C-IBVS} fails as the third joint reaches its allowable lower limit due to the large translational and rotational motions.
As can be clearly seen in Fig.~\ref{fig:2Dtrajectories_S45_ibvs}, the trajectories of the $4$ point-like features in the image in the constrained case are almost similar to that in the unconstrained case.  
The corresponding translational and rotational motions of the camera expressed in $\mathcal{F}_{o}$ are given in Fig.~\ref{fig:CameraPose_S45_ibvs}.
We can notice that the camera converges to the desired camera configuration ${ }^{o}\boldsymbol{P}_{c_1}^{\ast}$ with: (i) translational errors less than \SI{2}{\milli\metre} in both $x$- and $y$-directions and less than \SI{0.5}{\milli\metre} in the $z$-direction, and (ii) rotational errors less than \SI{0.003}{\radian} (i.e., \SI{0.17}{\degree}). 
It is also interesting to observe that the lower limit of the third joint is reached, as in \textit{C-IBVS}, at $t=\SI{3.16}{\second}$. Accordingly, the robot controller set the joints' velocities to zero (see the gray highlighted region in Fig.~\ref{fig:CameraPose_S45_ibvs}). Nevertheless, our proposed control scheme has the capability of getting out of it at $t=\SI{5.8}{\second}$, due to the stochastic nature of the optimal control sequence obtained by the \textit{MPPI} algorithm that mainly based on a stochastic sampling of system trajectories.
Finally, Figs.~\ref{fig:cameraVelocity_S45_mppi_with} and \ref{fig:cameraVelocity_S45_mppi_without} display the constrained and unconstrained optimal control input of the camera velocity screw $\mathbf{v}_{c}$. 
It is observed that the clamping function $g(\mathbf{v})$ that is used to restrict the control input $\mathbf{v}\equiv \mathbf{v}_{c}$ is only applied to the $w_z$ component (see Fig.~\ref{fig:cameraVelocity_S45_mppi_with}) since its value exceeds the given bound which is \SI{0.3}{\radian/ \second} (see Fig.~\ref{fig:cameraVelocity_S45_mppi_without}), while the remaining components of the camera velocity behave almost the same with unnoticeable changes.
\begin{figure}[th!]
\centering
\begin{subfigure}{.25\textwidth}
  \centering
  \includegraphics[width=1\columnwidth,height=1.4in]{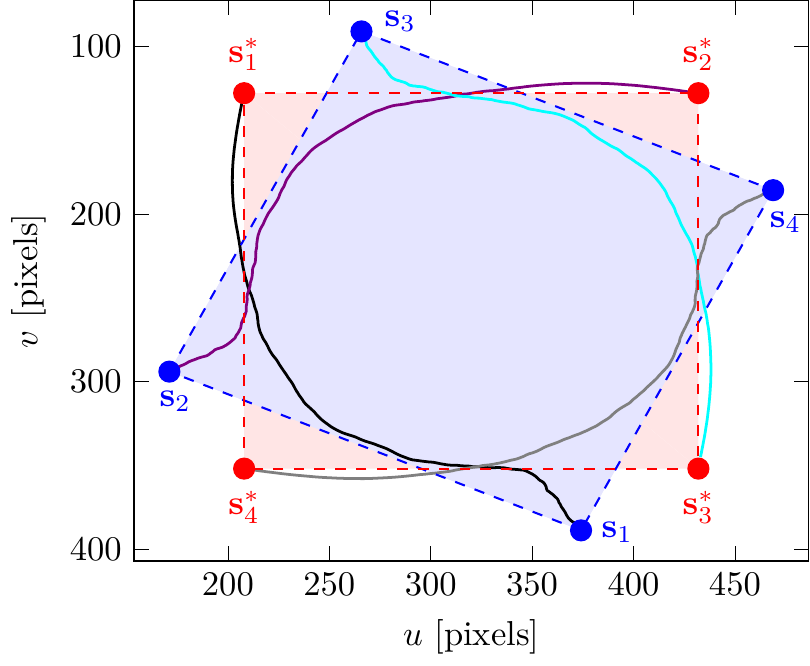}
  \caption{Image points trajectories} 
  \label{fig:2Dtrajectories_Treat_Problem}
\end{subfigure}%
\begin{subfigure}{.25\textwidth}
  \centering
  \includegraphics[width=1\columnwidth,height=1.4in]{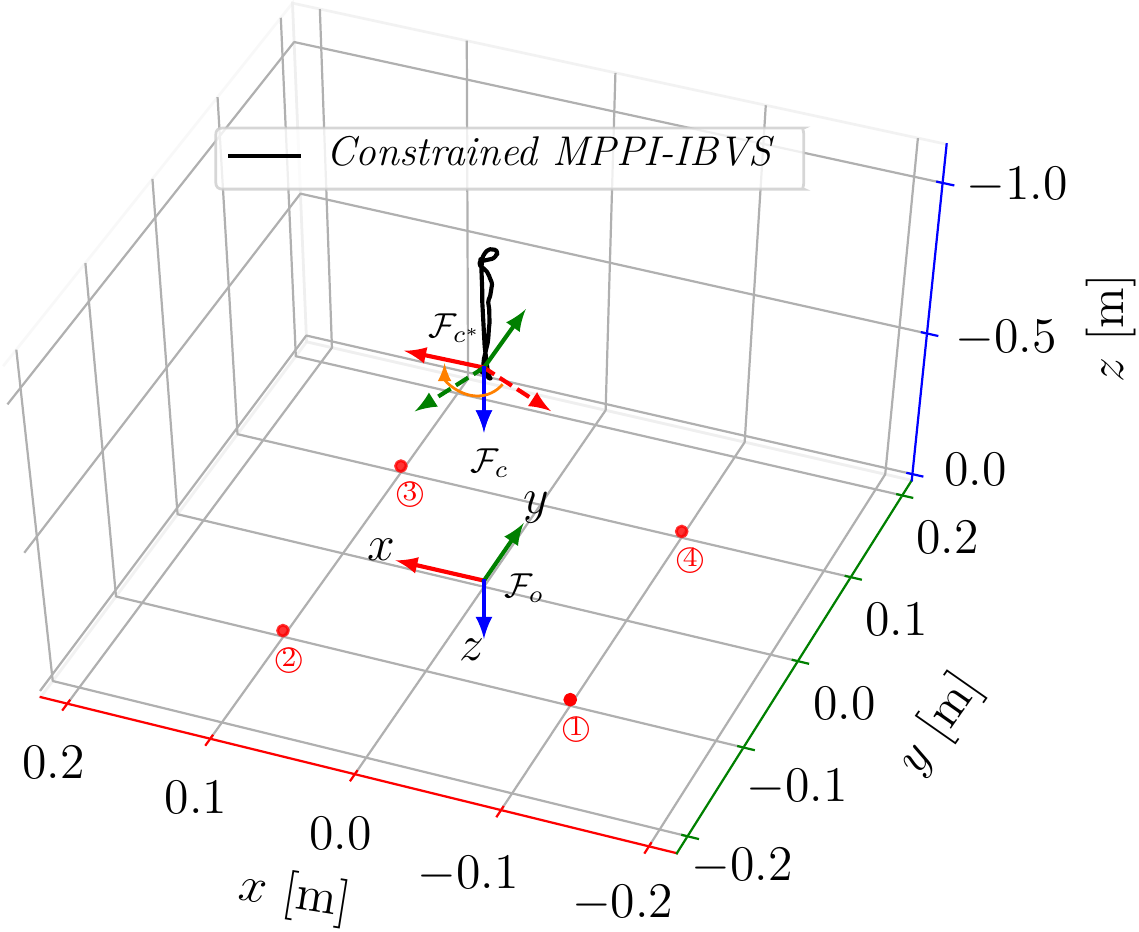}
  \caption{\textit{3D} camera trajectory}
  \label{fig:3DCamera_Treat_Problem}
\end{subfigure}
\par\medskip 
\begin{subfigure}{.25\textwidth}
  \centering
  \includegraphics[width=1\columnwidth,height=0.75in]{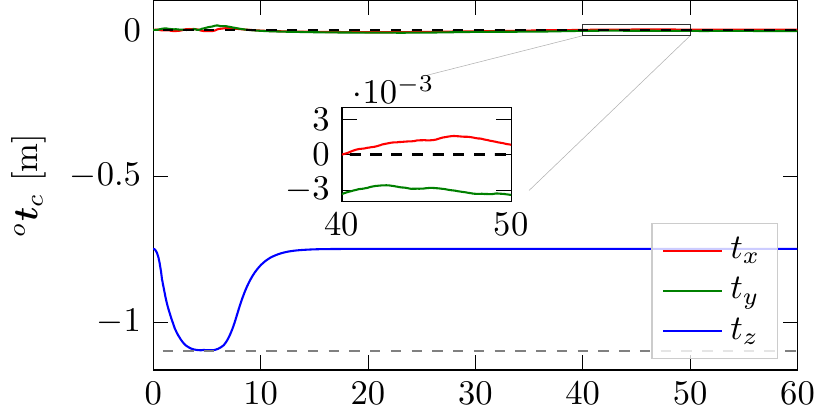}
  \includegraphics[width=1\columnwidth,height=0.85in]{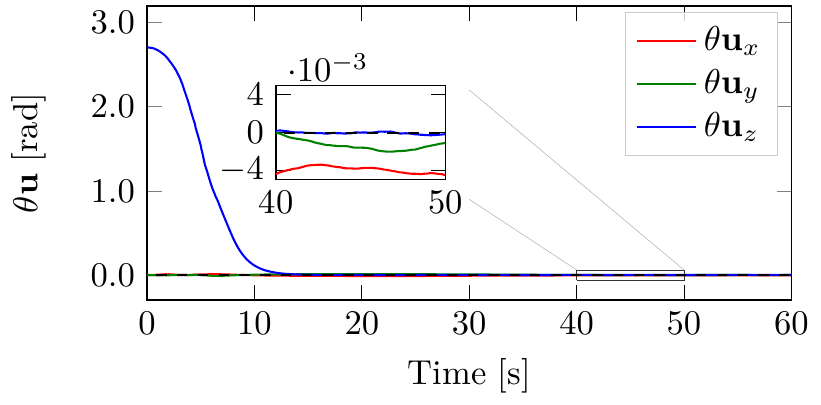}
  \caption{Camera pose ${ }^{o}\boldsymbol{P}_{c}$ in $\mathcal{F}_{o}$}
  \label{fig:cameraPose_Treat_Problem}
\end{subfigure}%
\begin{subfigure}{.25\textwidth}
  \centering
  \includegraphics[width=1\columnwidth,height=0.75in]{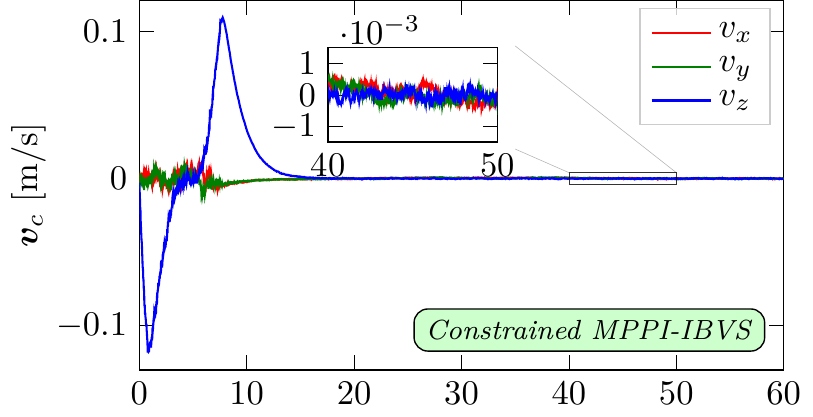}
  \includegraphics[width=1\columnwidth,height=0.85in]{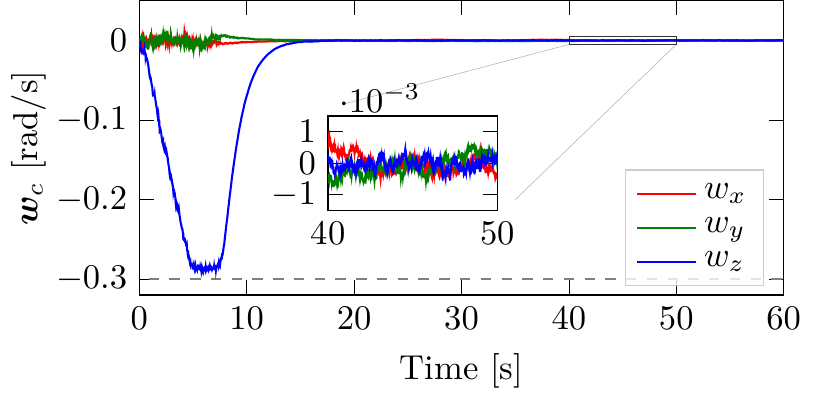}
  \caption{Constrained camera velocity}
  \label{fig:cameraVelocity_Treat_Problem}
\end{subfigure}
\caption{Behavior of constrained \textit{MPPI-IBVS} for a pure rotation of \SI{155}{\degree} around the camera's optical axis, considering ${ }^{o}\boldsymbol{P}_{c_1}^{\ast}$ as the desired configuration.}
\label{fig:treat_problem}
\end{figure}

Apart from the intensive simulations, we considered another servoing task, in which the camera performs a pure rotation of \SI{155}{\degree} about its optical axis\footnote{Notice that the maximum allowable rotational motion of our Cartesian robot around $Z$-axis (i.e., camera optical axis) is approximately \SI{157}{\degree} expressed in the object frame $\mathcal{F}_{o}$.}, as an example to illustrate the \textit{3D} constraints handling and how the camera retreat motion can be avoided as previously discussed in Section~\ref{sec:MPPI-VS Control Strategy}. 
In such a configuration, both \textit{C-IBVS} and unconstrained \textit{MPPI-IBVS} (with \textit{CASE \#0}) got stuck in the lower limit of the third joint due to the camera retreat motion.
However, once we set the maximum allowable camera retreat motion $Z_{\max}$ to \SI{-1.1}{\metre} with respect to $\mathcal{F}_{o}$, the constrained \textit{MPPI-IBVS} algorithm has the capability of converging easily to the desired camera configuration by penalizing the $Z$-axis translation motion of each \textit{3D} point as illustrated in Figs.~\ref{fig:2Dtrajectories_Treat_Problem}-\ref{fig:cameraPose_Treat_Problem}, without violating the control constraints (see Fig.~\ref{fig:cameraVelocity_Treat_Problem}). 

\subsubsection{\textcolor{babyblue}{\textbf{Handling Visibility Constraints}}} 
In all tests where \textit{MPPI-IBVS} is utilized, the simulations are carried out taken into account the visibility constraints, which are defined by the following inequalities:
\begin{equation}
\left[\begin{array}{l}
u_{\min }=0 \\
v_{\min }=0
\end{array}\right] \leq \mathbf{s}_t \leq\left[\begin{array}{l}
u_{\max }=640 \\
v_{\max }=480
\end{array}\right] \text{in [\si{\pixels}]}.
\end{equation}
Additionally, in order to better evaluate its capability of handling the constraints, the simulations of Test \#1 are repeated in Test \#12, considering ${ }^{o}\boldsymbol{P}_{c_2}^{\ast}$ as the desired configuration instead of ${ }^{o}\boldsymbol{P}_{c_1}^{\ast}$. 
The intensive simulations presented in Table~\ref{table:intensiveSimulationTable} demonstrate the efficiency and capability of our proposed control strategy in coping easily with the visibility constraints, as the visual features always remain within the camera's FoV (i.e., $\mathcal{P}_{\text{out}}=0$ for all tests).
\begin{figure}[th!]
\centering
\begin{subfigure}{.25\textwidth}
  \centering
  \hspace{-0.2in}\includegraphics[width=0.98\columnwidth,height=1.5in]{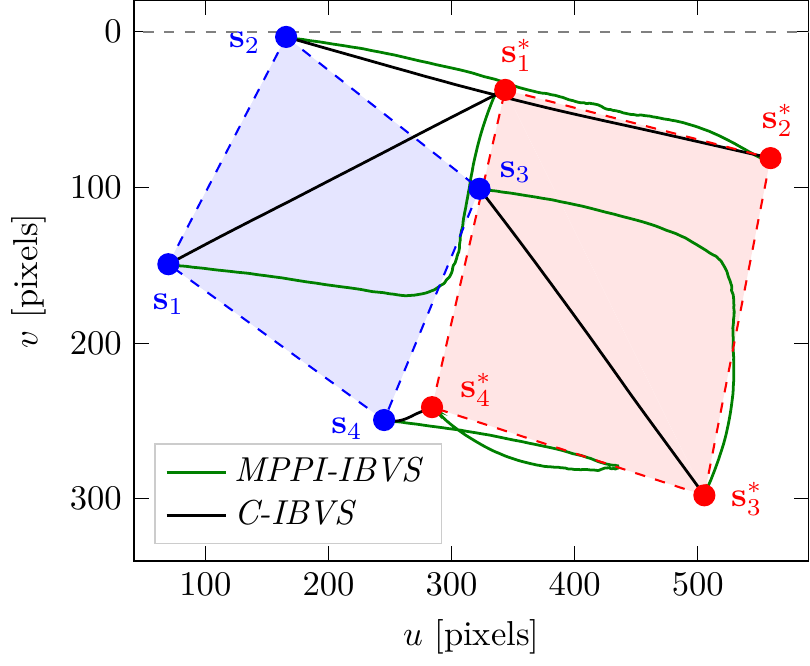}
  \caption{Positioning task \#12} 
  \label{fig:2Dtrajectories_S12_ibvs}
\end{subfigure}%
\begin{subfigure}{.25\textwidth}
  \centering
  \hspace{-0.1in}\includegraphics[width=0.98\columnwidth,height=1.5in]{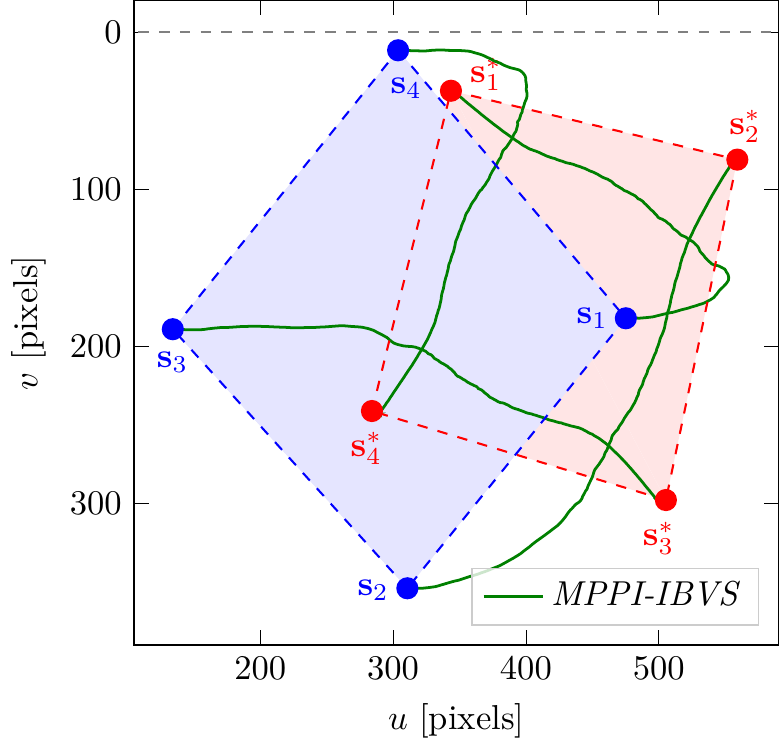}
  \caption{Positioning task \#100}
  \label{fig:2Dtrajectories_S100_ibvs}
\end{subfigure}
\caption{Features trajectories in the image for two successful servoing tasks, considering ${ }^{o}\boldsymbol{P}_{c_2}^{\ast}$ as the desired configuration.}
\label{fig:s12-100_ibvs}
\end{figure}
Two positioning tasks from Test \#12 are displayed in Fig.~\ref{fig:s12-100_ibvs}, illustrating the superiority of \textit{MPPI-IBVS} although one of the features is initially located (i) on the border of the image (see $\mathbf{s}_2$ in Fig.~\ref{fig:2Dtrajectories_S12_ibvs}) or (ii) near the border with large rotation along the optical axis (see $\mathbf{s}_4$ in Fig.~\ref{fig:2Dtrajectories_S100_ibvs}).
The initial camera configurations of tasks \#12 and \#100 are $\left(\left[-0.397, 0.001, -0.811\right], \left[2.59, 26.59, 60.83\right]\right)^T$ and $\left(\left[0.024, 0.038, -0.696\right], \left[2.95, 7.45, -133.56\right]\right)^T$ in ([\si{\metre}], [\si{\deg}]), respectively. 
Moreover, in Test \#13, we repeated the simulations of \textit{C-IBVS} given in Test \#6 with respect to ${ }^{o}\boldsymbol{P}_{c_2}^{\ast}$, not ${ }^{o}\boldsymbol{P}_{c_1}^{\ast}$. 
We can clearly observe that the $\mathcal{R}_{\text{JL}}$ failure cases increased owing to the complex translational and rotational motions associated with ${ }^{o}\boldsymbol{P}_{c_2}^{\ast}$, compared to ${ }^{o}\boldsymbol{P}_{c_1}^{\ast}$; one of the failure cases is task \#100, while \textit{C-IBVS} converges to the desired pose ${ }^{o}\boldsymbol{P}_{c_2}^{\ast}$ in task \#12, as shown in Fig.~\ref{fig:s12-100_ibvs}.

\subsubsection{\textcolor{asparagus}{\textbf{MPPI-IBVS Robustness}}}
To test the robustness of our proposed control strategy with respect to camera modeling errors and measurement noise, a set of intensive simulations has been considered in the next five tests. 
More precisely, in Test \#14, a white noise generated from a uniform distribution is added to the visual features and their \textit{3D} information (i.e., depth estimation) with a maximum error of $\pm$\SI{1}{\pixel} for \textit{2D} features and $\pm$\SI{0.5}{\cm} for \textit{3D} points. 
The influence of having a $+$\SI{30}{\%} and $-$\SI{60}{\%} error in the camera's focal length $f$ is individually investigated in Tests \#15 and \#16, respectively, whereas in Test \#17 we considered the case that combines various errors in the camera intrinsic parameters $\Gamma$: $+$\SI{30}{\%} in $f$, $-$\SI{20}{\%} in $\rho_u$, $+$\SI{20}{\%} in $\rho_v$, $-$\SI{15}{\%} in $u_0$, and $+$\SI{15}{\%} in $v_0$. Finally, the simulations carried out in Test \#18 take into consideration both the added measurement noise and camera calibration error given in Tests \#14 and \#15. 
\begin{figure}[!ht]
\centering
\begin{subfigure}{.25\textwidth}
  \centering
  \includegraphics[width=1\columnwidth,height=1.4in]{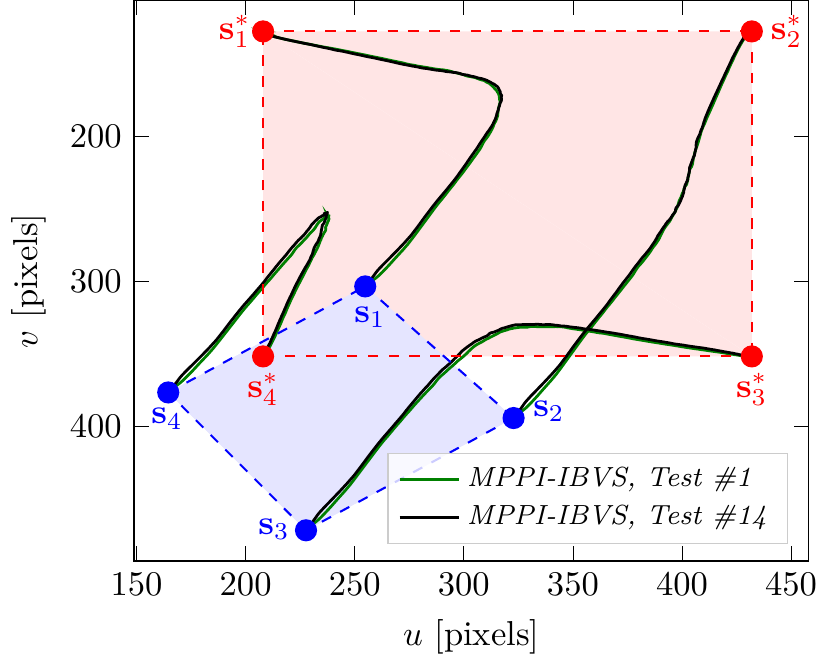}
  \caption{Image points trajectories} 
  \label{fig:2Dtrajectories_S113_ibvs_2pixels_errors}
\end{subfigure}%
\begin{subfigure}{.25\textwidth}
  \centering
  \includegraphics[width=1\columnwidth,height=1.4in]{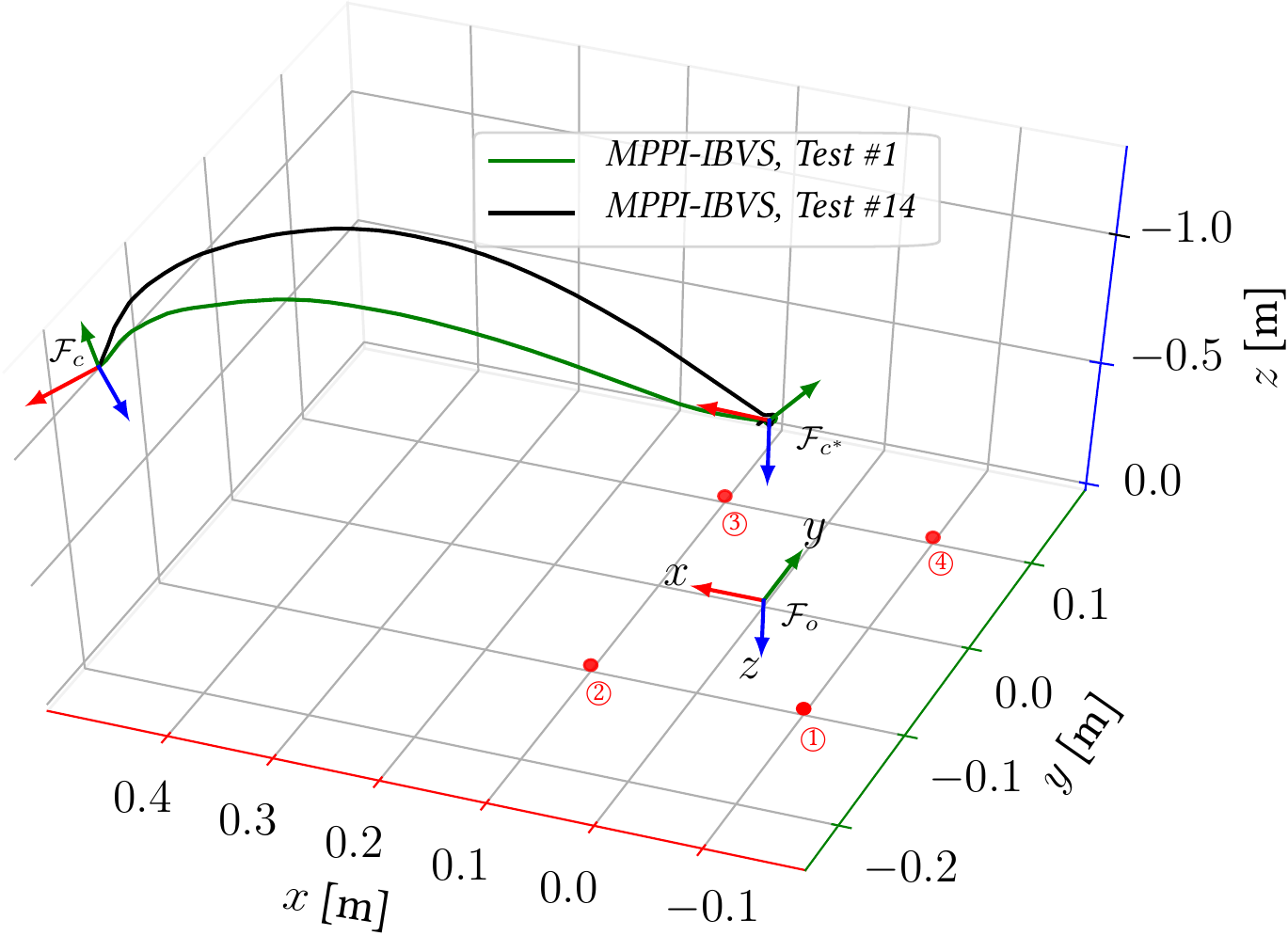}
  \caption{\textit{3D} camera trajectories}
  \label{fig:3Dtrajectory_S113_ibvs_2pixels_errors}
\end{subfigure}
\par\medskip 
\begin{subfigure}{.25\textwidth}
  \centering
  \includegraphics[width=1\columnwidth,height=0.7in]{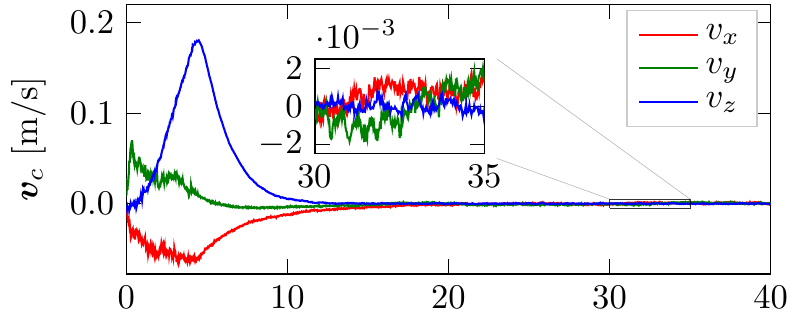}
  \includegraphics[width=1\columnwidth,height=0.9in]{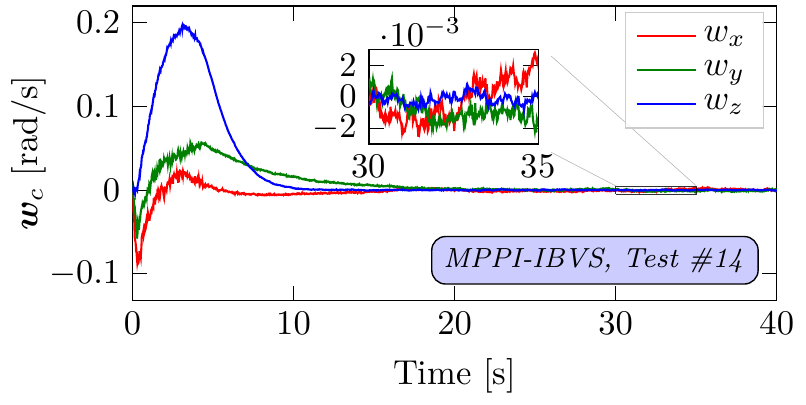}
  \caption{Camera velocity \textit{(MPPI-IBVS)}}
  \label{fig:cameraVelocity_S113_mppi_ibvs_2pixels_errors}
\end{subfigure}%
\begin{subfigure}{.25\textwidth}
  \centering
  \includegraphics[width=1\columnwidth,height=0.7in]{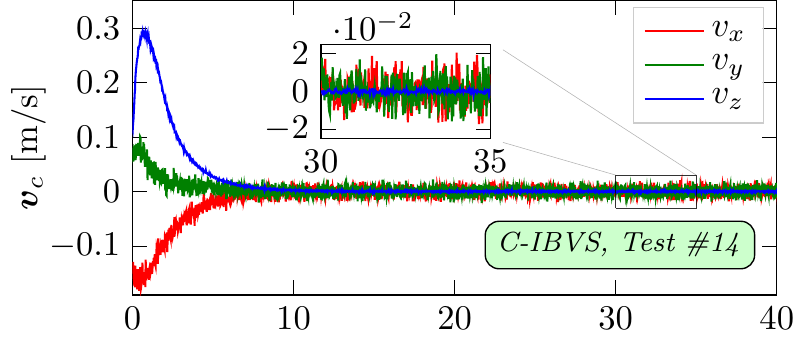}
  \includegraphics[width=1\columnwidth,height=0.89in]{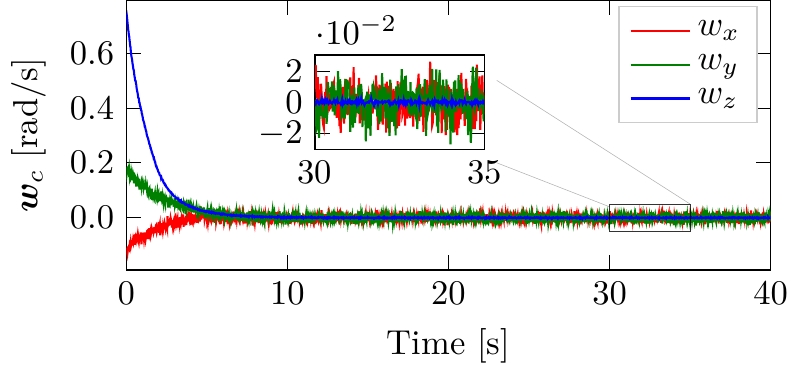}
  \caption{Camera velocity \textit{(C-IBVS)}}
  \label{fig:cameraVelocity_S113_classic_ibvs_2pixels_errors}
\end{subfigure}
\par\medskip 
\begin{subfigure}{.25\textwidth}
  \centering
  \hspace{-0.02in}\includegraphics[width=1.01\columnwidth,height=0.75in]{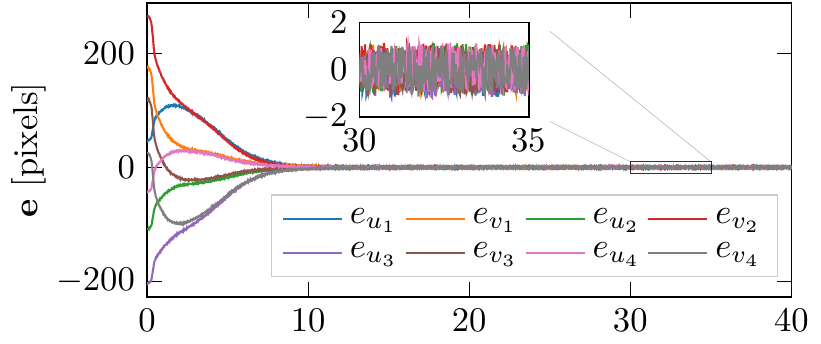}
  \includegraphics[width=1\columnwidth,height=0.75in]{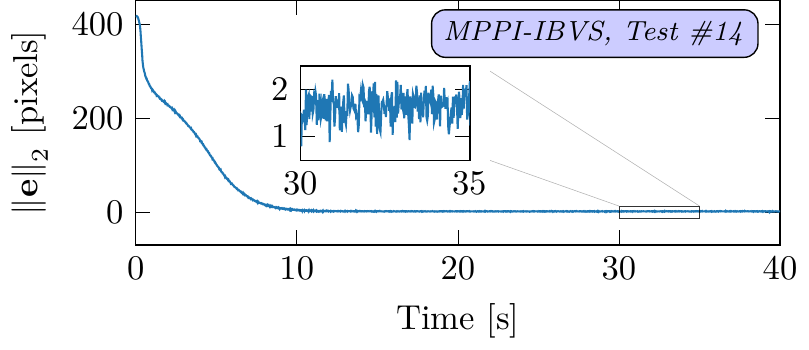}
  \caption{\textit{2D} features error \textit{(MPPI-IBVS)}}
  \label{fig:2DErrors_S113_mppi_ibvs_2pixels_errors}
\end{subfigure}%
\begin{subfigure}{.25\textwidth}
  \centering
  \hspace{-0.02in}\includegraphics[width=1.01\columnwidth,height=0.75in]{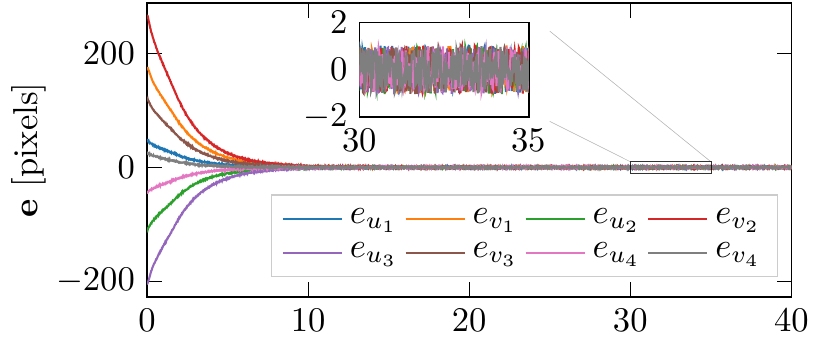}
  \includegraphics[width=1\columnwidth,height=0.75in]{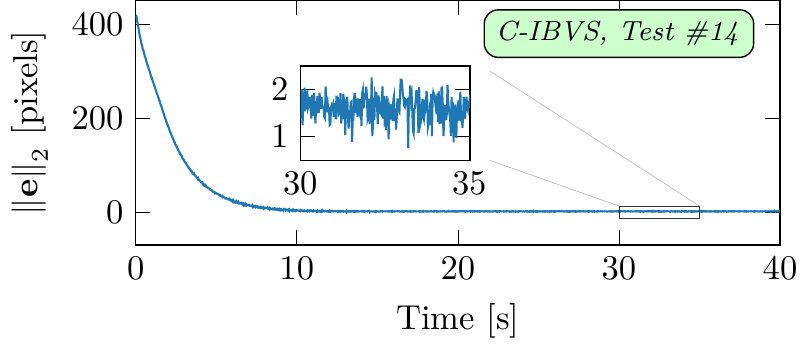}
  \caption{\textit{2D} features error \textit{(C-IBVS)}}
  \label{fig:2DErrors_S113_classic_ibvs_2pixels_errors}
\end{subfigure}
\par\medskip 
\begin{subfigure}{.25\textwidth}
  \centering
  \hspace{-0.02in}\includegraphics[width=1.01\columnwidth,height=0.85in]{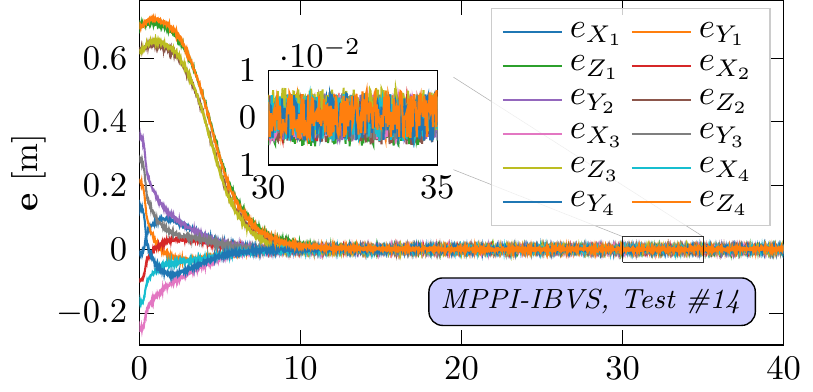}
  \includegraphics[width=1\columnwidth,height=0.75in]{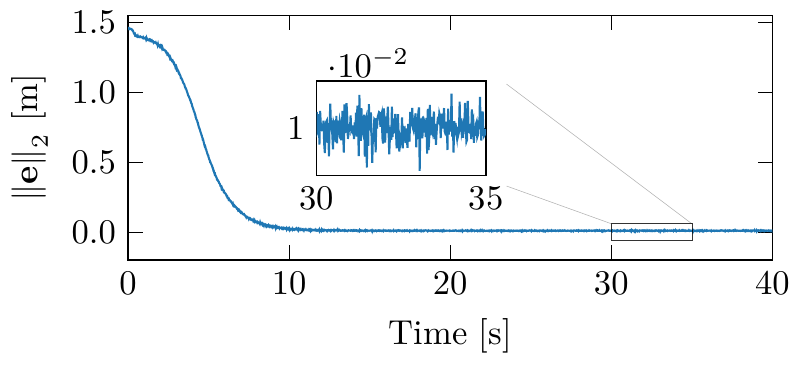}
  \caption{\textit{3D} points error \textit{(MPPI-IBVS)}}
  \label{fig:3DErrors_S113_mppi_ibvs_2pixels_errors}
\end{subfigure}%
\begin{subfigure}{.25\textwidth}
  \centering
  \hspace{-0.02in}\includegraphics[width=1.01\columnwidth,height=0.85in]{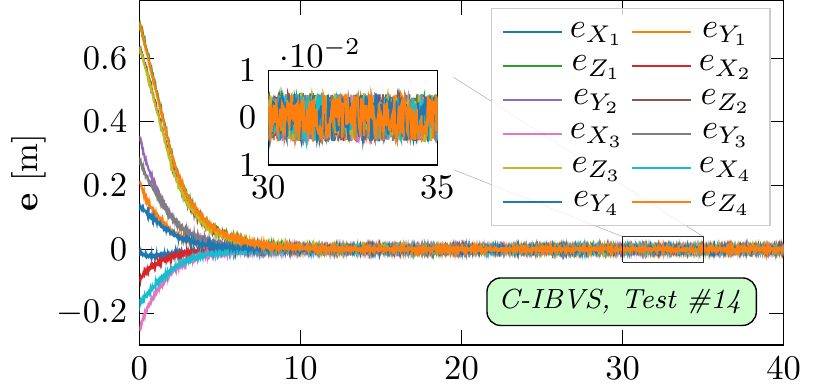}
  \includegraphics[width=1\columnwidth,height=0.75in]{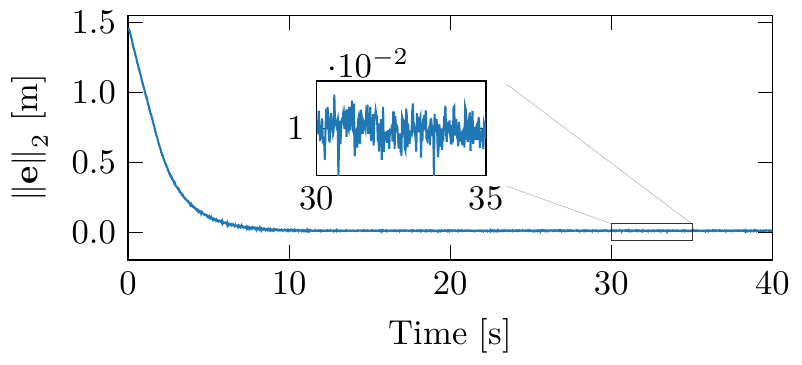}
  \caption{\textit{3D} points error \textit{(C-IBVS)}}
  \label{fig:3DErrors_S113_classic_ibvs_2pixels_errors}
\end{subfigure}
\caption{Simulation results of \textit{MPPI-IBVS} and \textit{C-IBVS} for the positioning task \#113 taken into account the measurement noise described in Test \#14.}
\label{fig:s113_ibvs_2pixels_errors_test14}
\end{figure}
\begin{figure}
\centering
\begin{subfigure}{.25\textwidth}
  \centering
  \includegraphics[width=1\columnwidth,height=1.4in]{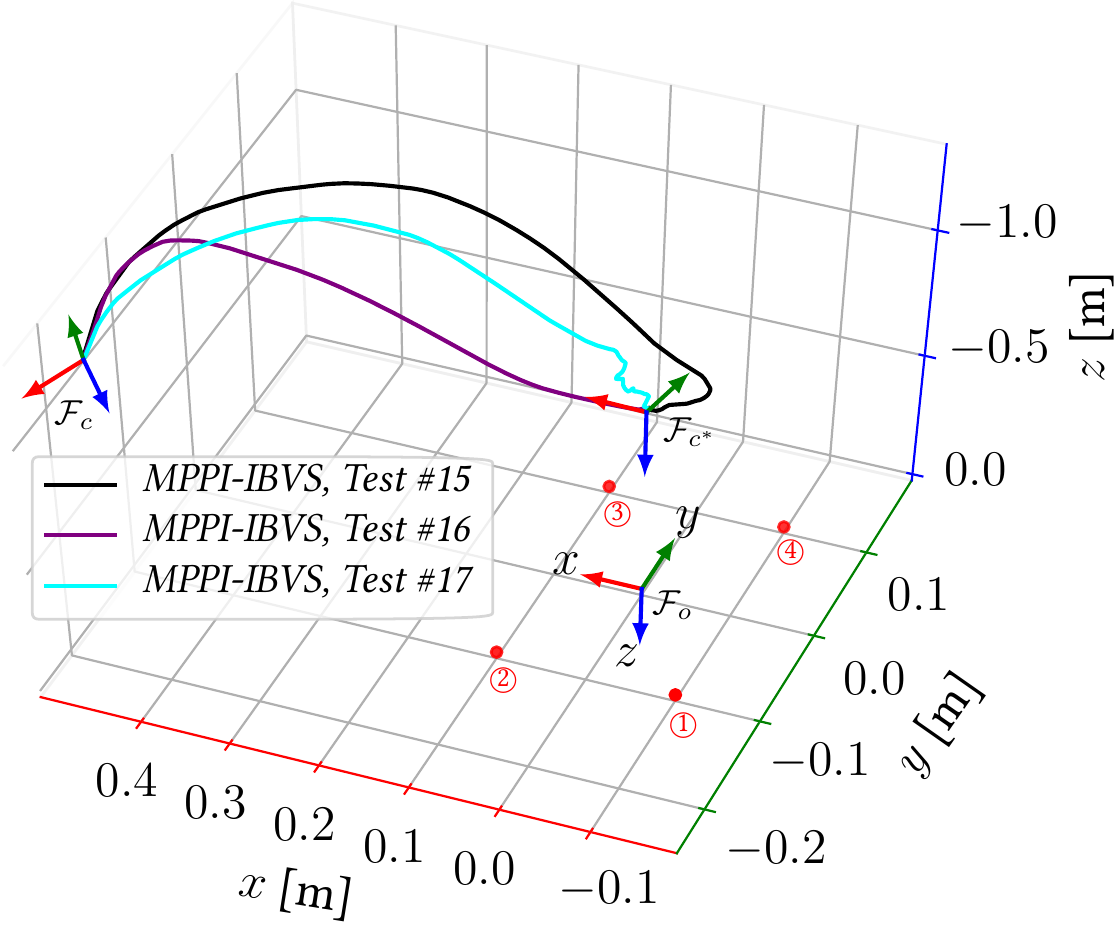}
  \caption{\textit{3D} camera trajectories} 
  \label{fig:3Dtrajectories_S113_ibvs_mppi_camera_errors}
\end{subfigure}%
\begin{subfigure}{.25\textwidth}
  \centering
  \includegraphics[width=1\columnwidth,height=1.4in]{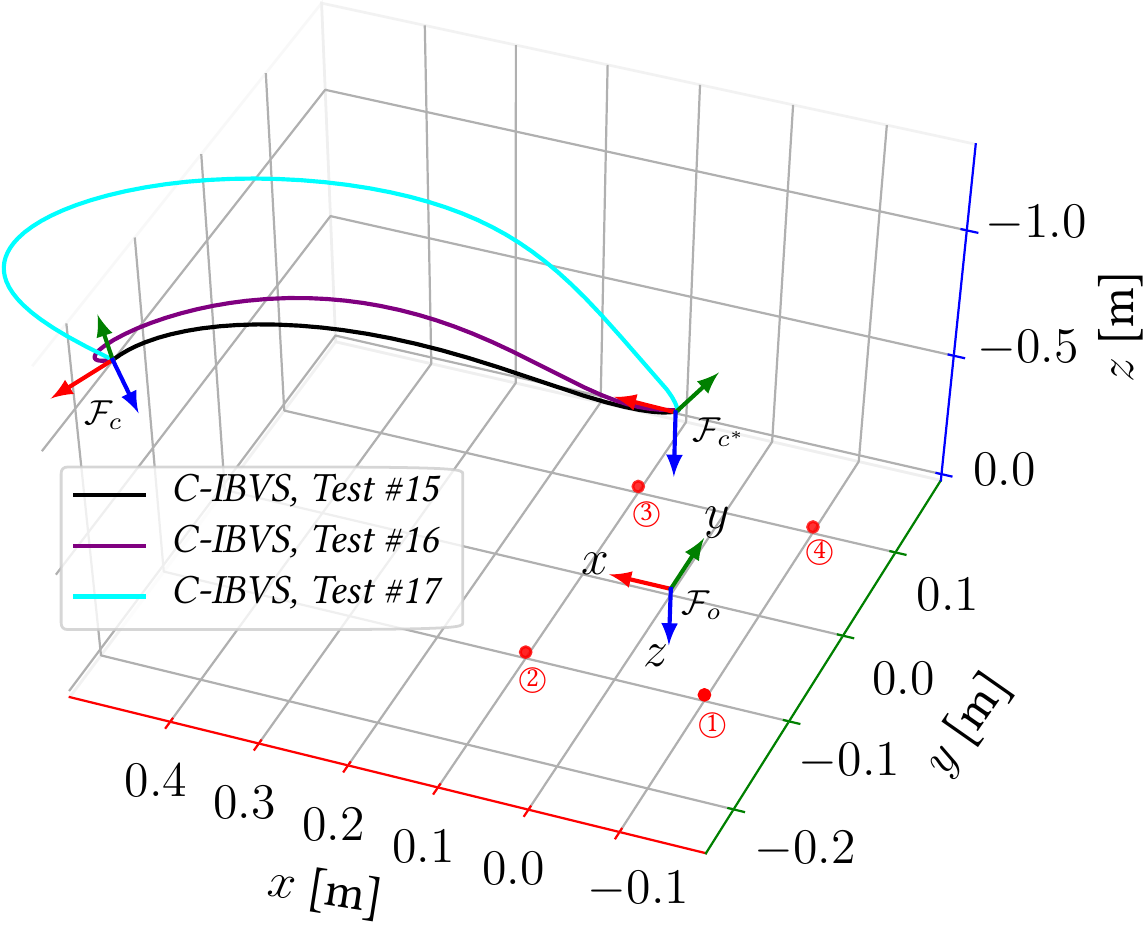}
  \caption{\textit{3D} camera trajectories}
  \label{fig:3Dtrajectories_S113_ibvs_classic_camera_errors}
\end{subfigure}
\par\medskip 
\begin{subfigure}{.25\textwidth}
  \centering
  \includegraphics[width=1\columnwidth,height=1.4in]{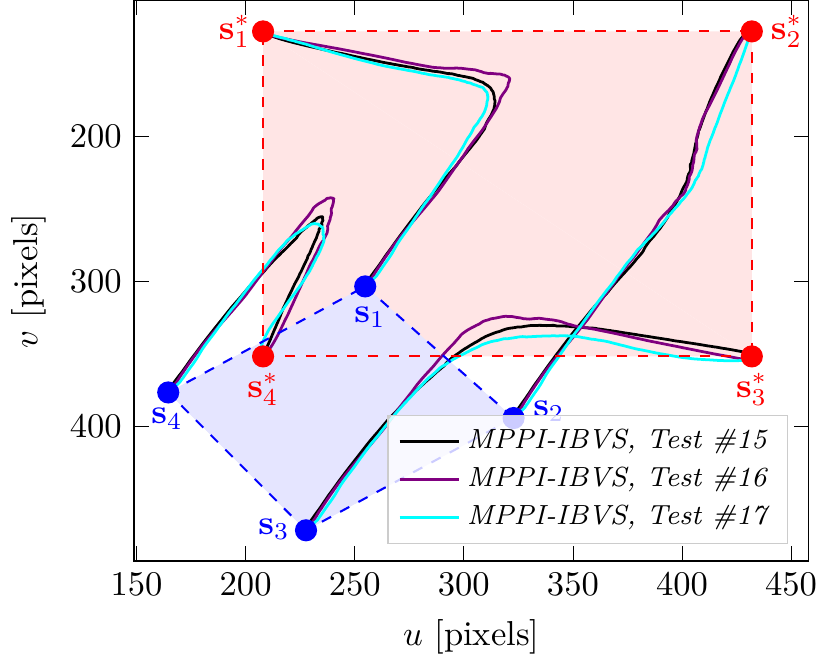}
  \caption{Image points trajectories} 
  \label{fig:2Dtrajectories_S113_ibvs_mppi_camera_errors}
\end{subfigure}%
\begin{subfigure}{.25\textwidth}
  \centering
  \includegraphics[width=1\columnwidth,height=1.4in]{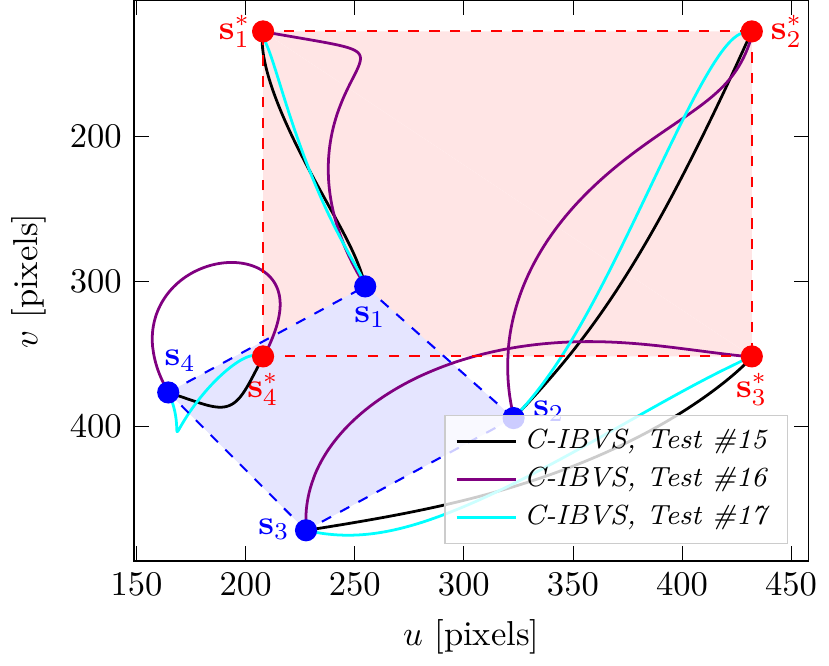}
  \caption{Image points trajectories}
  \label{fig:2Dtrajectories_S113_ibvs_classic_camera_errors}
\end{subfigure}
\par\medskip 
\begin{subfigure}{.25\textwidth}
  \centering
  \includegraphics[width=1\columnwidth,height=0.75in]{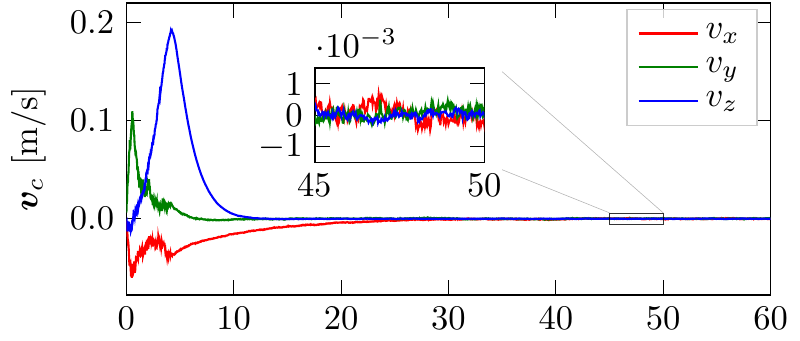}
  \includegraphics[width=1\columnwidth,height=0.9in]{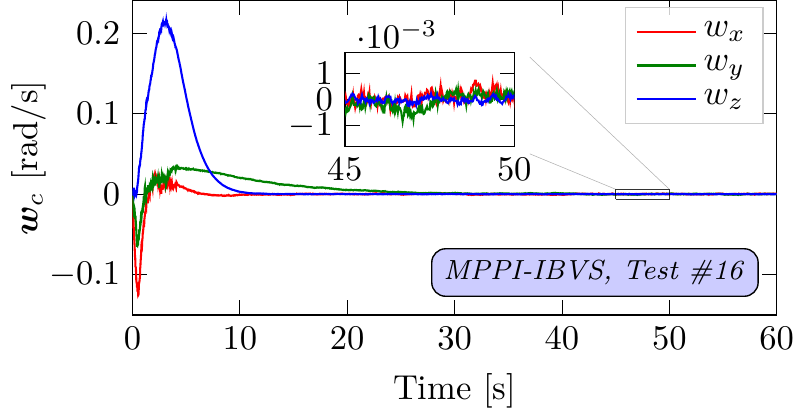}
  \caption{Camera velocity \textit{(MPPI-IBVS)}}
  \label{fig:cameraVelocity_S113_mppi_ibvs_camera_errors}
\end{subfigure}%
\begin{subfigure}{.25\textwidth}
  \centering
  \includegraphics[width=1\columnwidth,height=0.75in]{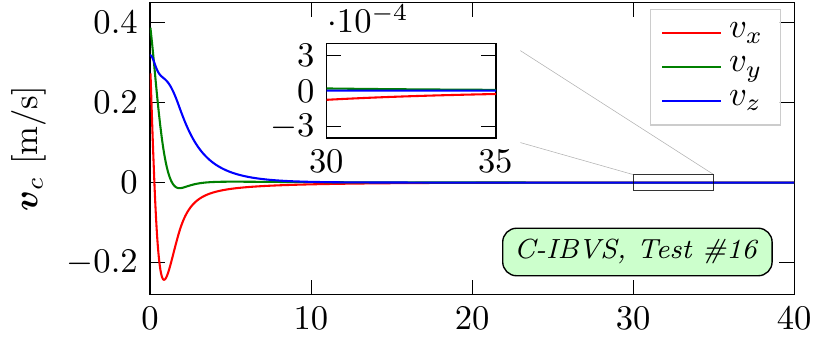}
  \includegraphics[width=1\columnwidth,height=0.9in]{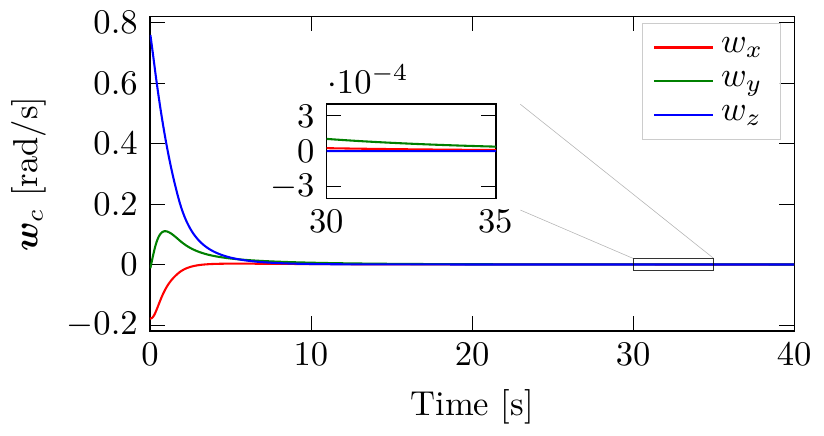}
  \caption{Camera velocity \textit{(C-IBVS)}}
  \label{fig:cameraVelocity_S113_classic_ibvs_camera_errors}
\end{subfigure}
\par\medskip 
\begin{subfigure}{.25\textwidth}
  \centering
  \includegraphics[width=1\columnwidth,height=0.9in]{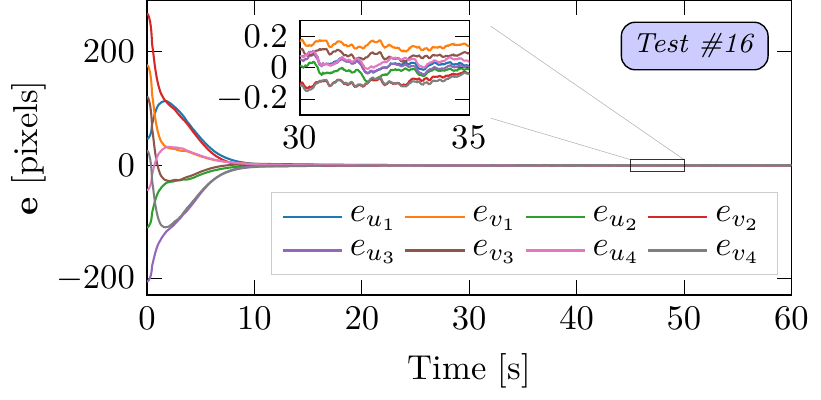}
  \caption{\textit{2D} features error \textit{(MPPI-IBVS)}}
  \label{fig:2DErrors_S113_mppi_ibvs_camera_errors}
\end{subfigure}%
\begin{subfigure}{.25\textwidth}
  \centering
  \includegraphics[width=1\columnwidth,height=0.9in]{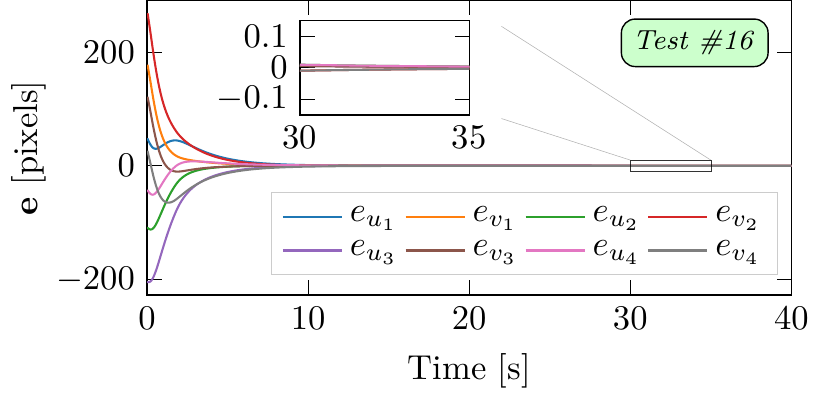}
  \caption{\textit{2D} features error \textit{(C-IBVS)}}
  \label{fig:2DErrors_S113_classic_ibvs_camera_errors}
\end{subfigure}%
\caption{Simulation results of \textit{MPPI-IBVS} and \textit{C-IBVS} for the positioning task \#113 taken into account the camera parameters errors described in Tests \#15, \#16, and \#17.}
\label{fig:s113_ibvs_camera_errors_test15_16_17}
\end{figure}
As can be seen in Table~\ref{table:intensiveSimulationTable}, our proposed control scheme behaves perfectly as there exists a maximum of \num{17}-$\mathcal{R}_{\text{JL}}$ failure cases that occurred in Test \#17, 
demonstrating its robustness in the presence of large errors in camera parameters and measurements, in addition to ensuring the convergence of the system.

Figure~\ref{fig:s113_ibvs_2pixels_errors_test14} demonstrates a comparison between the behavior of \textit{MPPI-IBVS} and \textit{C-IBVS} in terms of the measurement noise described in Test \#14. 
We can observe that \textit{MPPI-IBVS} follows almost exactly the same trajectories in the image, while it provides a bit long camera trajectory compared to the nominal case with very low fluctuations in the camera velocity components, as illustrated in Figs.~\ref{fig:2Dtrajectories_S113_ibvs_2pixels_errors}, \ref{fig:3Dtrajectory_S113_ibvs_2pixels_errors}, and \ref{fig:cameraVelocity_S113_mppi_ibvs_2pixels_errors}. 
On the other side, in the case of \textit{C-IBVS}, in spite of both the points' and \textit{3D} camera trajectories follow almost straight lines, the visual servoing is unstable due to high oscillations in the camera velocity components produced by the classical control law (see Fig.~\ref{fig:cameraVelocity_S113_classic_ibvs_2pixels_errors}).
Figures \ref{fig:2DErrors_S113_mppi_ibvs_2pixels_errors} to \ref{fig:3DErrors_S113_classic_ibvs_2pixels_errors} show the evolution of the error $\mathbf{e}$ of both image features and \textit{3D} points for the two control schemes, as well as their corresponding error norm ${\lVert \mathbf{e} \lVert}_2$. 
It should noticed that the maximum steady-state features error norm, for both schemes, is less than \SI{2}{\pixels} which is significantly lower than the added noise, while the steady-state \textit{3D} points error norm is mainly centered around \SI{1}{\cm}.

In Fig.~\ref{fig:s113_ibvs_camera_errors_test15_16_17}, we present the behavior of both control schemes with respect to camera calibration errors. 
As can be seen in Figs.~\ref{fig:3Dtrajectories_S113_ibvs_mppi_camera_errors} and \ref{fig:3Dtrajectories_S113_ibvs_classic_camera_errors}, the behavior of our proposed control scheme in Cartesian space is more satisfactory than that of \textit{C-IBVS}, especially in the case where the combined errors are considered (namely, Test \#17). 
Furthermore, the more interesting property is that those calibration errors have no remarkable effect on both: (i) the followed trajectories in the image (see Fig.~\ref{fig:2Dtrajectories_S113_ibvs_mppi_camera_errors}), (ii) the quality of optimal control input of the camera velocity screw as illustrated in Fig.~\ref{fig:cameraVelocity_S113_mppi_ibvs_camera_errors}, and (iii) the convergence rate (see Fig.~\ref{fig:2DErrors_S113_mppi_ibvs_camera_errors}).  
In practice, this clearly means that \textit{MPPI-IBVS} is largely compatible with a rough calibration (i.e., no need for an accurate calibration step), with sustaining a wide margin of the overall stability of the system.
In addition, concerning \textit{C-IBVS}, it can be seen in Figs.~\ref{fig:3Dtrajectories_S113_ibvs_classic_camera_errors} and \ref{fig:2Dtrajectories_S113_ibvs_classic_camera_errors} that neither the \textit{3D} camera trajectory nor the image points trajectories follow straight lines owning to bad calibration, which might increase the possibility of the visual features leaving the camera's FoV. Sequentially, the evolution of the error is no longer perfectly exponential (see Fig.~\ref{fig:2DErrors_S113_classic_ibvs_camera_errors}).

\begin{figure}[!ht]
\centering
\begin{subfigure}{.25\textwidth}
  \centering
  \includegraphics[width=1\columnwidth,height=1.4in]{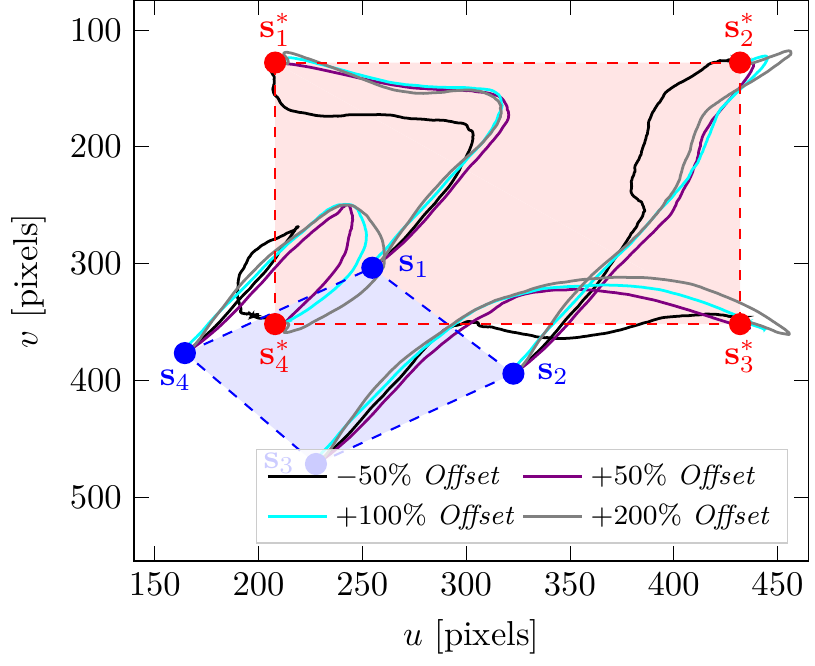}
  \caption{Image points trajectories} 
  \label{fig:2Dtrajectories_S113_Z0_Estimation}
\end{subfigure}%
\begin{subfigure}{.25\textwidth}
  \centering
  \includegraphics[width=1\columnwidth,height=1.4in]{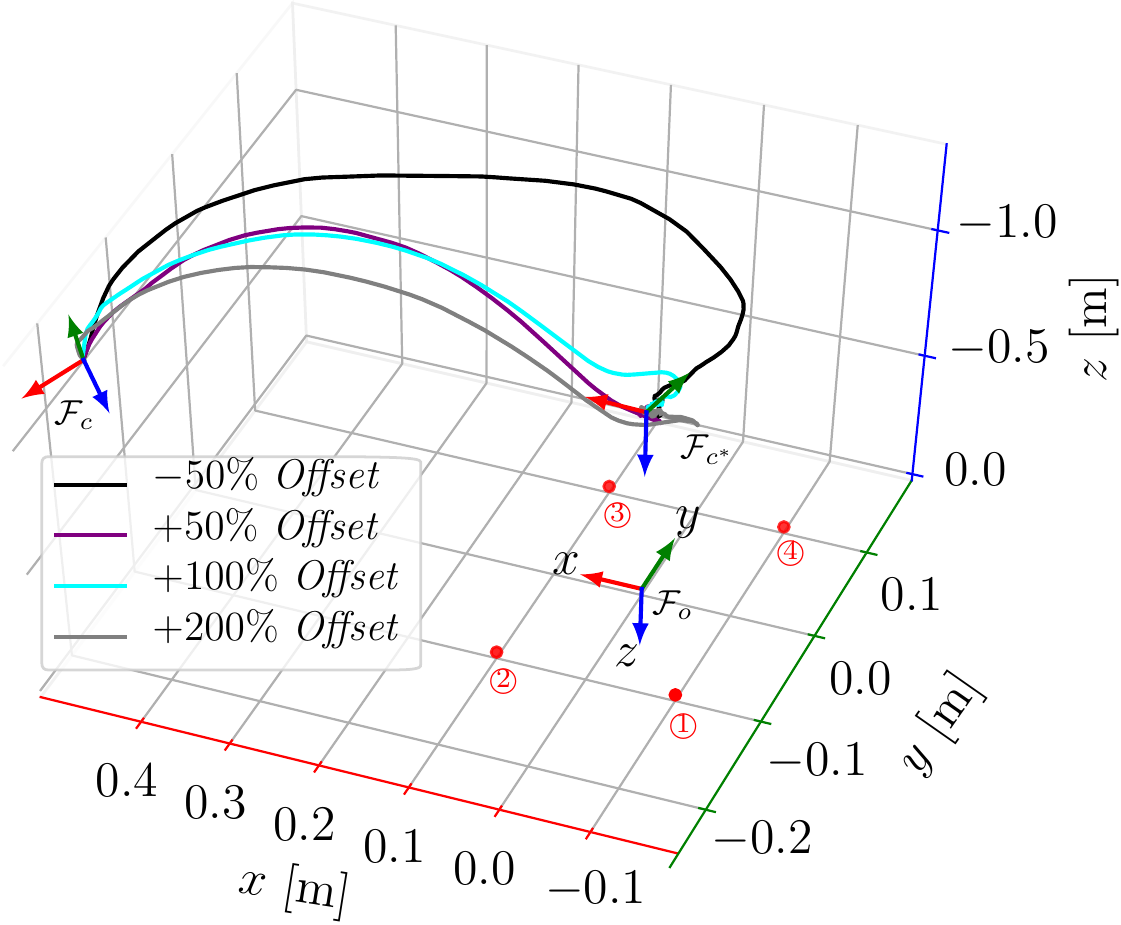}
  \caption{\textit{3D} camera trajectories}
  \label{fig:3Dtrajectory_S113_Z0_Estimation}
\end{subfigure}
\caption{Simulation results of \textit{MPPI-IBVS} for task \#113 taken into account an offset in the estimation of $Z_i$.}
\label{fig:S113_Z0_Estimation}
\end{figure}

Figure~\ref{fig:S113_Z0_Estimation} demonstrates the robustness of our proposed control scheme with respect to a bad estimation of the object depths $Z_i(t)$, starting from having an offset of $-$\SI{50}{\%} (i.e., half the true value) and ending with an offset of $+$\SI{200}{\%} (i.e. three times the true value). Moreover, the obtained results clearly demonstrate the fact that \textit{MPPI-VS} does not require an accurate mathematical model of the system dynamics; i.e., in our case, an approximate interaction matrix is sufficient to compute the optimal control sequence. 

\subsubsection{\textcolor{blue-violet}{\textbf{\textit{3DVS} Control Schemes}}}
\begin{figure}[!ht]
\centering
\begin{subfigure}{.25\textwidth}
  \centering
  \includegraphics[width=1\columnwidth,height=1.4in]{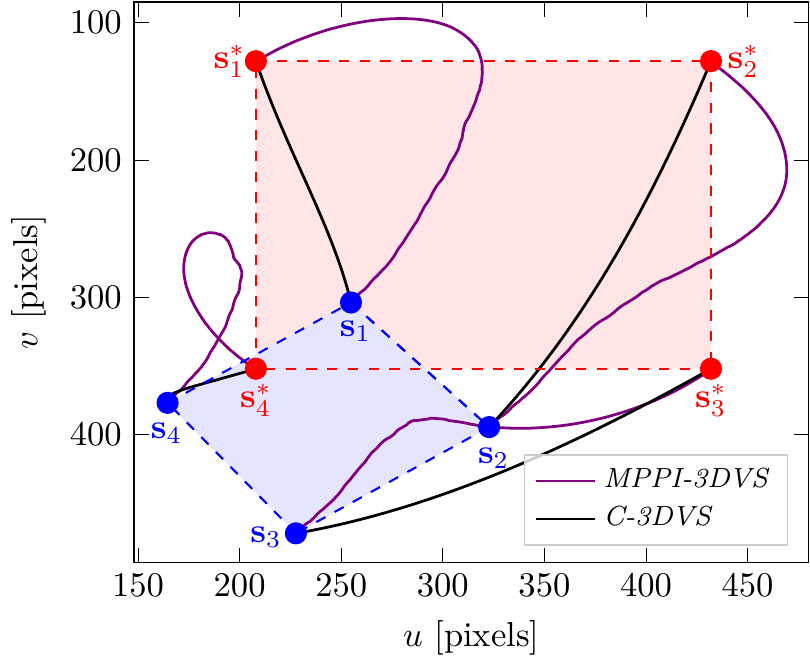}
  \caption{Image points trajectories} 
  \label{fig:2Dtrajectories_S113_3bvs}
\end{subfigure}%
\begin{subfigure}{.25\textwidth}
  \centering
  \includegraphics[width=1\columnwidth,height=1.4in]{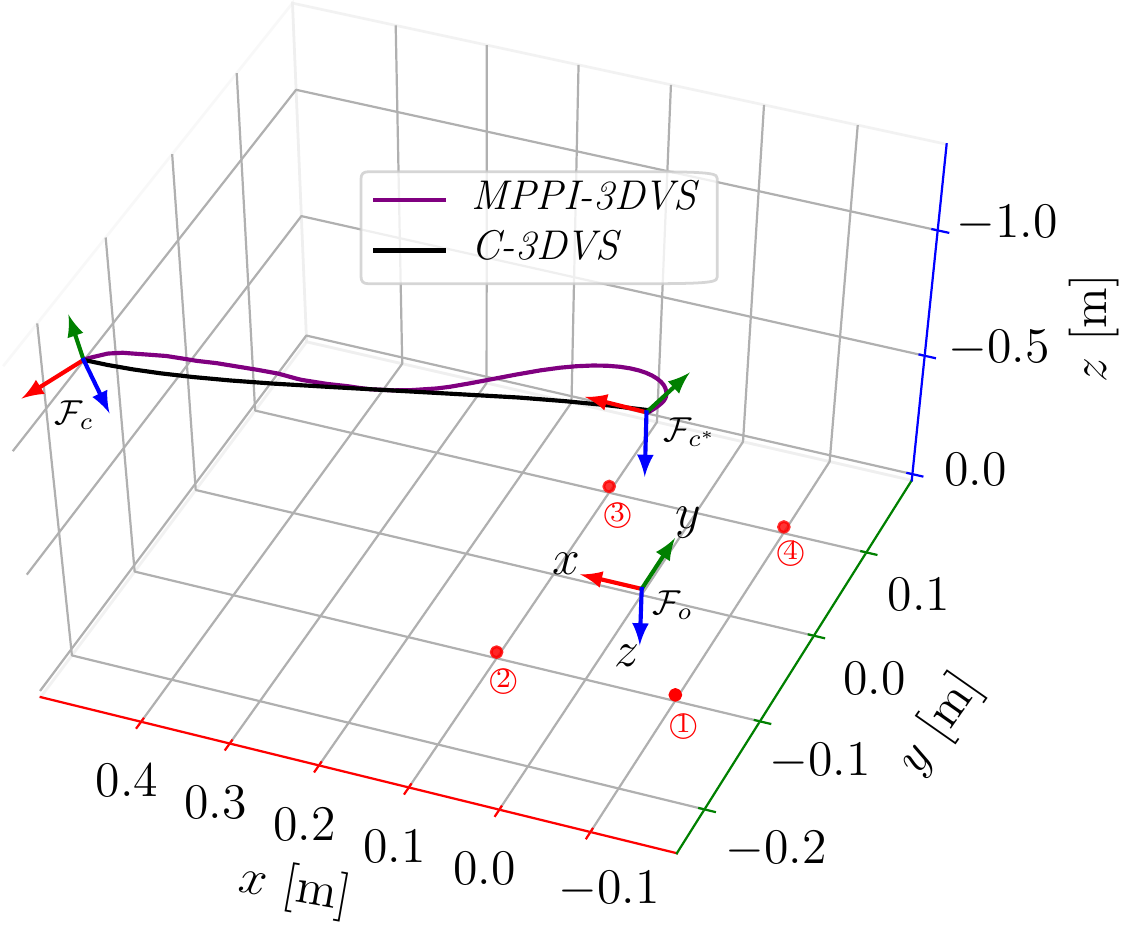}
  \caption{\textit{3D} camera trajectories}
  \label{fig:3Dtrajectory_S113_3dvs}
\end{subfigure}
\par\medskip 
\begin{subfigure}{.25\textwidth}
  \centering
  \includegraphics[width=1\columnwidth,height=0.7in]{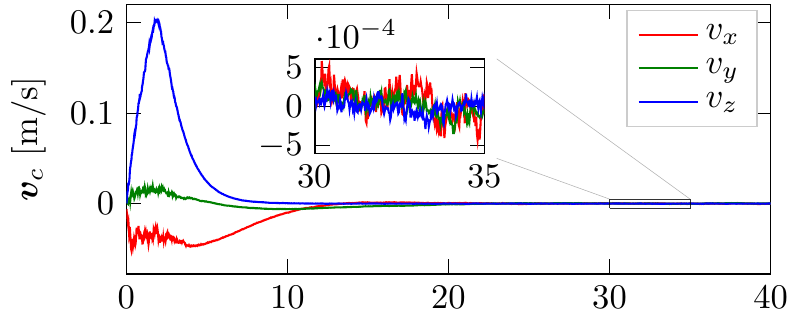}
  \includegraphics[width=1\columnwidth,height=0.85in]{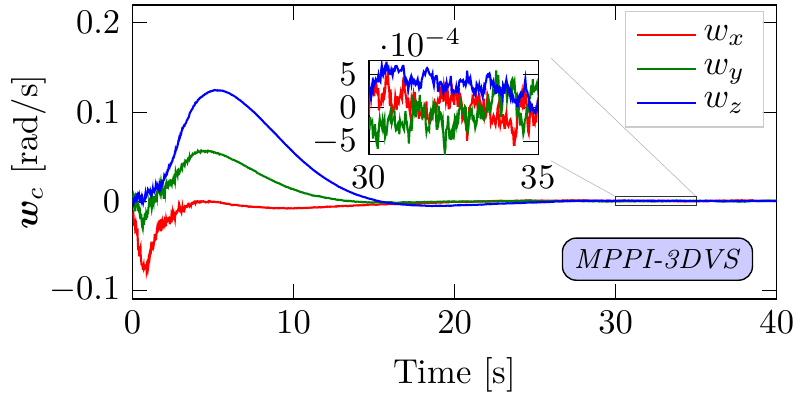}
  \caption{Camera velocity \textit{(MPPI-3DVS)}}
  \label{fig:cameraVelocity_S113_mppi_3dvs}
\end{subfigure}%
\begin{subfigure}{.25\textwidth}
  \centering
  \includegraphics[width=1\columnwidth,height=0.7in]{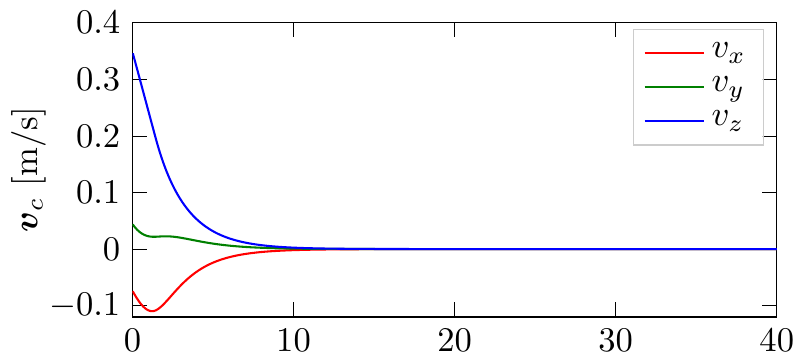}
  \includegraphics[width=1\columnwidth,height=0.85in]{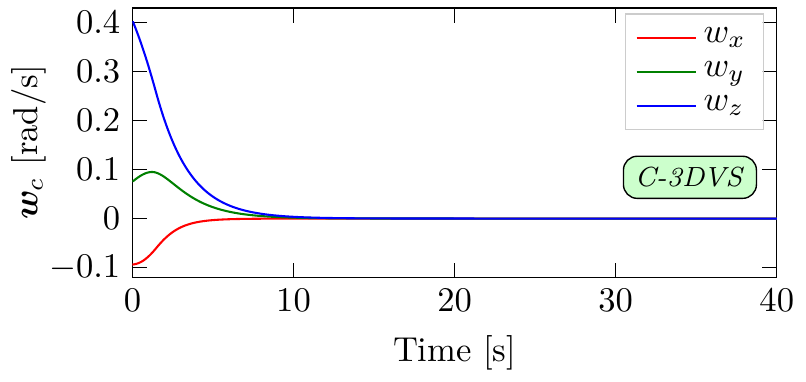}
  \caption{Camera velocity \textit{(C-3DVS)}}
  \label{fig:cameraVelocity_S113_classic_3dvs}
\end{subfigure}
\par\medskip 
\begin{subfigure}{.25\textwidth}
  \centering
  \includegraphics[width=1\columnwidth,height=0.95in]{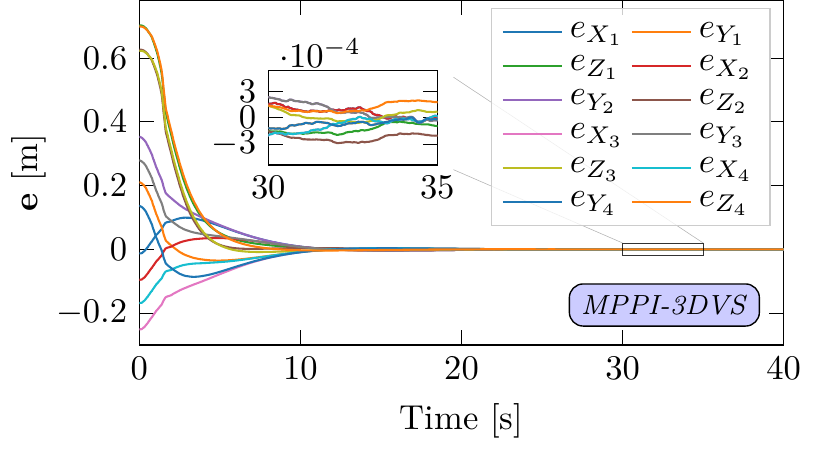}
  \caption{\textit{3D} points error \textit{(MPPI-3DVS)}}
  \label{fig:3DErrors_S113_mppi_3dvs}
\end{subfigure}%
\begin{subfigure}{.25\textwidth}
  \centering
  \includegraphics[width=1\columnwidth,height=0.95in]{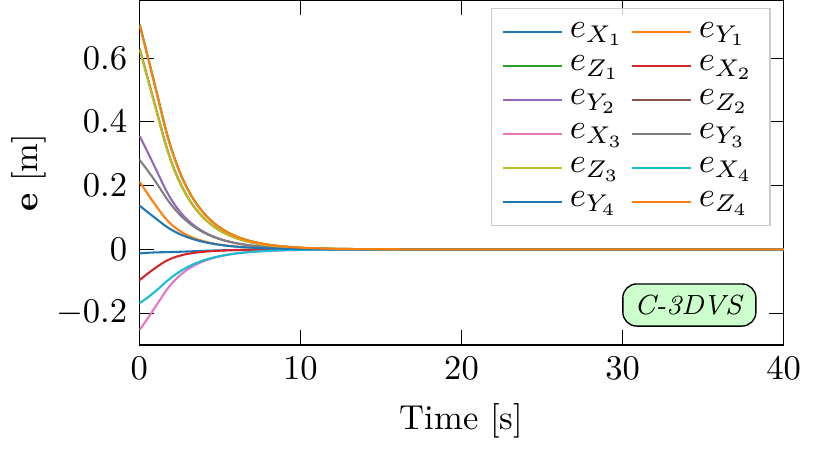}
  \caption{\textit{3D} points error \textit{(C-3DVS)}}
  \label{fig:3DErrors_S113_classic_3dvs}
\end{subfigure}%
\caption{Simulation results of \textit{MPPI-3DVS} and \textit{C-3DVS} for the positioning task \#113.}
\label{fig:s113_3dvs_mppi_classic}
\end{figure}

In this work, \textit{3DVS} has been mainly utilized by the \textit{MPPI} algorithm either for predicting the future evolution of the object depths $Z_i(t)$ at each iteration in case of \textit{MPPI-IBVS} or for coping with the visibility constraints of \textit{MPPI-PBVS} as previously described in (\ref{eq:18}).
Nevertheless, four intensive simulations are carried out in Tests \#19 to \#22 for assessing the performance of both \textit{MPPI-3DVS} and \textit{C-3DVS} (with $\lambda_{\mathbf{s}}=0.5)$, considering the two desired camera configurations and the \textit{MPPI-3DVS} parameters listed in Table~\ref{table:SysParameters}.  
The simulations show that both control strategies give very satisfactory results, where the maximum failure cases were \num{7} (\num{6} for  $\mathcal{R}_{\text{JL}}$ and only one for $\mathcal{P}_{\text{out}}$) occurred in Test \#22.  
Figure~\ref{fig:s113_3dvs_mppi_classic} gives an example illustrating the behavior of both control schemes.
Despite the \textit{2D} features in case of \textit{MPPI-3DVS} follow more complex trajectories compared to that in \textit{MPPI-IBVS} (see Figs.~\ref{fig:2Dtrajectories_S113_3bvs} and ~\ref{fig:2Dtrajectories_S113_ibvs}), the fluctuations of the camera velocity components and the evolution of the error are significantly lower (where lower is better) in the steady-state, as can be seen in Figs.~\ref{fig:cameraVelocity_S113_mppi_3dvs} and \ref{fig:3DErrors_S113_mppi_3dvs}. While \textit{C-3DVS} provides almost straight-line trajectories in the image and \textit{3D} space as in \textit{C-IBVS}.
\subsubsection{\textcolor{capri}{\textbf{Unconstrained \textit{MPPI-PBVS} V.S. \textit{C-PBVS}}}}
In the last six tests, intensive simulations are conducted to verify the performance of our proposed \textit{PBVS} control scheme.
In the first three tests (namely, from Test \#23 to \#25), we studied the behavior of \textit{MPPI-PBVS} without applying the visibility constraints, together with the performance of the classical scheme. 
The obtained results demonstrate that both control strategies produce control input ultimately leading to that some visual features leave the camera’s FoV (i.e., $\mathcal{P}_{\text{out}}\neq 0$). 
Moreover, it is interesting to notice that (i) \textit{C-PBVS} provides more satisfactory results than \textit{MPPI-PBVS} as our proposed scheme has in total \num{45}-$\mathcal{P}_{\text{out}}$ failure cases, in which at least one point leaves the FoV, compared to a total of \num{26}-$\mathcal{P}_{\text{out}}$ cases obtained by \textit{C-PBVS} in Test \#24, and (ii) the robot joint limits are totally avoided (i.e., there exist no-$\mathcal{R}_{\text{JL}}$ failure cases). 

\subsubsection{\textcolor{dogwoodrose}{\textbf{Constrained \textit{MPPI-PBVS}}}} 
Now, it is the time to assess the performance of \textit{MPPI-PBVS} taken into consideration the two methods proposed in (\ref{eq:17}) and (\ref{eq:18}) to cope with the visibility constraints.
Tests from \#26 to \#28 demonstrate the validness and effectiveness of our proposed methods in handling visibility constraints, regardless of which desired camera configuration is considered (i.e., whether ${ }^{o}\boldsymbol{P}_{c_1}^{\ast}$ or ${ }^{o}\boldsymbol{P}_{c_2}^{\ast}$), as there exist neither $\mathcal{R}_{\text{JL}}$ nor $\mathcal{P}_{\text{out}}$ failure cases.

\begin{figure}[!ht]
\centering
\begin{subfigure}{.25\textwidth}
  \centering
  \includegraphics[width=1\columnwidth,height=1.4in]{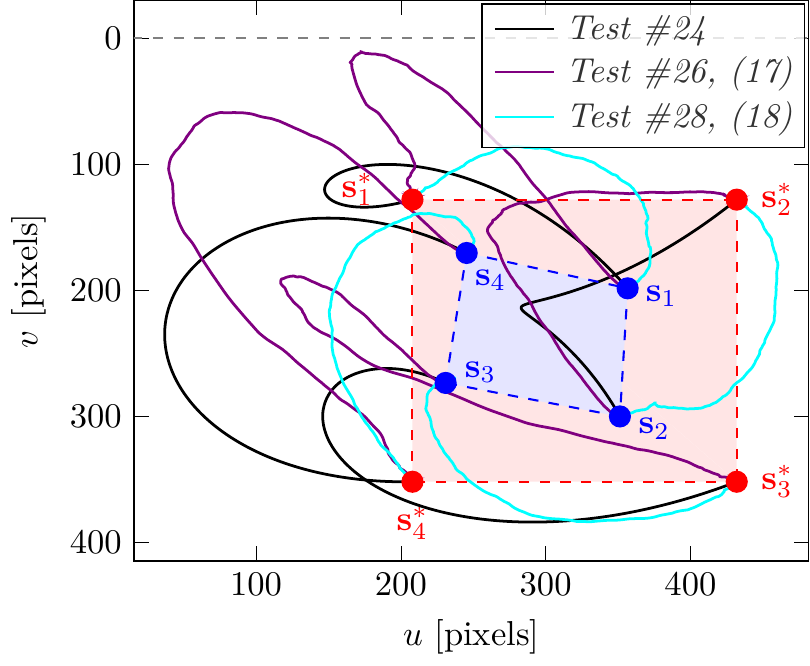}
  \caption{Image points trajectories} 
  \label{fig:2Dtrajectories_S15_pbvs}
\end{subfigure}%
\begin{subfigure}{.25\textwidth}
  \centering
  \includegraphics[width=1\columnwidth,height=1.4in]{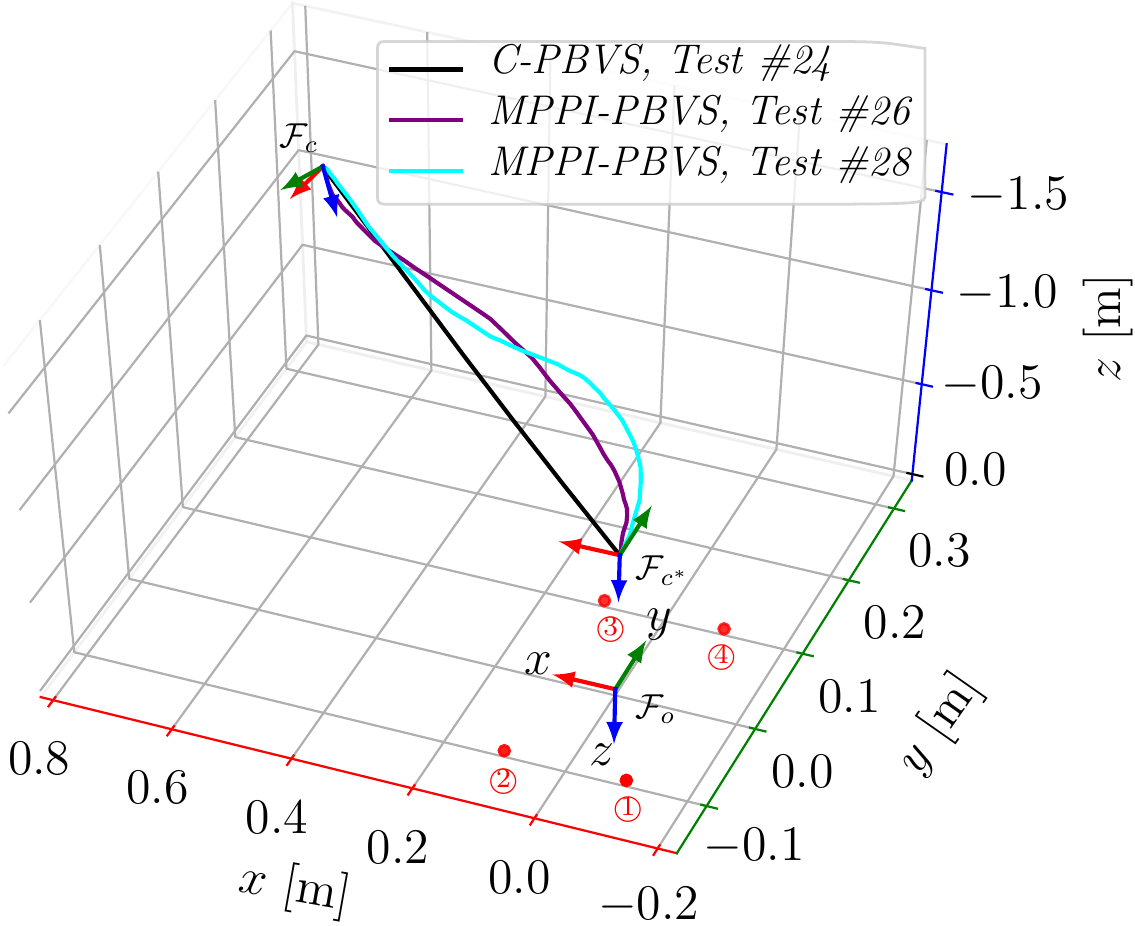}
  \caption{\textit{3D} camera trajectories}
  \label{fig:3Dtrajectory_S15_pbvs}
\end{subfigure}
\par\medskip 
\begin{subfigure}{.25\textwidth}
  \centering
  \includegraphics[width=1\columnwidth,height=0.75in]{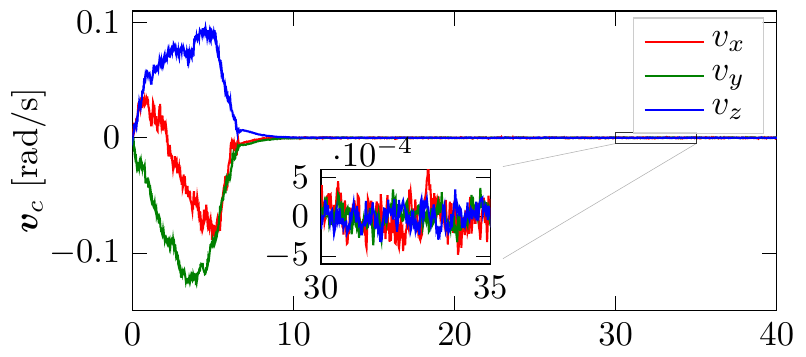}
  \includegraphics[width=1\columnwidth,height=0.85in]{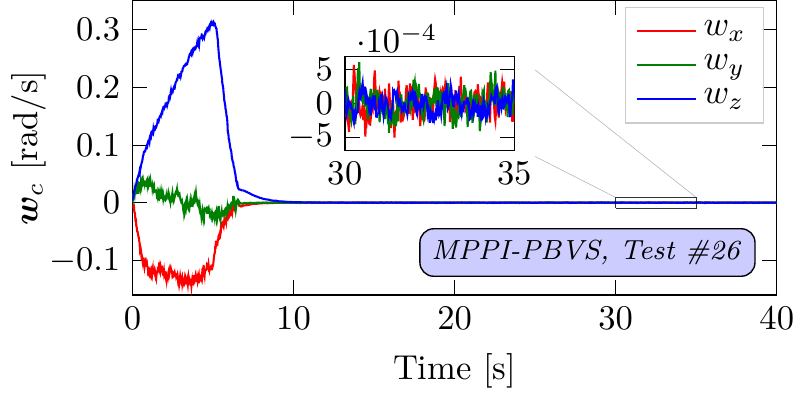}
  \caption{Camera velocity \textit{(MPPI-PBVS)}}
  \label{fig:cameraVelocity_S15_pbvs_mppi}
\end{subfigure}%
\begin{subfigure}{.25\textwidth}
  \centering
  \includegraphics[width=1\columnwidth,height=0.75in]{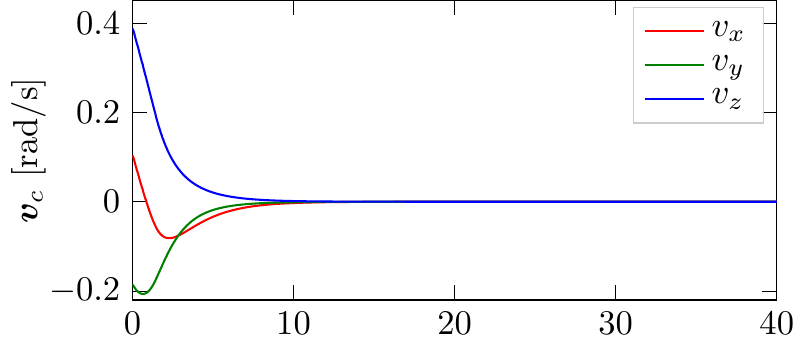}
  \includegraphics[width=1\columnwidth,height=0.85in]{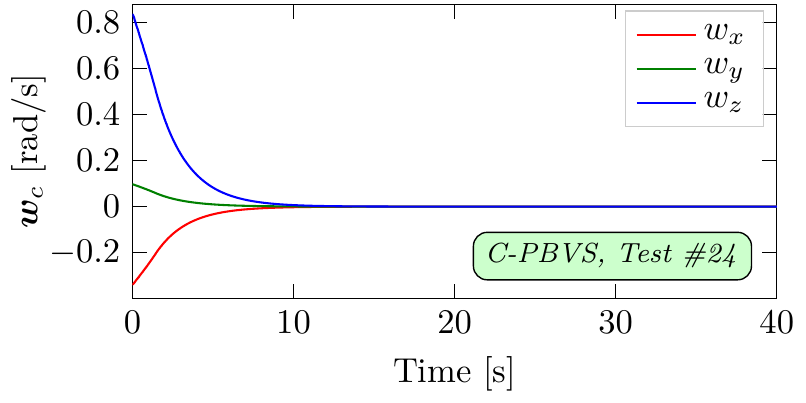}
  \caption{Camera velocity \textit{(C-PBVS)}}
  \label{fig:cameraVelocity_S15_pbvs_classic}
\end{subfigure}%
\par\medskip 
\begin{subfigure}{.25\textwidth}
  \centering
  \includegraphics[width=1\columnwidth,height=0.9in]{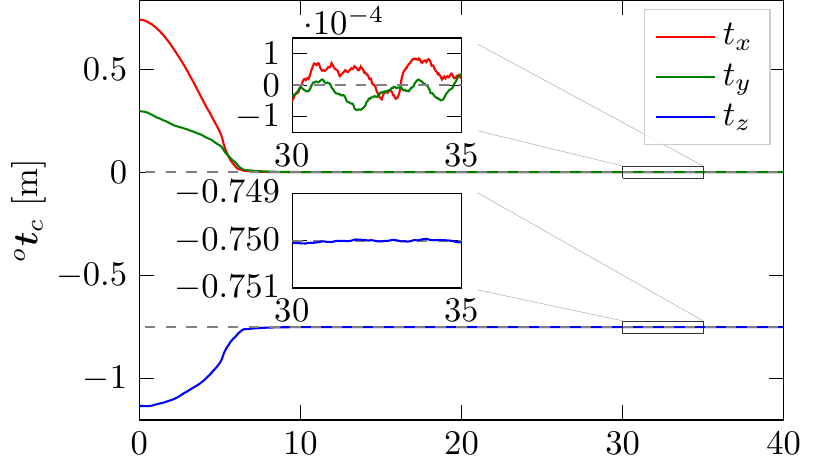}
  \includegraphics[width=1\columnwidth,height=0.85in]{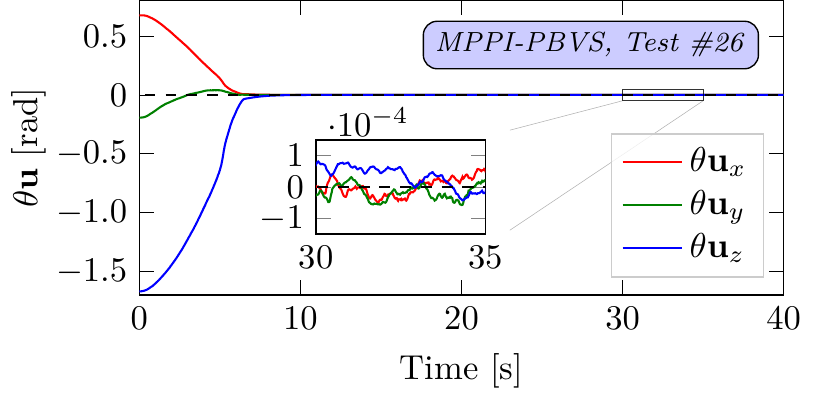}
  \caption{Camera pose ${ }^{o}\boldsymbol{P}_{c}$ in $\mathcal{F}_{o}$}
  \label{fig:poseErrors_S15_mppi_3dvs}
\end{subfigure}%
\begin{subfigure}{.25\textwidth}
  \centering
  \includegraphics[width=1\columnwidth,height=0.8in]{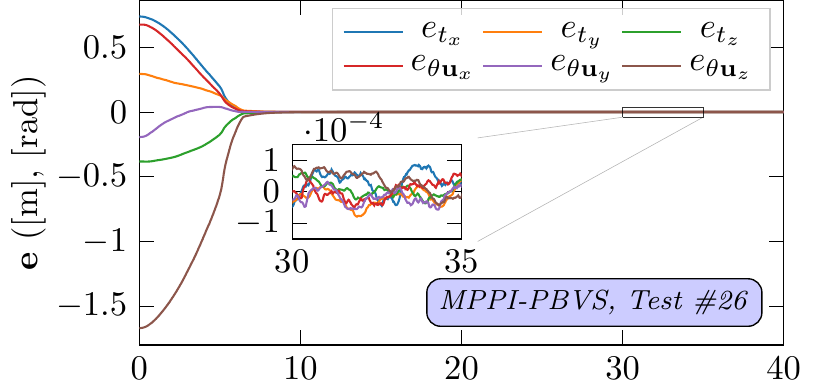}
  \includegraphics[width=1\columnwidth,height=0.92in]{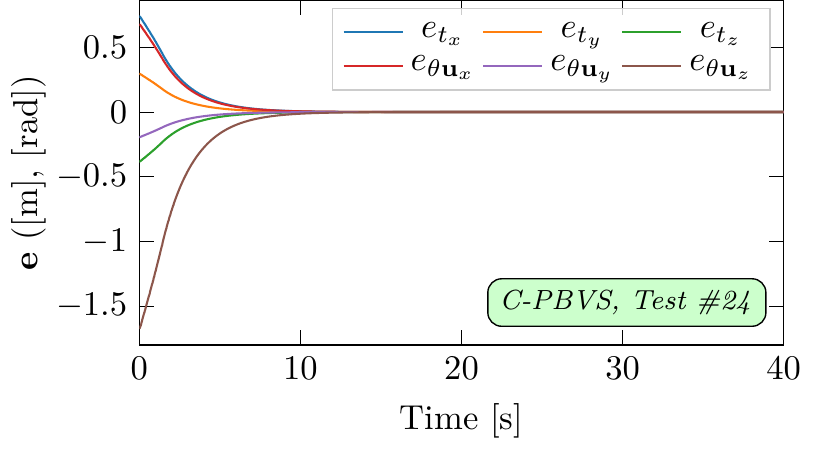}
  \caption{Camera pose error $\mathbf{e}$ in $\mathcal{F}_{c^{*}}$}
  \label{fig:poseErrors_S15_classic_pdvs}
\end{subfigure}%
\caption{Simulation results of \textit{MPPI-PBVS} and \textit{C-PBVS} for the positioning task \#15 that has an initial camera configuration of $\left(\left[0.74, 0.29 , -1.13\right], \left[38.69, -11.09, -95.65\right]\right)^T$ in ([\si{\metre}], [\si{\deg}]), considering ${ }^{o}\boldsymbol{P}_{c_1}^{\ast}$ as the desired configuration.}
\label{fig:s15_pdvs_mppi_classic}
\end{figure}

Figure~\ref{fig:s15_pdvs_mppi_classic} illustrates the behavior of \textit{MPPI-PBVS} based on (\ref{eq:17}) and (\ref{eq:18}) for the servoing task \#15. 
In such a configuration, only \textit{C-IBVS} successfully fulfills the servoing task as the four points always remain in the image as shown in Fig.~\ref{fig:2Dtrajectories_S15_pbvs}, while it is observed that the unconstrained \textit{MPPI-PBVS} that presented in Test \#23 causes failure of servoing since three points leave the image (namely, $\mathbf{s}_1, \mathbf{s}_2$, and $\mathbf{s}_4$).
However, the visibility constraints have been respected, once one of our proposed methods is involved in the \textit{MPPI} algorithm as the visual features have been forced to stay in the image (see again Fig.~\ref{fig:2Dtrajectories_S15_pbvs}).
It is also observed that despite \textit{MPPI-PBVS} on the basis of (\ref{eq:18}) provides better motion in the image space, the \textit{3D} camera trajectory obtained by applying (\ref{eq:17}) is considerably shorter than that obtained by applying (\ref{eq:18}), as illustrated in Fig.~\ref{fig:3Dtrajectory_S15_pbvs}. 
It is noteworthy that the quality of trajectories obtained by (\ref{eq:18}), particularly in the image plane, and satisfying the visibility constraints are mainly based on the assigned value to the cost weighting of $q(\mathbf{s})$ particularly $w_2$, as previously discussed in Section~\ref{sec:MPPI-VS Control Strategy}.
The impact of changing $w_2$ on the performance of \textit{MPPI-PBVS} is studied in Table~\ref{table:effect_w2}. 
It can be clearly noticed that by assigning low values to $w_2$, our control scheme is unable to constantly satisfying the visibility constraints. For instance, if $w_2$ is assigned to \num{1} instead of \num{150} (the value we have set during our intensive simulations), three points will leave the FoV (i.e., $p_{\text{out}}=3$).
Thus, the higher this value, the better the performance. 
For this reason, it is highly recommended to utilize (\ref{eq:17}) in the prediction algorithm as the constraints have not been violated.
\definecolor{applegreen}{rgb}{0.7, 1, 0.0}
\definecolor{LightCyan}{rgb}{0.88,1,1}
\definecolor{atomictangerine}{rgb}{1.0, 0.6, 0.4}
\begin{table}[!ht]
\caption{Influence of changing $w_2$}
\centering
\begin{tabular}{|c| c||c| c||c| c|}
\hline
 \cellcolor{auburn!15} $w_2$  & \cellcolor{pink!50} $p_{\text{out}}$  &  \cellcolor{auburn!15} $w_2$  & \cellcolor{pink!50} $p_{\text{out}}$ &  \cellcolor{auburn!15} $w_2$ & \cellcolor{pink!50} $p_{\text{out}}$\\
 \hline  
 \hline
 \cellcolor{auburn!15} 1 & \cellcolor{pink!50} 3 & \cellcolor{auburn!15} 2 & \cellcolor{pink!50} 2 & \cellcolor{auburn!15} 5 & \cellcolor{pink!50} 1 \\
 \hline
 \cellcolor{auburn!15} 10 & \cellcolor{pink!50} 1 & \cellcolor{auburn!15} 20 & \cellcolor{pink!50} None &  \cellcolor{auburn!15} 50 & \cellcolor{pink!50} None\\
 \hline
\end{tabular}
\label{table:effect_w2}
\end{table}

Figures~\ref{fig:cameraVelocity_S15_pbvs_mppi} and \ref{fig:cameraVelocity_S15_pbvs_classic} show the evolution of the six components of the camera velocity for the constrained \textit{MPPI-PBVS} based on (\ref{eq:17}) and \textit{C-PBVS}, while the evolution of the error of the camera pose relative to the desired camera frame $\mathcal{F}_{c^{*}}$ is shown in Fig.~\ref{fig:poseErrors_S15_classic_pdvs}.
We can notice that our control scheme achieves a slightly faster convergence; the system converges to the desired configuration within \SI{9.44}{\second} compared to \SI{14.08}{\second} when \textit{C-PBVS} is used. 
Furthermore, its behavior in the steady state is roughly similar to that of \textit{MPPI-3DVS} and, once again, better than that of \textit{MPPI-IBVS}. Finally, the evolution of the camera pose with respect to the object reference frame $\mathcal{F}_{o}$ is shown in Fig.~\ref{fig:poseErrors_S15_mppi_3dvs}.  
It is quite interesting to observe that the positioning error precision obtained in this servoing task is less than \SI{0.1}{\milli\metre} for the translation and \SI{0.01}{\degree} for the rotation.

\subsubsection{\textbf{Reaching Local Minimum}}
Concerning all tests where \textit{MPPI} is deployed, it should be clearly observed from Table~\ref{table:intensiveSimulationTable} that the controller performs perfectly without the camera reaching a local minimum (i.e., $R_{\text{LM}}=0$ for all tests).
We noticed during our intensive simulations that the most significant parameter for avoiding reaching a local minimum or crossing a singularity of the interaction matrix is the inverse temperature $\lambda$, as $\lambda$ determines how tightly peaked the optimal distribution is. 
In other words, for instance, in the case of \textit{MPPI-IBVS}, we observed that low values of $\lambda$ result in many trajectories being rejected due to their costs are too high. 
In addition, low values lead to empirically slow convergence to the desired configuration, with a slightly fluctuating motion of the camera. 

\subsubsection{\textbf{MPPI-VS Convergence Time}} 

\begin{figure}[!ht]
\centering
\begin{subfigure}{.25\textwidth}
  \centering
  \includegraphics[width=0.95\columnwidth,height=1.3in]{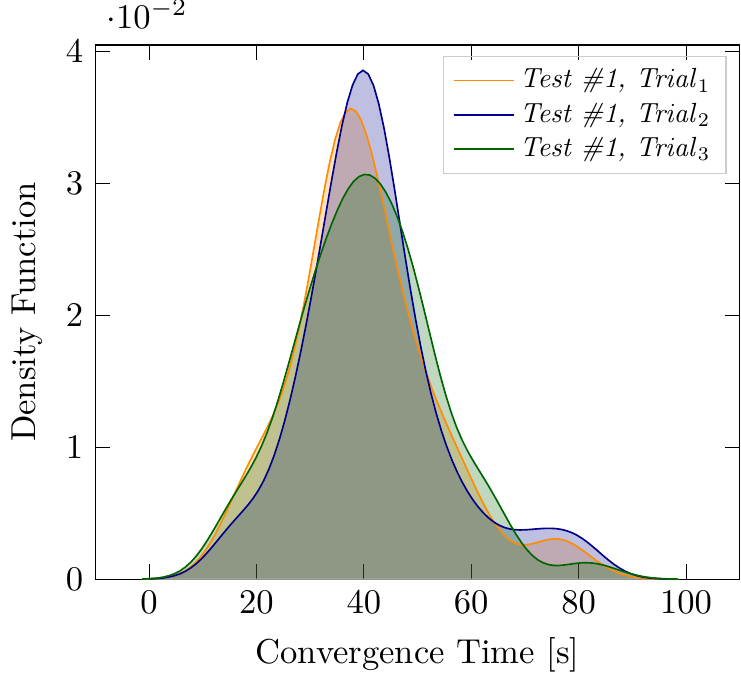}
  \caption{Three trials of Test \#1} 
  \label{fig:density_3trials_Convergence-Time}
\end{subfigure}%
\begin{subfigure}{.25\textwidth}
  \centering
  \includegraphics[width=0.95\columnwidth,height=1.3in]{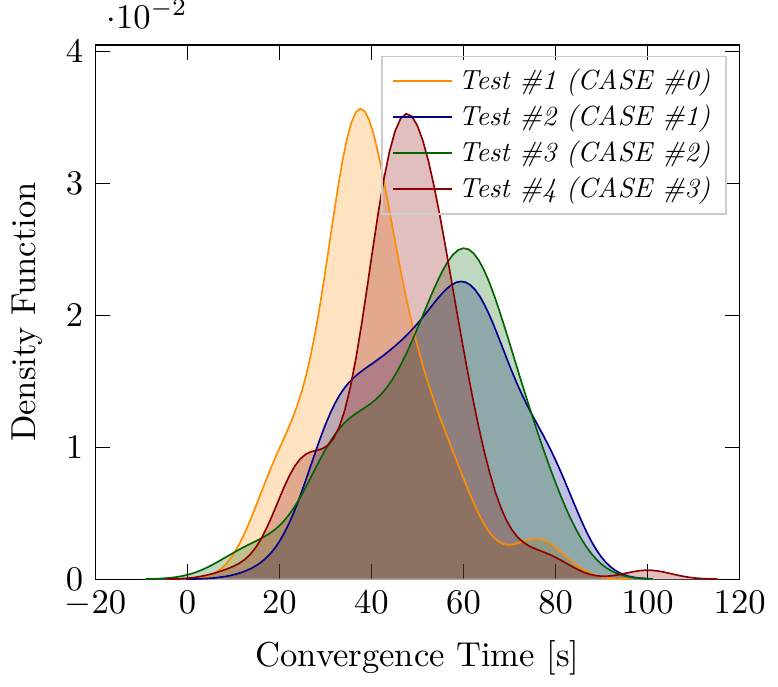}
  \caption{Four cases of ${}^{2d}\widehat{\mathbf{L}}_{\mathbf{s}}$}
  \label{fig:density_4Cases_Convergence-Time}
\end{subfigure}
\par\medskip 
\begin{subfigure}{.25\textwidth}
  \centering
  \includegraphics[width=0.95\columnwidth,height=1.3in]{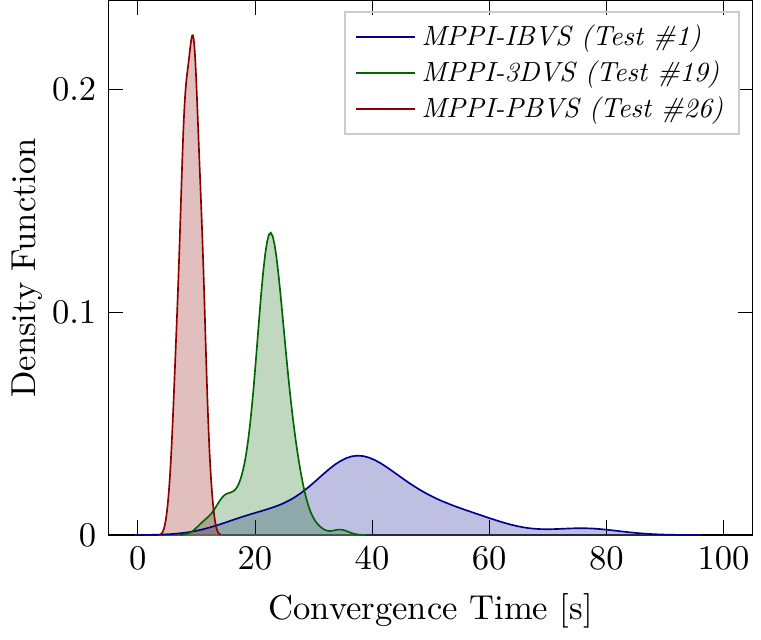}
  \caption{Three \textit{MPPI-VS} schemes}
  \label{fig:density_mppi_3_VS_Convergence-Time}
\end{subfigure}%
\begin{subfigure}{.25\textwidth}
  \centering
  \includegraphics[width=0.95\columnwidth,height=1.3in]{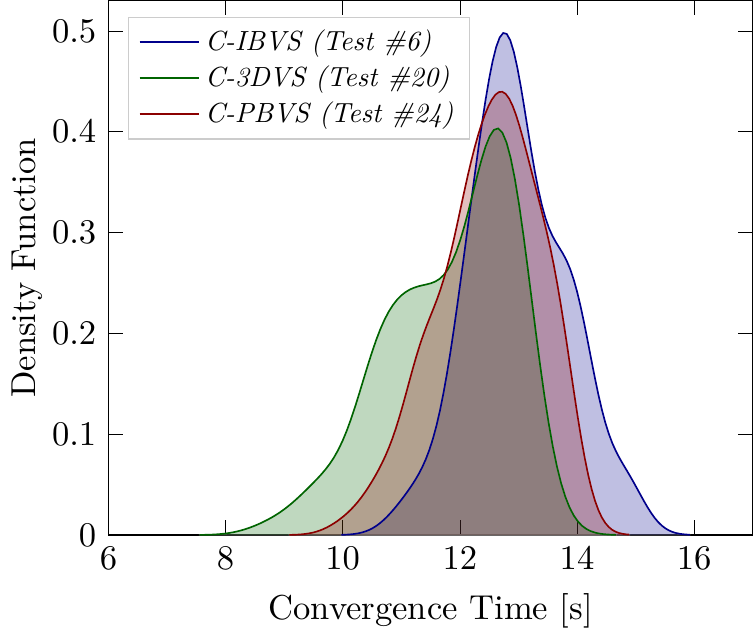}
  \caption{Classical control schemes}
  \label{fig:density_Classic_3_VS_Convergence-Time}
\end{subfigure}%
\caption{Histogram of the convergence time of: (a) three conducted trials of Test \#1, (b) \textit{MPPI-IBVS} considering four cases of ${}^{2d}\widehat{\mathbf{L}}_{\mathbf{s}}$, (c) \textit{MPPI-VS}, and (d) classical control schemes.} 
\label{fig:convergence-time}
\end{figure}

To quantitatively evaluate the performance of the convergence of our proposed control schemes, 
the index of the convergence time of the successful servoing tasks $\mathcal{N}_{\text{success}}$ is first utilized; then, Gaussian distribution of those successful tasks was fitted with a histogram plot as shown in Fig.~\ref{fig:convergence-time}. 
The convergence is considered to be achieved if the error norm ${\lVert \mathbf{e} \lVert}_2$ is less than \SI{0.6}{\pixel} for \textit{MPPI-IBVS} and \SI{3}{\milli\metre} for both \textit{MPPI-3DVS} and \textit{MPPI-PBVS}.
Figures~\ref{fig:density_3trials_Convergence-Time} and \ref{fig:density_4Cases_Convergence-Time} show the histogram of the convergence time of \textit{MPPI-IBVS} for three conducted trials of Test \#1 and the four cases of ${}^{2d}\widehat{\mathbf{L}}_{\mathbf{s}}$ that previously discussed in Section~\ref{classic-ibvs}. From the three trials, we can infer that the majority of those tasks converge to the desired camera configuration within approximately \SI{40.91}{\second}. 
Moreover, as anticipated, the average convergence time of \textit{MPPI-IBVS} considering both CASE \#0 and \#3 is quite shorter than that of CASE \#1 and \#2, thanks to the estimation of the visual features and their depth information. 

In Figs.~\ref{fig:density_mppi_3_VS_Convergence-Time} and~\ref{fig:density_Classic_3_VS_Convergence-Time}, we present the histograms of the convergence time of our proposed control schemes compared to the classical schemes. We can clearly observe that the convergence time of \textit{MPPI-PBVS} is remarkably shorter (with very low standard deviation) than that of \textit{MPPI-IBVS} and classical schemes. This relates to the fact that the control law given in (\ref{eq:mppi_optimal-control}) and the state vector $\mathbf{s}$ of the former control strategy are expressed in the same space (i.e., Cartesian space), while the state vector and, sequentially, the state-dependent running cost function $q(\mathbf{s})$ of the latter strategy are expressed in the image plane which results in a slightly low convergence to the desired camera configuration. 
Furthermore, the standard deviation in the case of \textit{MPPI-IBVS} is quite large since, in some particular configurations, the controller got stuck in the robot joints' limits for few seconds before reaching the desired pose, as previously discussed in Fig~\ref{fig:s45_ibvs}. 

\section{Conclusion and future work}\label{sec:conclusion}
In this work, a \textit{real-time} sampling-based \textit{MPC} strategy (so-called \textit{MPPI-VS}) has been successfully developed for predicting the future behavior of the visual servoing systems such as \textit{IBVS}, \textit{3DVS}, and \textit{PBVS} control schemes, without solving the online optimization problem which usually exceeds the real system-sampling time and suffers from the computational burden.
This control strategy leverages the approximate interaction matrix; i.e., there is no need for estimating the interaction matrix inversion or performing the pseudo-inversion. 
Our proposed control strategy, as well as the classical control strategies, has been successfully tested on realistic and intensive simulations via a 6-DoF Cartesian robot with an \textit{eye-in-hand} camera configuration. 
Through those intensive studies, it is demonstrated that \textit{MPPI-VS} has the following properties. First, it has the capability of coping easily with both hard and soft constraints including visibility, 3D, and control constraints, without adding additional complexity to the optimization problem. Second, contrary to classical \textit{IBVS}, it is highly robust against not only the uncertainties associated with the robot and camera models but also against the measurement noise of the visual features. 
Finally, for \textit{PBVS}, it ensures that the object always remains within the camera’s FoV and, also, ensures a faster convergence rate to the desired pose, compared to other proposed schemes. Our future work will be devoted to experiments on a real robot. 
Moreover, the possibility of running the \textit{MPPI} algorithm on CPUs will be investigated, instead of using GPUs, with the aim of reducing the computational burden.

\section*{Declarations}

\subsubsection*{\textbf{Funding}} This work was supported by the French ANR CLARA (ANR-18-CE33-0004), Université Côte d’Azur, INRIA Sophia Antipolis, France.

\subsubsection*{\textbf{Acknowledgements}} 
The author would like to thank Guillaume Allibert, Philippe Martinet, Fabien Spindler, Ezio Malis, and Lantao Liu for their assistance and valuable comments.

\bibliographystyle{IEEETran}
\bibliography{references}            

\end{document}